\documentclass{dmathesis}

\usepackage{color}
\usepackage{array}
\usepackage{amsmath, amssymb}
\usepackage{amsbsy}
\usepackage{amsfonts}

\usepackage{graphicx}
\usepackage{epstopdf}
\usepackage{hyperref}
\usepackage{float}
\usepackage{natbib}
\setcitestyle{numbers,square}

\usepackage{bibentry}
\usepackage{diagbox}
\usepackage{slashed}
\usepackage{framed}
\usepackage{leftidx}
\usepackage{braket}
\usepackage{bm}
\usepackage{epsfig}
\usepackage{braket}

\usepackage{epigraph}
\usepackage{tcolorbox}
\usepackage{bm}
\usepackage{dsfont}
\usepackage[scr=boondoxo]{mathalfa}
\usepackage{xcolor}

\graphicspath{{Figures/}}
\long\def\/*#1*/{}

\usepackage{tikz}

  \usepackage{wasysym}

\usepackage{graphicx}

\usepackage{fancyhdr}
\usepackage{epsfig}
\usepackage{graphicx}
\usepackage{amsmath}
\usepackage{theorem}
\usepackage{amssymb}
\usepackage{latexsym}
\usepackage{epic}
\usepackage{braket}

\usepackage{lipsum}
\usepackage{graphicx}
\usepackage[english]{babel}
\usepackage{psfrag}
\usepackage{color}
\usepackage{tabularx}
\usepackage{hyperref}
\usepackage{float}

\usepackage{microtype}
\usepackage{graphicx}
\usepackage{booktabs}
\usepackage{url}
\usepackage{multirow}
\usepackage{balance}

\usepackage{subcaption}

\newcounter{ind}

{\theorembodyfont{\rmfamily}}
{\theorembodyfont{\rmfamily}\newtheorem{The}{{\textbf Theorem}}[section]}
{\theorembodyfont{\rmfamily}\newtheorem{Def}[The]{{\textbf Definition}}}
{\theorembodyfont{\rmfamily}}
{\theorembodyfont{\rmfamily}}
{\theorembodyfont{\rmfamily}\newtheorem{Exp}{{\textbf Example}}[section]}

\def\bproof{\textbf{Proof}: }
\def\eproof{\hfill$\Box$}
\def\lsim{\mathrel{\rlap{\lower4pt\hbox{\hskip1pt$\sim$}}\raise1pt\hbox{$<$}}}
\def\gsim{\mathrel{\rlap{\lower4pt\hbox{\hskip1pt$\sim$}}\raise1pt\hbox{$>$}}}

\newcommand{{\evh}}{{E_{\rm vH}}}

\ifx \showCODEN    \undefined \def \showCODEN     #1{\unskip}     \fi
\ifx \showISBNx    \undefined \def \showISBNx     #1{\unskip}     \fi
\ifx \showISBNxiii \undefined \def \showISBNxiii  #1{\unskip}     \fi
\ifx \showISSN     \undefined \def \showISSN      #1{\unskip}     \fi
\ifx \showLCCN     \undefined \def \showLCCN      #1{\unskip}     \fi
\ifx \shownote     \undefined \def \shownote      #1{#1}          \fi
\ifx \showarticletitle \undefined \def \showarticletitle #1{#1}   \fi

\providecommand\bibfield[2]{#2}
\providecommand\bibinfo[2]{#2}
\providecommand\natexlab[1]{#1}
\providecommand\showeprint[2][]{arXiv:#2}

\begin{document}
\renewcommand{\topmargin}{1.1in}
\pagenumbering{roman} \setcounter{page}{1} \pagestyle{empty}
\newpage
\thispagestyle{empty}
\begin{center}
\vspace*{0.0cm}
{\LARGE Boston College\\
     The Graduate School of Arts and Sciences\\
\vspace*{0.6cm}
     Department of Physics}
\end{center}

\begin{center}
\vspace*{1cm}
{\Large \bf Algebraic Learning: Towards Interpretable Information Modeling}
\vspace*{1cm}

{\Large a dissertation\\}

{\Large by\\}
\vspace*{1cm}

  {\large \bf TONG YANG}\vspace*{0.5cm}

 {\Large submitted in partial fulfillment of the requirements\\

           for the degree of\\}

          {\Large Doctor of Philosophy\\}

\vspace*{0.5cm}
  {\Large DEC 2021}
\end{center}

%%%%%%%%%%%%%%%%%%%%%%%%%%%%%%%%%%%%%%%%%%%%%%%%%%%%%%%%%%%%%%%%%%%%%%%%%%%
%%%%%%%%%%%%%%%%%%          The Copyright page           %%%%%%%%%%%%%%%%%%
%%%%%%%%%%%%%%%%%%%%%%%%%%%%%%%%%%%%%%%%%%%%%%%%%%%%%%%%%%%%%%%%%%%%%%%%%%%
\newpage
\thispagestyle{empty}

\begin{center}
\vspace*{15.5cm}
\vspace{2in} \textbf{\copyright\ copyright by TONG YANG}\\
\vspace*{-0.3cm}
 2021
\end{center}

%%%%%%%%%%%%%%%%%%%%%%%%%%%%%%%%%%%%%%%%%%%%%%%%%%%%%%%%%%%%%%%%%%%%%%%%%%%
%%%%%%%%%%%%%%%%%%           The abstract page         %%%%%%%%%%%%%%%%%%%%
%%%%%%%%%%%%%%%%%%%%%%%%%%%%%%%%%%%%%%%%%%%%%%%%%%%%%%%%%%%%%%%%%%%%%%%%%%%
\newpage
\thispagestyle{empty}
%\addcontentsline{toc}{chapter}{\numberline{}Abstract}
\begin{center} 
\vspace*{0.6cm}
{\Large \bf Algebraic Learning: Towards Interpretable Information Modeling}

\vspace*{0.6cm}
\textbf{\Large TONG YANG}

\vspace*{0.5cm}
{\Large Dissertation advisor: Dr. Jan Engelbrecht}
\vspace*{0.5cm}

\textbf{\Huge Abstract}
\end{center}
Along with the proliferation of digital data collected using sensor technologies and a boost of computing power, Deep Learning (DL) based approaches have drawn enormous attention in the past decade due to their impressive performance in extracting complex relations from raw data and representing valuable information.
At the same time, though, rooted in its notorious black-box nature, the appreciation of DL has been highly debated due to the lack of interpretability.
On the one hand, DL only utilizes statistical features contained in raw data while ignoring human knowledge of the underlying system, which results in both data inefficiency and trust issues; on the other hand, a trained DL model does not provide researchers with any extra insight about the underlying system beyond its output, which, however, is the essence of most fields of science, e.g. physics and economics.

The interpretability issue, in fact, has been naturally addressed in physics research.
Conventional physics theories develop models of matter to describe experimentally observed phenomena.
Tasks in DL, instead, can be considered as developing models of information to match with collected datasets.
Motivated by techniques and perspectives in conventional physics, this thesis addresses the issue of interpretability in general information modeling.
%% Align with two drawbacks of DL approaches mentioned above, the current thesis %% endeavors to ease the problem from two scopes.
This thesis endeavors to address the two drawbacks of DL approaches mentioned above.

Firstly, instead of relying on an intuition-driven construction of model structures, a problem-oriented perspective is applied to incorporate knowledge into the modeling practice, where interesting mathematical properties emerge naturally which cast constraints on modeling.
Secondly, given a trained model, various methods could be applied to extract further insights about the underlying system, which is achieved either based on a simplified function approximation of the complex neural network model, or through analyzing the model itself as an effective representation of the system.
These two pathways are termed as \emph{guided model design} (GuiMoD) and \emph{secondary measurements} ($\partial$M), respectively, which, together, present a comprehensive framework to investigate the general field of interpretability in modern Deep Learning practice.  

% The presented framework exhibits the power of mathematics (e.g. algebra, geometry and analysis) in interpretable information modeling, based on which, a novel scheme for statistics based learning practice is proposed: \emph{Deductive Learning} (DeLr).
% Instead of being restricted to the discussion of any specific model structure or dataset, DeLr starts from a modeling task itself and studies idiosyncrasies of a legitimate model class in general.
% Under this novel learning scheme, GuiMoD aims at specifying the legitimate general class, while $\partial$M reveals the characteristic information digested during the learning process.

Remarkably, during the study of GuiMoD, a novel scheme emerges for the modeling practice in statistical learning: \emph{Algebraic Learning} (AgLr).
Instead of being restricted to the discussion of any specific model structure or dataset, AgLr starts from the idiosyncrasies of a learning task itself, and studies the structure of a legitimate model class in general.
This novel modeling scheme demonstrates the noteworthy value of abstract algebra for general artificial intelligence, which has been overlooked in recent progress, and could shed further light on interpretable information modeling by offering practical insights from a formal yet useful perspective.

%%%%%%%%%%%%%%%%%%%%%%%%%%%%%%%%%%%%%%%%%%%%%%%%%%%%%%%%%%%%%%%%%%%%%%%%%%%
%%%%%%%%%%%%%%%%%%     The acknowledgments page         %%%%%%%%%%%%%%%%%%
%%%%%%%%%%%%%%%%%%%%%%%%%%%%%%%%%%%%%%%%%%%%%%%%%%%%%%%%%%%%%%%%%%%%%%%%%%%
\newpage
\textbf{\Huge\\Acknowledgments} 
\vspace*{1.0cm}\pagestyle{plain}

"No man is an Iland, intire of it selfe; every man is a peece of the Continent, a part of the maine."
It was the duty and the responsibility of exploration of the nature that inspired me to serve as a researcher of science in the past six years, which has been a promise to myself since my graduation from the primary school.
Performing the role of a scientist, meanwhile, I have kept querying the true value, if any, of my own work all the time, constantly reminded that my life has already been generously supported by enormous efforts and progress of others on the planet, which could never be taken for granted.
I would, therefore, at the very first place, sincerely acknowledge all the people who, while serving in their own role, have, directly or incidentally, supported me in such a munificent way that I fortunately get an opportunity to serve back. 
Indeed, it has been a journey of serving.
As academia has explicitly become a sector of economy in the modern society, academic research nowadays, although partially motivated by pure curiosity from human nature, has already closely entangled with other sectors.
Far from being talented or creative enough to advance human's fundamental understanding of the universe, I found, until now, the only way that I myself could potentially serve back to produce as much value as possible is to exploit integrated solutions in cross-sector practice.
This could never be possible without the help and support from many special ones in my life.

As the very first person I would like to acknowledge, my advisor, Prof. Jan Engelbrecht, has provided me with incredible supports during my Ph.D. life.
Being one of the most open minded people I have ever met, Jan has been always curious and passionate about new insights and progress from other fields.
With a widely spread spectrum of knowledge, his intellectual competency has been proved in nearly every discussion we had during the past years.
Untypically for a doctor student in physics, my personal interests have broadly extended into various subjects and even industrial and business practices, which would never be possible without Jan's constant encouragement and acts of support.
With more than three years of collaboration associated with random chats, I have been impressed by both his insights about science, and his characters of being humble, patient, and genuine all the time.
Also, importantly, I want to thank Jan's wife, Valerie Wyhte, who is always unbelievably caring and kindhearted, for generously providing me online physical diagnosis and instructions on rehabilitation exercise when I broke both of my ankles in a series of trail-running in the summer of 2020 during the COVID-19 pandemic.

The other special person I would like to stress with gratitude is Prof. Tao Li from Department of Physics in Remin University of China.
Our connection started in a supervisor-student mode during my undergraduate years, and has since then extended way above over many aspects of my life.
He is one of the people I would consult with whenever I find it difficult to make a decision.
The communication between us during the past years has provided me with incredible courage and strength to learn, to doubt, to choose, to reboot, and to progress with improvements.

I am deeply thankful to Prof. Ying Ran at Department of Physics (Boston College), Prof. Renato Mirollo at Department of Mathematics (Boston College), and Prof. Pengyu Hong at Department of Computer Science (Brandeis University), for their bountiful advisory supports across various fields of academics.
Ying had been my first Ph.D. advisor for three years at the beginning of my research life at BC.
When I arrived at Boston in 2015, my research interest was majorly in strongly correlated physics.
Ying provided to me a well-designed training with both invaluable insights and understandings about fundamental philosophies of theoretical physics, and eventually a more generic scheme about defining scientific problems rigorously and plainly step by step.
I was fortunate enough to connect with and learn from Rennie by sneaking into his discussions with Jan casually.
Rennie has clearly demonstrated to me how genius and rigorous a person could be as a mathematician, and is always the one who brings the discussion back into a grounded setup when our thoughts get diverged.
Since 2018, I have been collaborating with Pengyu on research in deep learning and artificial intelligence, who has later become one of the most important mentors in my life.
Pengyu constantly shows a broad interest across various disciplines, and is always open to mentoring and collaborating with students from different background.
Starting from fundamental optimization algorithms in Deep Learning, Pengyu supervised me through numerous projects.
Pengyu and I, together with my closest collaborator Long Sha, have frequently battled on ideas and understandings about innovations and applications of Deep Learning.
Without him offering me the precious collaboration opportunity, I would not have been able to extend my research into cross-discipline studies.

Moreover, I wish to express my sincere gratitude to all my academic collaborators across the Big Boston Area. I would like to thank Prof. Zhijie Xiao at Economics Department (Boston College), Prof. Jos\'e Bento at Computer Science Department (Boston College), Prof. H. Eugene Stanley at Center for Polymer Studies (Boston University), and Prof. Christoph Riedl at Network Science Institute (Northeastern University), for their helpful discussions during collaborations on projects in econometrics, computational biology, network science, and computational sociology, respectively.
I would like to thank Long Sha at Department of Computer Science (Brandeis University) for his patient explanations about numerous Deep Learning and AI concepts during years of collaboration;
and would like to thank Xiangyi Meng at Department of Physics (Boston University) for insightful discussions on quantum and classical information theory when we collaborated since 2018, and also for his instrumental accompaniment in our randomly scheduled after-dinner-entertainment sessions.

Besides, I want to thank my friends and colleagues, either in industry and business sectors, or having a passion about real world impacts, who have offered me a chance to get a glimpse about the application oriented scenarios in real life.
I would like to thank my colleagues at Snowflake during my spring internship in 2021, especially to Swagat Behera for his patient support and help.
I want to thank Guochen Dai, Chaobing Yang, Yifu Liu, Yuan Xu, Ying Lu, Han Zhu, Lihong Xie, Peizhao Li, Shihui Chen, and Jianqiao Zhang, for bringing me with industrial and business insights and career advice, and even co-creating with me some exciting projects and products that are either already launched or to be launched soon in the recent future.

%I was fortunate enough to meet a set of close friends while constantly bothered by research ideas in my head, who have provided me with warmest support and greatest laughter all the time.
I met Bowen Zhao (BU) when taking class at MIT, 
%who was initially deemed by me as a post-doc due to his \textit{over-mature} look.
who then introduced me to two of his high school classmates, Yifeng Qi (MIT) and Jingkai Chen (MIT).
%Yifeng appeared in front of me on a ski-trip, and had impressed me by laughing out loudly, though pleasantly, through the whole day.
%Jingkai, on the other hand, had actually freaked me out with the amount of alcohol he intended to share on our first hangout, after which I decided (temporarily) to quit drink for quite a while.
Although with completely different academic backgrounds, we share moments with each other, especially during the pandemic.
% when all of us eventually decided to step out of academia.
We constantly exchange thoughts and ideas on technologies, industries, trends, and future.
On the edge of graduation, the discussions we had about designs of the upcoming stage have been continuously enlightening me, encouraging me, and inspiring me.

Six years at BC's Department of Physics has been such a great journey thanks to all those lovely people: Xu Yang, Shenghan Jiang, Kun Jiang, Joshua Heath, Xiaodong Hu, Zheng Ren, Xinyue Zhang, Lidong Ma, Yiping Wang, Bryan Rachmilowitz, Bolun Chen, Wenping Cui, Wentao Hou, Hong Li, Andrzej Herczynski, Ziqiang Wang, Yun Peng, and all the faculty and staff members in the department.
Meanwhile, the life in Boston could never be so charming and unforgettable without those precious connections with my close friends: Tong Tong (BC), Han Zhang (Harvard), Xiaoying Lan (BC), Dinghe Cui (BC), Fuxin Zhai (UCI), Hao Li (BC), Jing Ma (BU), Xinzhi Li (Northeastern), Fei Wu (Amazon), Ye Tian (Georgia), Yuanhui Li (Brandeis), Menglin Liu (Google), Yuyue Lou (Edelstein), Junyi Yang (Pointillist), Yang Bai (Deloitte), Qiyuan Sun(E\&Y), and Yifei Wang (Brandeis), with an emphasis on my outdoor partners: Yuwen Zeng (MIT), Yaling Tang (Harvard), Dainel Fu (Nuance), and friends in NASU hiking team, BSSC skiing club, and Ben running club.

Additionally, I would like to give a special "Thank You" to my teammates working together on the voluntary project of 1Point3Acres COVID-19 Information Integration Platform, including but not limited to: Lin Zuo, Peter Sun, Sixuan He, Kai Shen, Pingying Chen, Enyu Li, Jiayue Hu, Yiwen Mo, Weiwei Zhang, Haonan Zhang, Jingxue Chen, Yu Guo, and Wenge (Warald) Wang, for their selfless effort and incredibly decent working ethics in data collection, organization, and presentation tasks, who have attempted their best to minimize the information gap and bring data transparency in the era of COVID-19.

Finally, I would like to express my deepest and warmest gratitude to my grandmother, Xiufang Ma, and my parents, Jinming Yang and Jie Wei, for their endless unconditional support to me during all the days I learnt and served as a student, which, counted from the first day of my primary school, has already been summed up to twenty-one years; and the most unique thanks to my love, Xunmeng Du, for appearing in my life all of a sudden and then participating it all the way through the whole past towards the entire future.

\newpage
\vspace*{4.0cm}

            To My Anchor and My Parents.

%%%%%%%%%%%%%%%%%%%%%%%%%%%%%%%%%%%%%%%%%%%%%%%%%%%%%%%%%%%%%%%%%%%%%%%%%%%
%%%%%%%%    tableofcontents, listoffigures and listoftables       %%%%%%%%%
%%%%%%%%        Command if you do not have  them                  %%%%%%%%%
%%%%%%%%%%%%%%%%%%%%%%%%%%%%%%%%%%%%%%%%%%%%%%%%%%%%%%%%%%%%%%%%%%%%%%%%%%%
\tableofcontents
\renewcommand{\thepage}{\roman{page}}
\listoffigures
\renewcommand{\thepage}{\roman{page}}
%\listoftables
%\renewcommand{\thepage}{\roman{page}}
\clearpage

%%%%%%%%%%%%%%%%%%%%%%%%%%%%%%%%%%%%%%%%%%%%%%%%%%%%%%%%%%%%%%%%%%%%%%%%%%%
%%%%%%%%%%%%%%%%%%%%%%   END OF FRONT PAGE %%%%%%%%%%%%%%%%%%%%%%%%%%%%%%%%
%%%%%%%%%%%%%%%%%%%%%%%%%%%%%%%%%%%%%%%%%%%%%%%%%%%%%%%%%%%%%%%%%%%%%%%%%%%

\begin{sloppypar}
%% Main text
% set page number starts from 1
\pagenumbering{arabic} \setcounter{page}{1}
\setcounter{equation}{0}
\chapter{General Prologue}\label{ch:intro}
% \epigraph{ The best of artists hath no thought to show	\\	
%       which the rough stone in its superfluous shell	\\
%     doth not include; to break the marble spell	\\
%       is all the hand that serves the brain can do.		
% 	}{  {\it-Michelangelo}}

\section{Information Modeling and Deep Learning}\label{sec1:IM_DL}

We describe what we behold.
Science addresses the task of "understanding the world" based on testable explanations and predictions~\cite{science}.
We, as humans, or any specific form of existence, however, could only perceive what we can sense.
Theoretically, there are possible cases where a difference could not be sensed by human or any human-made device, and therefore does not require an "understanding".
The motivation to understand matter fostered the field of physics~\cite{matterandmotion}, where "the world" is perceived in various measurements, including heat, sound, light, force, and so on.
There is, at the same time, another equivalently important component to be understood: \textit{information}.

\begin{figure}[h!]
\centering
\begin{subfigure}{.4\textwidth}
	\centering
	\includegraphics[height = 4.5cm]{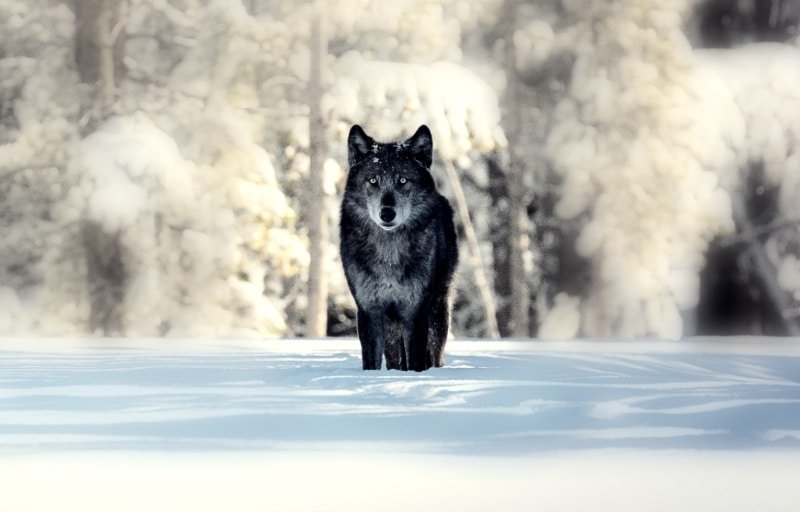}
	\caption{A picture of a wolf.}
	\label{wolf}
\end{subfigure}%
\begin{subfigure}{.6\textwidth}
	\centering
	\includegraphics[height = 4.5cm]{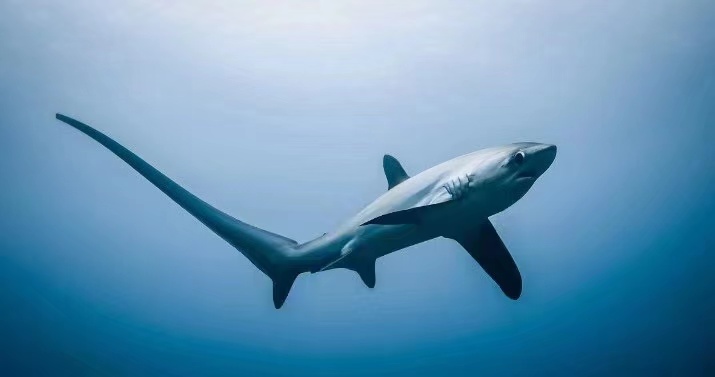}
	\caption{A picture of a thresh shark.}
	\label{shark}
\end{subfigure}
\caption{Different information represented by variables of image pixels.}
\label{pixels_info}
\end{figure}

Analogous to matter, which could be measured by comparing differences based on human sensations, e.g. mass, motion, temperature, or conductivity, information can also be measured by comparing differences based on human conceptualizations\footnote{The term \emph{information} here is more related to the concept of \emph{perception} in academia.}.
For instance, the two pictures in Fig.\ref{pixels_info}, both composed of $1024\times1024$ pixels, can be compared, leading to a conclusion that one represents a wolf and the other represents a thresher shark.

It is sometimes argued that information itself does not require a independent understanding, as its existence is based on matter as carriers and hence is not fundamental.
This statement could not explain the fact that, in many situations, a change of carriers does not change the meaning or conceptualization of information.
Furthermore, it is interesting to notice that modern physics theories are based on quantum mechanics~\cite{griffiths}, where the complex phases of quantum states governs change in time, which describes more about the role of "information" rather than "matter" itself~\cite{wen2015}.
More precisely, the behavior of matter could be better understood by studying its information perspective, which brings another reason to study and understand information itself as a fundamental topic~\cite{wen2015}.

Humans understand the world through \textit{modeling}, which is the procedure of describing observed phenomena using a certain type of language, i.e. the so-called formal science~\cite{hamilton1860}.
In modern science, mathematics has become the official choice to model the world~\cite{wigner1960}, and has been used to describe fundamental causal relationships between observables.
%While conventional physics can be regarded as a science modeling matters, it now becomes inevitable to discuss the general task of \textit{information modeling}.
%It would be helpful to make an analogy to the modeling procedure in physics.
In practice, roughly speaking, a modeling procedure could be decomposed into three general steps: 
\begin{enumerate}
    \item \textbf{data preparation}: experiments or data collections are implemented to acquire data for observables of interests;
    \item \textbf{relation extraction}: a relation between observables are described, usually as a function, through certain fitting procedures with sufficient statistical importance;
    \item \textbf{theoretic modeling}: a mathematical model, either microscopic or phenomenological, would be established to explain the relation observed in Step-2.
\end{enumerate}
Relation extraction is mostly a function fitting which summarizes raw experimental data to compactly describe the observed phenomena; theoretic modeling, in contrast, is the key step towards human understanding, aiming at explaining rather than describing the phenomena.

\begin{figure}
    \centering
    \includegraphics[height = 5cm]{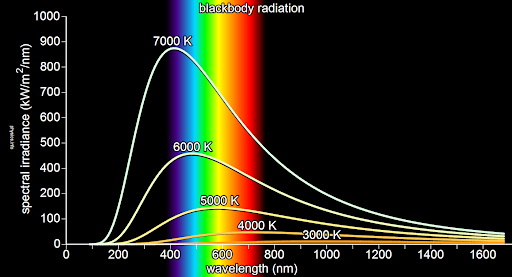}
    \caption{A relation extraction between wavelength and radiance under different temperatures in black-body radiation.}
    \label{fig:blackbody}
\end{figure}

The three steps mentioned above are quite common in traditional physics. 
For instance, in the problem of black-body radiation, experimental data was firstly collected, after which, a interpolation function fitting (as in Fig.~\ref{fig:blackbody}) was used to describe the relation between the wavelength and spectral radiance~\cite{planck1901}.
A energy quanta based theory~\cite{planck1901} was then developed to explain the empirical relation, which precipitated to the development of \emph{quantum mechanics}.
Analogously, information could also be modeled in a similar way.
Instead of recording data from designed experiments, the data preparation step in information modeling is simply the collection of real life data, e.g. images and sentences.
Different from the situation in physics, where experimentally observed variables are usually of low dimension (e.g. temperature and radiance) and a simple function can fit or describe the relation, the collected data in information could be extremely high-dimensional (e.g. pixels in an image and words in an article), and a simple function is not sufficient to capture the relation.

In the past decade, Deep Learning (DL)~\cite{bengio2016} has achieved impressive performance in various tasks~\cite{LeCun2015}.
Taken the setup of supervised learning~\cite{russell2010} for instance, given a collected dataset $\{(\mathbf{x}_i, \mathbf{y}_i)\}_{i=1}^{N}$, where $\mathbf{x}_i$ and $\mathbf{y}_i$ represent independent and dependent variables respectively of the $i$-th sample, a DL method trains a neural network (NN) model $\mathcal{M}_{w}$ to capture the complex relation between an arbitrary input $\mathbf{x}_i$ and the corresponding expected output $\mathbf{y}_i$~\cite{Mohri2012}.
Clearly, DL-based methods target information modeling tasks.
More precisely, DL based methods play the role of relation modeling as defined above in Step-2:
instead of \emph{explaining} the existence of the relation between $\mathbf{x}$ and $\mathbf{y}$, the trained model $\mathcal{M}_{w}$ only \emph{describes} this relation using a complicated NN function.
In other words, DL provides an efficient function fitting for problems with high-dimensional variables~\cite{Mohri2012}, and hence boosts the progress of information modeling.

\begin{table}[!htp]
    \centering
    \begin{tabular}{|c|c|c|}
        \hline
         & Matter & Information \\
        \hline
        Data preparation & experiments & dataset collection \\
        \hline
        Relation modeling & function fitting using regression & neural network fitting using DL \\
        \hline
        Theoretic modeling & physics theories & (\textit{new information theory}) \\
        \hline
    \end{tabular}
    \caption{A comparison between matter modeling and information modeling.}
    \label{tab:modeling}
\end{table}

We exhibit a direct comparison between matter and information modeling in Table.\ref{tab:modeling}. 
Reminded that the relation modeling in Step-2 is implemented to extract a compact description of the raw data, most DL methods nowadays fail to achieve this particular goal: its black-box nature does not support a comprehensive description for humans to capture and use in theoretic modeling later.
On the other hand, DL based methods are indeed practically useful, or even inevitable, as the complexity of relations in information modeling grows exponentially with variable dimension.
This motivates the research of interpretability in DL-based information modeling.

\section{Interpretability in Information Modeling}\label{sec1:inter_IM}
Different from regression-fitted functions in conventional physics, where a straightforward relationship between dependent and independent variables can be derived, the complexity of neural network structures inhibits researchers from gaining insights beyond output values, leaving  further theoretical modeling ungrounded.
Therefore, interpretable components in DL models are demanded. 
Generally speaking, there are two pathways to improve the interpretability of DL.
%explaining a trained black-box model to provide insights about the underlying system.

\subsection{GuiMoD: Guided Model Design}\label{subsec:inter_IM:design}
The first way is to incorporate knowledge into modeling to provide guidance for model design.
In general perception signals, e.g. images, sounds, and sentences, different features of the extremely high-dimensional input space are usually not independent to each other.
Due to this fact, although an explicit relation between variables is absent, there are indeed some \textit{global constraints} that could guide the model design, where the word "global" here refers to the fact that the constraint is not imposed on any single pair of variables or single pair of data samples.
And we term this modeling practice as \emph{guided model design} (GuiMoD).

In fact, there have already been many examples of such guided design of DL models.
For instance, in the problem of image processing~\cite{Gonzalez2018}, the spatial order of different pixels is important, since, obviously, a set of randomly distributed pixels does not compose a meaningful object at all.
This motivates the design of Convolutional Neural Network (CNN), where the convolution operator regards the translation equivariance~\cite{8308187}: when a translation operation is applied in lower layers, higher layer features would transform in a way that keeps the information of this operation\footnote{See a more rigorous mathematical definition associated with detailed explanation in Chapter.3}.
As another typical example, similarly, in natural language processing (NLP)~\cite{manning1999}, the order of words in a sentence is immutable to keep the same semantic meaning.
To incorporate this principle, the well-known Transformer model~\cite{attention} implements a \emph{positional encoding} of words, which applies Fourier series of sinusoidal functions to inject the order information.
By taking advantage of these global constraints, DL modeling is  enhanced in both efficiency and interpretability.

The GuiMoD methodology, in fact, investigates the general ``laws'' dictated by the task itself, and produces a set of constraints on the model structure. 
This eventually derives a general legitimate model class permitted by the task $\mathcal{T}$, which forms a constrained set $\mathcal{M}_{\mathcal{T}}$ from all possible candidates.
This procedure is  analogous to the Projected Symmetry Group approach in condensed matter physics~\cite{PhysRevB.44.2664, PhysRevB.65.165113}.
To construct a low energy effective theory, a mean field method could be studied as an approximation.
The choice of the order parameter specified in the mean field approximation, however, is not fully arbitrary.
In particular, if a certain symmetry is of research interest, then the order parameter should accommodate the concerned symmetry group during model construction, which therefore pre-selects a subset of candidate mean field theories for the modeling practice.
The spirit of  symmetry-constrained model construction is closely aligned with GuiMoD proposed herein.

\subsection{dM: Secondary Measurements}\label{subsec:inter_IM:secondary}
In addition to improving interpretability at the stage of model design, a post-training explanation about a learnt model could also be insightful: 
given a trained neural network function, one may make inference about the underlying system which generates the dataset.

More specifically, one could regard a trained NN model as an efficient, though approximate, representation of the underlying system.
Recall that NN functions in DL serve the role of relation modeling, and that the severest drawback comes from the complexity of the trained function.
Now, if a trained NN itself is treated as an efficient representation, one could instead make further measurements on the NN model, rather than being limited by the access to real life data collection, which, in the present thesis, are termed as \textit{secondary measurements} ($\partial$M).
In practice, $\partial$Ms could be either direct or indirect.

A direct $\partial$M measures the approximated relation between input $\mathbf{x}$ and output $\mathbf{y}$.
In fact, most model-agnostic black-box interpretation techniques~\cite{Tsang2018, zhang2019, Scholbeck2020, molnar2020} belong to this category.
For instance, the well-known framework SHapley Additive exPlanations (SHAP)~\cite{shap} approximate a trained black-box model based on a local expansion, up to the first order of derivatives, around a given data point.
And the resulting SHAP values describes the lowest order relations between input and output.
The family of GA${}^2$M models~\cite{ga2m} further captures interaction between input dimensions and hence produces an improved approximate relation.

%An indirect $\partial$M, in contrast, does not target on the input-output relation itself, but attempts to measure certain observables of interests on the approximate system, i.e. the trained NN model.
An indirect $\partial$M, in contrast, does not target on the input-output relation itself, but attempts to measure certain observables of interests on the \emph{approximate} system, i.e. the trained NN model.
This is extremely valuable in various aspects.
On the one hand, insights can be gained from observables measured in an indirect $\partial$M to assist theoretic modeling of the underlying system in the next stage.
On the other hand, this provides a straight path to black-box model assessment, which are of vital importance in real life scenarios towards safe \& reliable AI~\cite{vassev2016safe, shneiderman2020human}, e.g. model risk evaluation, fairness evaluation, and discrimination detection.
As technology assessment has addressed more and more concerns from  society, this category of research is producing a growing impact across academia, industry, and business sectors.

The $\partial$M probes a trained model to understand its uniqueness brought by the training procedure. 
This could be better explained from an information theory perspective.
Applying GuiMoD, one identifies a set $\mathcal{M}_{\mathcal{T}}$.
Without specifying a training dataset, the probability, over $\mathcal{M}_{\mathcal{T}}$, of an arbitrary model being the optimal one obeys a uniform distribution.
With an increasing amount of instructions dictated by a dataset (e.g. supervised learning or reinforcement learning), the training identifies an single model instance as the optimal one, which gradually changes the distribution over $\mathcal{M}_{\mathcal{T}}$.
The learning therefore is associated with an information change: information is transferred from the dataset into the model space (i.e. $\mathcal{M}_{\mathcal{T}}$).
After training, one regards the trained model as an efficient representation of the underlying data-generation system, and intends to infer the information about the system.
Since the model is guaranteed to be an instance in $\mathcal{M}_{\mathcal{T}}$, a measurement is meaningful only when it probes a characteristic quantity whose value differs on different instances in $\mathcal{M}_{\mathcal{T}}$.

\section{Synopsis of the Thesis}\label{sec1:structure}

Now we delineate the structure of this thesis along with an overview of contents covered herein.
Basically, the present thesis explores the issue of interpretability of information modeling in DL %by means of DeLr, 
using both methodologies of GuiMoD and $\partial$M.
More specifically, Chapter 2 and Chapter 3 propose new model structures following the GuiMoD methodology, while Chapter 4 delivers an indirect $\partial$M on a sequential model trained to capture the underlying dynamics of time series data.
As discussed above, information is carried and represented by high-dimensional signals.
While a generic modeling task in such high-dimensional space is intractable, fortunately, in real world cases, these signals are usually compactly organized by some kind of order, or say, correlations among different dimensions.
There are in general three types of correlations:
1. \emph{discrete correlation} (e.g. logic predicates);
2. \emph{spatial correlation} (e.g. pixels in images);
3. \emph{temporal correlation} (e.g. time series).
Motivated by this categorization, we study one instantiating problem in each correlation category in the following three chapters.

Chapter 2 targets  discrete correlations hidden in chain-rule logic inference, and introduces a novel knowledge graph embedding (KGE)~\cite{8047276} framework based on abstract algebra, which guides the design of the parameter space of embedding models and hence belongs to the first pathway.\footnote{Embedding, in the field of artificial intelligence and machine learning, refers to a high-dimensional array-like representation of the original object of interests.}
On a high-level perspective, we consider the signal of abstract conceptualizations, i.e. knowledge, and explore the global constraints~\cite{Yang2020Learn} which reveal the relations among different knowledge pieces.
Specifically, we observe the existence of a group algebraic structure hidden in relational knowledge embedding problems, which suggests that a group-based embedding framework is essential for designing embedding models. 
We propose a concept that captures the global constraints implied by these embedding tasks:
\textit{hyper-relation}, which, for the first time, provides a unified framework for many previously studied characteristics of knowledge embedding tasks.
The theoretical analysis explores merely the intrinsic property of the embedding problem itself and hence is model agnostic. 
Motivated by the theoretical analysis, a group theory-based knowledge graph embedding framework is proposed, in which relations are embedded as group elements and entities are represented by vectors in group action spaces, together with a generic recipe to construct embedding models associated with two instantiating examples: \textbf{SO3E} and \textbf{SU2E}, both of which apply a continuous non-Abelian group as the relation embedding. 
Empirical experiments using these two exampling models have shown state-of-the-art results on benchmark datasets.

In Chapter 3, another practice of GuiMoD is carried out, with a new model architecture for image processing proposed to accommodate \textit{a priori} knowledge about spatial correlations among pixels.
In particular, we consider the problem of information propagation under geometric transformations of input signals, which raises the concept of \textit{equivariance}~\cite{10.1007/978-3-642-21735-7}.
Equivariance of symmetry transformations is a demanded feature in vision tasks. 
Roughly speaking, when a geometric transformation is operated on an image, one, in general, would require a record of this transformation: 
a model structure that could not differ transformed signals from un-transformed ones would result into an information collapse.
Mathematically, this is usually formulated as an equivariance property of the model.
While most existing works design network structures in real space, in Chapter 3, a spectral representation of signals is discussed, which naturally hosts all irreducible representations of 2-dimensional space groups, and motivates a spectral induced equivariant neural network model structure, termed as \emph{SieNet}. 
The whole network structure is full-symmetry equivariant with respect to all space symmetry operations by design. 
The theoretic analysis and proposed model structure could be easily generalized to arbitrary dimensional spaces as well.

% Chapter.2 and Chapter.3 together could serve as a convincing demonstration about the necessity of the application of mathematical theories of geometry and algebra.

On the other side, given a model trained as a black-box, we investigate it by exploiting the secondary measurement philosophy, which is instantiated in Chapter 4, where an indirect $\partial$M is conduct in the scenario of time series modeling by treating a trained recurrent neural network (RNN)~\cite{Rumelhart1986} model itself as a dynamical system and studying the internal dynamics.
Specifically, a deep learning based training method is introduced, followed by a description of the learnt model from a dynamical system theoretic perspective.
The resulting hybrid provides both a model with promising prediction performance and a comprehension about the trained model itself, which reveals the learnt dynamics by analyzing its asymptotic behavior.
Regarded as an indirect $\partial$M, the analysis provides a way to assess the error accumulation behavior and hence the risk of deviated predictions in applications.
This attempt therefore suggests a mutual benefiting bridge, across which both communities could take advantage of their own knowledge to make contributions in the other field.

All together, we deliver a multi-scope study to the subject of interpretable information modeling, with instantiating studies in several typical fields that covers all major genres of signals, including spatial signals (images), sequential signals (time series), and discrete signals (knowledge and logic).
Interestingly, the three studies presented exploit concepts and techniques from different areas of mathematics:
the embedding framework proposed in Chapter 2 explores idiosyncrasies of general KGE tasks, which reveals an \emph{algebraic} nature of the problem;
Chapter 3 solves the equivariance problem required in image processing tasks by investigating generic \emph{geometric} transformations applicable to an arbitrary image;
while in Chapter 4, a trained RNN model is inspected based on concepts and techniques developed in the dynamical system theory, which is an important branch in \emph{analysis}.
Indeed, covering algebra, geometry, and analysis, the current thesis has demonstrated the power of mathematics in interpretable information modeling, and suggests that further exploration along this direction would be noteworthy.

With all three chapters providing concrete examples, in the last chapter, a highly abstract formal analysis is provided by separately discussing two fundamental modules of general intelligence: \emph{perception module} and \emph{reasoning module}.
In both cases, we demonstrate the emergence of a certain abstract algebraic structure, based on which, we propose a novel unified scheme for statistical learning based modeling practice, termed as \textbf{\emph{Algebraic Learning}} (\textbf{AgLr}).
We argue that, by casting constraints derived from the algebraic structure, AgLr could improve both efficiency and interpretability in generic information modeling practice.

% \newpage

\setcounter{equation}{0}
\chapter{Algebraic Knowledge Graph Embedding}\label{ch:kge}
% \epigraph{ The best of artists hath no thought to show	\\	
%       which the rough stone in its superfluous shell	\\
%     doth not include; to break the marble spell	\\
%       is all the hand that serves the brain can do.		
% 	}{  {\it-Michelangelo}}

In this chapter, we would exhibit our first demonstration of \textit{guided model design} as introduced in the prologue.
More specifically, we consider a relatively abstract problem: knowledge representation.
The signal considered here are general conceptualizations, and the correlation among signals corresponds to logic rules.
For example, from the fact that \textit{"object $A$ is a cat"}, and that \textit{"object $B$ gave birth to $A$"}, one could directly infer that \textit{"object $B$ is a cat"}.
The above three facts could be deemed as three signals, which are correlated by a logic rule that is consistent with commonsense of human. 
Clearly, these rules describe discrete correlations.
To efficiently capture it, we apply an algebra-based theoretic study, which accommodate the intrinsic attribute of the correlations concerned, by considering the most essential problem of knowledge representation: \textit{knowledge graph embedding}.

\section{Background}

Knowledge graphs (KGs) are prominent structured knowledge bases for many downstream semantic tasks~\cite{hao-2017}. A KG contains an entity set $\mathcal{E} = \{e_i\}$, which correspond to vertices in the graph, and a relation set $\mathcal{R} = \{r_k\}$, which forms edges. 
The entity and relation sets form a collection of factual triplets, each of which has the form $(\mathbf{e}_i, \mathbf{r}_k, \mathbf{e}_j)$ where $\mathbf{r}_k$ is the relation between the head entity $\mathbf{e}_i$ and the tail entity $\mathbf{e}_j$.  
Since large scale KGs are usually incomplete due to missing links (relations) amongst entities, an increasing amount of recent works~\cite{transe, dismult, complex, transr} have devoted to the graph completion (i.e., link prediction) problem by exploring a low-dimensional representation of entities and relations. 

More formally, each relation $\mathbf{r}$ acts as a mapping $\mathbf{O}_{\mathbf{r}}[\cdot]$ from its head entity $\mathbf{e}_1$ to its tail entity $\mathbf{e}_2$:
\begin{align}
    \mathbf{r}: \mathbf{e}_1 \mapsto
    \mathbf{O}_{\mathbf{r}}[\mathbf{e}_1]=:\mathbf{e}_2.
\end{align}
The original KG dataset represents these mappings in a tabular form, and the task of knowledge graph embedding (KGE) is to find a better numeric (array-like) representation for these abstract mappings, that can be later efficiently used in computation tasks. 
For example, in the TransE model~\cite{transe}, relations and entities are embedded in the same vector space, and the operation $\mathbf{O}_{\mathbf{r}}[\cdot]$ is simply a vector summation: $\mathbf{O}_{\mathbf{r}}[\mathbf{e}]=\mathbf{e}+\mathbf{r}$. In general, the operation could be either linear or nonlinear, either pre-defined or learned.
Importantly, graph completion relies on the fact that relations are not independent. For example, the hypernym and hyponym are inverse to each other; while kinship relations usually support mutual inferences. These dependencies would impose constraints on the operation design. Previous studies~\cite{rotate, dihedral} have concerned some specific cases of inter-relation dependencies, including (anti-)symmetry and compositional relations. 

Here, however, we attempt to deliver a high-level analysis from a general perspective. More particularly, we ask the following three questions:
\begin{enumerate}
    \item What constraints/requirements does a general KGE task impose on embeddings?
    \item What kind of embedding method would satisfy these constraints?
    \item How to explicitly construct embedding models?
\end{enumerate}
Note that the first question concerns general KGE tasks rather than specific datasets nor embedding models, and, therefore, requires an analysis including all possible knowledge graph structures. We find that, to accommodate all possible KG datasets, there are five requirements for the relation embedding: \textbf{\textit{closure, identity, inverse, associativity,}} and \textbf{\textit{non-commutativity}}. The first four coincide with the algebraic definition of groups in mathematics, and imply a direct answer to the second question: embedding all relations into a group manifold (and designing mapping operations as group actions) would automatically satisfy all requirements; 
in addition, the last requirement, non-commutativity, further suggests implementing non-Abelian groups for the most general KGE tasks.
The third question asks for a general recipe to embed relations as group elements.

One main contribution of this work is it provides a framework for addressing the KG embedding problem from a novel and more rigorous perspective: the group-theoretic perspective. We prove that the intrinsic structure of general KGE tasks coincides with the complete definition of groups. To our best knowledge, this is the first proof that rigorously legitimates the application of group theory in KG embedding. With this framework, we also establish connections with many existing models (see Sec \ref{3.3}), including: TransE~\cite{transe}, TransR~\cite{transr}, TorusE~\cite{toruse}, RotatE~\cite{rotate}, ComplEx~\cite{complex}, DisMult~\cite{dismult}. 

The remaining sections are organized as following: 
Section~\ref{relatedworks} mentions some related works, and emphasizes the distinction between our analysis and others; 
in Section~\ref{general_kge}, we answer the first two question proposed above, achieving a conclusion that continuous non-Abelian groups suit general KGE tasks well, which leads to the exampling continuous non-Abelian group embedding models (\textbf{NagE}) in later sections; 
in Section~\ref{implementing}, we provide a general recipe for group-embedding approach of KGE problems, associated with two novel instantiating models: \textbf{SO3E} and \textbf{SU2E}, which completes the answer for all three questions; 
we demonstrate the power of the proposed embedding framework by comparing the experimental results of our two example models with other state-of-the-art models in Section~\ref{experiments}, and conclude all discussions in Section~\ref{conclusion}.

\section{Related Works}\label{relatedworks}

From the group theory perspective, our work may be related to the TorusE~\cite{toruse}, the RotatE~\cite{rotate}, DihEdral~\cite{dihedral}, and QuatE~\cite{quate} models. 

The TorusE model frames the KG entity embedding in a Lie group manifold to deal with the \emph{compactness problem}. The authors proved that the additive nature of the operation in the TransE model contradicts the entity regularization. However, if a non-compact embedding space is used, entities must be regularized to prevent the divergence of negative scores. Therefore the TorusE model used $n$-torus, a compact manifold, as the entity embedding space. In other words, TorusE embeds entities in group manifolds, while our work embeds relations as group manifolds. 

In the DihEdral model, the $D_{4}$ group the author used plays the same role as groups in our work: group elements correspond to relations. The motivation of DihEdral is to resolve the non-Abelian composition (i.e., the compositional relation formed by $\mathbf{r}_1$ and $\mathbf{r}_2$ would change if the two are switched). Nevertheless, DihEdral applies a discrete group $D_8$ for relation embedding while using a continuous entity embedding space, which may suffer two problems as discussed in the later Section \ref{3.3}. The RotatE model was designed to accommodate symmetric, inversion, and (abelian) composition of triplets at the same time. 
Different from these previous works, our work does not target at one or few specific cases but aims at answering the more general question: finding the generative principle of all possible cases, and thus to provide guidance for model designs that can accommodate all cases. 

More importantly, in most preceding works related to groups~\cite{dihedral, cai2019group}, group theory serves as an alternative perspective to explain the efficiency of specific models; while the theoretical analysis in our current work in Section~\ref{general_kge} is model/dataset independent and is merely initiated by KGE tasks themselves, and the group embedding approach automatically emerges as the natural method which could satisfy all constraints for a general KGE task.
To our the best of our knowledge, this is the first proof that rigorously legitimates the implementation of group theory in KG embeddings.

Another interesting connection refers to QuatE ~\cite{quate} model, where the authors proposed quaternions (also octonions and sedenions in the appendix), which served as an extension of complex numbers for KGE. The intuition of their model was not related to group theory at all. However, in Model Analysis (Section 5.3 in [4]), the authors stated that: “Normalizing the relation to unit quaternion is a critical step for the embedding performance.” While this was empirically observed, an explicit reason is absent. This phenomena can be easily understood from the group theory perspective, as long as one realizes the mathematical correspondence: SU(2) group is an isomorphism to unit quaternions. This explains the necessity of applying “normalization” on quaternions: only unit quaternions are consistent with the group structure (SU(2) in specific), while non-unit ones cannot form a group. One of our newly proposed model, \textbf{SU2E}, is therefore closely related to QuatE model with "unit-scaling", although our proposal does not concern different number systems at all. It is also worth to mention that when comparing with other proceeding models, the authors applied a completely different criteria for the QuatE experiments: a "type constraint" was introduced in their experiments, which filtered out a significant portion of challenging relation types during the evaluation phase. As a contrast, our \textbf{SU2E} model proposed in later sections investigated the similar setting but compared with other models under the common criteria (without "type constraint"), and showed superior results as a \emph{continuous non-Abelian group embedding} for the first time.

\section{Group theory in relational embeddings}\label{general_kge}
In this section, we formulate the group-theoretic analysis for relational embedding problems. 
%For simplicity, our discussion would start from 1-to-1 mappings and discuss the generalization to more complicated cases in the Supplemental Materials.
Firstly, as most embedding models represent objects (including entities and relations) as vectors, the task of operation design thus can be stated as finding proper transformations of vectors.
Secondly, as we mentioned in the introduction, our ultimate goal of reproducing the whole knowledge graph using atomic triplets further requires certain types of local patterns to be accommodated. We now discuss these structures, which in the end naturally leads to the definition of groups in mathematics.

\subsection{Hyper-relation Patterns: relation-of-relations}\label{hyper_relation}

One difficulty of generating the whole knowledge graph from atomic triplets lies in the fact that different relations are not independent of each other. The task of relation inference relies exactly on their mutual dependency. In other words, there exist certain \emph{relation of relations} in the graph structure, which we term as \emph{hyper-relation patterns}. A proper relation embedding method and the associated operations should be able to capture these hyper-relations.

Now instead of studying exampling cases one by one, we ask the most general question: what are the most fundamental hyper-relations? The answer is quite simple and only contains two types, namely, \emph{inversion} and \emph{composition}:
\begin{itemize}
    \item \textbf{Inversion}: given a relation $\mathbf{r}$, there \emph{may exist} an inversion $\mathbf{\bar{r}}$, such that, $\forall \mathbf{e}_1, \mathbf{e}_2\in \mathcal{E}$:
    \begin{align}
        \mathbf{r}:\mathbf{e}_1\mapsto\mathbf{e}_2
        \;\; \longrightarrow \;\; 
        \mathbf{\bar{r}}:\mathbf{e}_2\mapsto\mathbf{e}_1.
    \end{align}
    The inversion captures any relation path with a length equal to 1 (in the unit of relations).
    
    \item \textbf{Composition}: given two relations $\mathbf{r}_1$ and $\mathbf{r}_2$, there \emph{may exist} a third relation $\mathbf{r}_3$, such that, $\forall \mathbf{e}_1, \mathbf{e}_2, \mathbf{e}_3\in \mathcal{E}$:
    \begin{align}\label{composition}
    \left\{
        \centering
        \begin{tabular}{l}
             $\mathbf{r}_1:\mathbf{e}_1\mapsto\mathbf{e}_2$\\ $\mathbf{r}_2:\mathbf{e}_2\mapsto\mathbf{e}_3$
        \end{tabular}
    \right. 
    \quad \longrightarrow \quad 
    \mathbf{r}_3:\mathbf{e}_1\mapsto\mathbf{e}_3.
    \end{align}
    Any relation paths longer than 1 can be captured by a sequence of compositions.
\end{itemize}
One may notice the phrase \emph{may exist} in the above definition, this simply emphasizes that the existence of these derived conceptual relations $\bar{\mathbf{r}}$ and $\mathbf{r}_3$ depends on the \textbf{specific KG dataset}; while, on the other hand, to accommodate general KG datasets, the \textbf{embedding space} should always contains the mathematical representations of these conceptual relations.

An important feature of KG is that with the above two hyper-relations, one could generate any local graph pattern and eventually the whole graph, as relational paths with arbitrary length have been captured.  
Note the term of \emph{inversion} and \emph{composition} might have different meanings from ones in other works: most existing works study triplets to analyze hyper relations, while the definition we provide above is based purely on relations. This is more general in the sense that any conclusion derived would not depend on entities at all, and some different hyper relations could, therefore, be summarized as a single one. For example, there are enormous discussions on \emph{symmetric} triplets and \emph{anti-symmetric} triplets~\cite{rotate}, which are defined as:
\begin{align*}
    \centering
    \left.
    \begin{tabular}{rccr}
        \textbf{symmetric}:  &  $(\mathbf{e}_1, \mathbf{r}, \mathbf{e}_2)$ & $\longrightarrow$ & $(\mathbf{e}_2, \mathbf{r}, \mathbf{e}_1)$,\\
        \textbf{anti-symmetric}:  &  $(\mathbf{e}_1, \mathbf{r}, \mathbf{e}_2)$ & $\longrightarrow$ & $\neg(\mathbf{e}_2, \mathbf{r}, \mathbf{e}_1)$.\\ 
    \end{tabular}
    \right.
\end{align*}
In fact, if for any choice of $\mathbf{e}_{1,2}$, one could produce a symmetric pair of true triplets using $\mathbf{r}$, this would imply a property of $\mathbf{r}$ itself, and in which case, one could then simply derive:
\begin{align}
    \mathbf{\bar{r}} = \mathbf{r}.
\end{align}
This is a special case of the inversion hyper-relation; and similarly, the anti-symmetric case simply implies $\mathbf{\bar{r}} \neq \mathbf{r}$, which is quite common, and does not require extra design. 
The deep reason for discussing hyper-relations which relies merely on relations rather than triplets is that the logic of relation inference problem itself is not entity-dependent.

\subsection{Emergent Group Theory}\label{theory}
To accommodate both general inversions and general compositions, we now derive explicit requirements on the relation embedding model. We start by defining the \emph{product} of two relations $\mathbf{r}_1$ and $\mathbf{r}_2$: $\mathbf{r}_1\cdot\mathbf{r}_2$, as subsequently "finding the tail" twice according to the two relations, i.e. 
\begin{align}\label{product}
\mathbf{O}_{\mathbf{r}_1\cdot\mathbf{r}_2}\big[\cdot\big] :=\mathbf{O}_{\mathbf{r}_1}\big[\mathbf{O}_{\mathbf{r}_2}[\cdot]\big].
\end{align}
With the above definition, \eqref{composition} can be rewritten as: $\mathbf{r}_3 = \mathbf{r}_1 \cdot \mathbf{r}_2$.
One would realize that the following properties should be supported by a proper embedding model:
\begin{enumerate}
    \item \textbf{Inverse element}: to allow the possible existence of inversion, the elements $\mathbf{\bar{r}}$ should also be an element living in the same relation-embedding space\footnote{Given a graph, not all inversions correspond to meaningful relations, but an embedding model should be able to capture this possibility in general.}.
    
    \item \textbf{Closure}: to allow the possible existence of composition, in general, the elements $\mathbf{r}_1 \cdot \mathbf{r}_2$ should also be an element living in the same relation-embedding space\footnote{Given a graph, not all compositions correspond to meaningful relations, but an embedding model should be able to capture this possibility in general.}.
    
    \item \textbf{Identity element}: the possibly existing inversion and composition together define another special and unique relation:
    \begin{align}
        \mathbf{i} = \mathbf{r} \cdot \mathbf{\bar{r}}, \qquad \forall \mathbf{r}\in\mathcal{R}.
    \end{align}
    This element should map any entity to itself, and thus we call it \emph{identity element}.
    
    \item \textbf{Associativity}: In a relational path with the length longer than three (containing three or more relations $\{r_1, r_2, r_3, ...\}$), as long as the sequential order does not change, the following two compositions should produce the same result:
    \begin{align}\label{associativity}
        (\mathbf{r}_1 \cdot \mathbf{r}_2)\cdot \mathbf{r}_3
        = \mathbf{r}_1 \cdot (\mathbf{r}_2\cdot \mathbf{r}_3).
    \end{align}
    The associativity is actually rooted in our definition of $\mathbf{r}_1\cdot\mathbf{r}_2$ in \eqref{product} through the subsequent operating sequence in the entity space,
    from which, we can derive directly that:
    \begin{align}
        \mathbf{O}_{(\mathbf{r}_1 \cdot \mathbf{r}_2)\cdot \mathbf{r}_3}\big[\cdot\big] &= \mathbf{O}_{\mathbf{r}_1\cdot\mathbf{r}_2}\big[ \mathbf{O}_{\mathbf{r}_3}[\cdot]\big]  \nonumber \\
        &= \mathbf{O}_{\mathbf{r}_1}\big[ \mathbf{O}_{\mathbf{r}_2}\big[ \mathbf{O}_{\mathbf{r}_3}[\cdot]\big]\big]\\
        &= \mathbf{O}_{\mathbf{r}_1}\big[ \mathbf{O}_{\mathbf{r}_2\cdot\mathbf{r}_3}[\cdot]\big]
        = \mathbf{O}_{\mathbf{r}_1 \cdot (\mathbf{r}_2\cdot \mathbf{r}_3)}\big[\cdot\big], \nonumber
    \end{align}
    which then leads to the association \eqref{associativity}.
    To help readers understand the practical meaning of associativity in real life cases, here we provide a simple example of the relational associativity: 
    \begin{align*}
        \mathbf{r}_1 = \textbf{\text{isBrotherOf}}, \;  
        \mathbf{r}_2 = \textbf{\text{isMotherOf}}, \;  
        \mathbf{r}_3 = \textbf{\text{isFatherOf}}.
    \end{align*}
    Meanwhile, the following compositions are also meaningful:
    \begin{align*}
        \mathbf{r}_1\cdot\mathbf{r}_2 = \textbf{\text{isUncleOf}}, \quad
        \mathbf{r}_2\cdot\mathbf{r}_3 = \textbf{\text{isGrandmotherOf}}.
    \end{align*}
    In this example, one could easily see that:
    \begin{align*}
        (\mathbf{r}_1 \cdot \mathbf{r}_2)\cdot \mathbf{r}_3
        = \mathbf{r}_1 \cdot (\mathbf{r}_2\cdot \mathbf{r}_3)
        = \textbf{\text{isGranduncleOf}}.
    \end{align*}
    This is a simple demonstration of the associativity.

    \item \textbf{Commutativity/Nonconmmutativity}: In general, commuting two relations in a composition, i.e. $\mathbf{r}_1\cdot\mathbf{r}_2 \leftrightarrow \mathbf{r}_2\cdot\mathbf{r}_1$, may compose either the same or different results.
    We provide a simple illustrative examples for non-commutative compositions. Consider the following real world kinship:
    \begin{align}
        \mathbf{r}_1 = \textbf{\text{isMotherOf}}, \quad  
        \mathbf{r}_2 = \textbf{\text{isFatherOf}}.
    \end{align}
    Clearly, the composition $\mathbf{r}_1\cdot\mathbf{r}_2$ and $\mathbf{r}_2\cdot\mathbf{r}_1$ correspond to $\textit{\text{isGrandmotherOf}}$ and $\textit{\text{isGrandfatherOf}}$ relations respectively, which are different. This is a simple example of non- commutative cases.
    In real graphs, any cases may exist, and a proper embedding method should be able to accommodate both.
\end{enumerate}
The first four properties are exactly the definition of a \textbf{\textit{group}}. In other words, \emph{the group theory automatically emerges from the relational embedding problem itself, rather than being applied manually.} This is a quite convincing evidence that group theory is indeed the most natural language for relational embeddings if one aims at ultimately reproducing all possible local patterns in graphs.
Besides, the fifth property on \emph{commutativity/nonconmmutativity} are actually termed as \textbf{abelian/non-Abelian} in the group theory language. Since abelian is only a special case, to accommodate all possibilities, one should, in general, consider a \textbf{non-Abelian group for the relation embedding}, and guarantee at the same time it contains at least one nontrivial abelian subgroup. We would term the corresponding embedding method as \textbf{NagE: the non-Abelian group embedding method}.

More explicitly, given a graph, to implement a group structure in embedding, one should embed all relations as group elements, which are parametrized by certain group parameters. For instance: 
the translation group $T$ can be parametrized by a real number $\delta$. And correspondingly, due to its vector nature, the embedding of entities could be regarded as a \textbf{representation (rep)} space of the same group. For the translation group, $\mathbb{R}$ (the real field) is a rep space of $T$.

This suggests the group representation theory is useful in knowledge graph embedding problems when talking about entity embeddings, and we leave this as a separate topic for subsequent works later. In the later section, we provide a general recipe for the graph embedding implementation.

\subsection{Embedding models using different groups}\label{3.3}
In this section, we discuss embedding methods using different groups, from simple ones as $T$ (the translation group) and $U(1)$, to complicated ones including $SU(2)$, $GL(n, \mathbb{V})$ (where $\mathbb{V}$ could be any type of fields), or even \textit{Aff}$(\mathbb{V})$. It is important to note that, in practice, \textbf{continuous groups} are more reasonable than discrete ones, due to the two following reasons:
\begin{itemize}
    \item The entity embedding space is usually continuous, which matches reps of the continuous group better. If used to accommodate a discrete group, a continuous space always contains infinite copies of irreducible reps of that group, which makes the analysis much more difficult.
    
    \item When training the embedding models, a gradient-based optimization search would be applied in the parameter space. However, different from continuous groups whose group parameter are also continuous, the parametrization of a discrete group uses discrete values, which brings in extra challenges for the training procedure.
\end{itemize}
With the two reasons above, we thus mainly consider continuous groups which are more reasonable choices. The other important feature of a group is commutativity, which we would mention for each group below. 
Besides the relational embedding group $\mathcal{G}$, the entity embedding space and the similarity measure also need to be determined. As discussed above, the entity embedding should be a proper rep space of $\mathcal{G}$. 
While for similarity measure $d(\cdot)$, we choose among the popular ones including $L_p$-norms ($L_p$) and the $\cos$-similarity ($\cos$), and a complete score function $s_{\mathbf{r}}(\mathbf{e_1}, \mathbf{e_2})$ for a triplet $(\mathbf{e_1}, \mathbf{r}, \mathbf{e_2})$ would be the distance from the relation-transformed head entity to the tail entity. 
One would notice many choices reproduce precedent works, and we show two examples below.

\subsubsection{Example group: \texorpdfstring{$T$}{Lg}}\label{T}
One could use $n$-copies of $T$, the translation group, for the relation embedding. This is a \textit{noncompact abelian} group. The simplest rep-space would be the real field $\mathbb{R}$, which should also appear $n$ times as $\mathbb{R}^n$. The group embedding then produces the following embedding vectors:
\begin{align*}
    \mathbf{e} \quad \Longrightarrow \quad 
    \Vec{v}_{\mathbf{e}}&= \big(x_1, x_2, \;\;\cdots, \;\;x_n\big), \qquad \forall \mathbf{e}\in \mathcal{E}; \nonumber \\
    \mathbf{r} \quad \Longrightarrow \quad 
    \Vec{v}_{\mathbf{r}} &= \big(\delta_1, \; \delta_2, \;\;\,\cdots, \;\;\delta_n\big), \qquad \forall \mathbf{r}\in \mathcal{R};
\end{align*}
both of which are $n$-dim. Here both $x_i$ and $\delta_i$ are real numbers. In a triplet, the relation $\Vec{v}_r$ acts as an addition vector added to the head entity $\mathbf{e}_1$.
If one further chooses $L_p$-norm as the similarity measure, the complete score function $s_{\mathbf{r}}(\mathbf{e_1}, \mathbf{e_2})$ would be:
\begin{align}
    \|(\Vec{v}_{\mathbf{r}} + \Vec{v}_{\mathbf{e_1}}) - \Vec{v}_{\mathbf{e_2}}\|_p,
\end{align}
this actually corresponds to the well-known \textbf{TransE} model~\cite{transe}.
There was a regularization in the original TransE model that changes the entity rep-space, which however has been removed in many later works by properly bounding the negative scores.

\subsubsection{Example group: \texorpdfstring{$U(1)$}{Lg}}\label{u1}

One could use $n$-copies of $U(1)$, the 1-dim unitary transformation group, for the relational embedding. This is a \textit{compact abelian} group. 
The simplest rep-space would be the real field $\mathbb{C}$, which should also appear $n$ times as $\mathbb{C}^n$. The group embedding then produces the following embedding vectors:
\begin{align*}
    \mathbf{e} \quad \Longrightarrow \quad 
    \Vec{v}_{\mathbf{e}}&= \big(x_1, x_2, \;\;\cdots, \;\;x_n\big), \qquad \forall \mathbf{e}\in \mathcal{E}; \nonumber \\
    \mathbf{r} \quad \Longrightarrow \quad 
    \Vec{v}_{\mathbf{r}} &= \big(\phi_1,  \phi_2, \;\;\,\cdots, \;\phi_n\big), \qquad \forall \mathbf{r}\in \mathcal{R};
\end{align*}
where $x_i$ is a complex number containing a both real and imaginary part, while $\phi_i$ is a phase variable take values from $0$ to $2\pi$. Therefore the entity-embedding dimension is $2n$, while the relation dimension is $n$.
In a triplet, the relation $\Vec{v}_{\mathbf{r}}$ acts as a phase shift on the head entity $\mathbf{e}_1$. In a matrix form, one could define $\mathbf{R}_{\mathbf{r}}$ as the diagonal matrix with the $i$-th diagonal element being $e^{i\phi_i}$.
If one further chooses $L_p$-norm as the similarity measure, the complete score function $s_{\mathbf{r}}(\mathbf{e_1}, \mathbf{e_2})$ would be:
\begin{align}
    \|\mathbf{R}_{\mathbf{r}}\cdot\Vec{v}_{\mathbf{e}_1} - \Vec{v}_{\mathbf{e}_2}\|_p =
    \|\big[e^{i\Vec{v}_{\mathbf{r}}}\big]\circ\Vec{v}_{\mathbf{e}_1} - \Vec{v}_{\mathbf{e}_2}\|_p,
\end{align}
where $\circ$ means a Hadamard product.
This precisely leads to the \textbf{RotatE} model~\cite{rotate}. 

On the other hand, one could also use the $n$-torus $\mathbb{T}^n$ as the rep-space:
\begin{align*}
    \mathbf{e} \quad \Longrightarrow \quad 
    \Vec{v}_{\mathbf{e}}&= \big(\theta_1, \theta_2, \;\;\cdots, \;\;\theta_n\big), \qquad \forall \mathbf{e}\in \mathcal{E}; \nonumber \\
    \mathbf{r} \quad \Longrightarrow \quad 
    \Vec{v}_{\mathbf{r}} &= \big(\phi_1,  \phi_2, \;\;\,\cdots, \,\phi_n\big), \qquad \forall \mathbf{r}\in \mathcal{R};
\end{align*}
where $\theta_i$ represents a coordinate on the torus. Still using the $L_p$-norm similarity measure, the complete score function $s_{\mathbf{r}}(\mathbf{e_1}, \mathbf{e_2})$ now is:
\begin{align}
    \|\mathbf{R}_{\mathbf{r}}\cdot\Vec{v}_{\mathbf{e}_1} - \Vec{v}_{\mathbf{e}_2}\|_p =
    \|e^{i\Vec{v}_{\mathbf{r}}}\circ e^{i\Vec{v}_{\mathbf{e}_1}} - e^{i\Vec{v}_{\mathbf{e}_2}}\|_p,
\end{align}
which leads to the \textbf{TorusE} model~\cite{toruse}.
In the original implementation of TorusE, there is an additional projection $\pi$ from $\mathbb{R}^n$ to $\mathbb{T}^n$.\footnote{Due to the special relation between $T$ and $U(1)$, i.e. $U(1)\cong T/(2\pi\mathbb{Z})$, one could also regard TorusE as an implementation of group-embedding with $T$, which is more similar to the motivation in the original paper~\cite{toruse}.}

\subsubsection{A summary of some example groups}
We summarize the results of several chosen examples in Table~\ref{tab:groups}.

\begin{table*}[!ht]
\centering
\begin{tabular}{|c|c|c|c|c|c|}
\hline
Group & Space & Abelian & $d(\cdot)$ & Studied & Related Work \\
\hline
$T$ & $\mathbb{R}^n$ & YES & $L_p$ & $\checkmark$ & TransE~\cite{transe}\\
\hline
$U(1)$  & $\mathbb{C}^n$ & YES & $L_p$ & $\checkmark$ & RotatE~\cite{rotate}\\
\hline
$U(1)$  & $\mathbb{T}^n$ & YES & $L_p$ & $\checkmark$ & TorusE~\cite{toruse}\\
\hline
$SO(3)$  & $\mathbb{R}^{3 n}$ & NO & $L_p$ & -- & --\\
\hline
$SU(2)$  & $\mathbb{C}^{2 n}$ & NO & $L_p$ & -- & -- \\
\hline
$GL(1, \mathbb{R})$  & $\mathbb{R}^n$ & YES & $\cos$ & $\checkmark$ & DisMult~\cite{dismult}\\
\hline
$GL(1, \mathbb{C})$  & $\mathbb{C}^n$ & YES & $\cos$ & $\checkmark$ & ComplEx~\cite{complex}\\
\hline
$GL(n, \mathbb{R})$  & $\mathbb{R}^n$ & NO & $\cos$ & -- & RESCAL~\cite{rescal}\\
\hline
\textit{Aff}$(\mathbb{R}^n)$ & $\mathbb{R}^n$ & NO & $L_p$ & -- & TransR~\cite{transr}\\
\hline
$D_4$  & $\mathbb{R}^n$ & NO & $L_p$ & $\checkmark$ & DihEdral~\cite{dihedral}\\
\hline
\end{tabular}
\caption{Examples of the group embedding.}
\label{tab:groups}
\end{table*}

Note in the Table~\ref{tab:groups}, some groups have not been studied, but there are still some existing models which use a quite similar embedding space; while the major gap, between the existing models and their group embedding counterparts, is the constraint of group structures on the parametrization.
For example, implementing group embedding with $GL(n, \mathbb{R})$, the $n$-dim general linear groups defined on field-$\mathbb{R}$, would lead to a model similar to RESCAL~\cite{rescal}. However, the original RESCAL model does not have a built-in group structure: it uses arbitrary $n\times n$ real matrices, some of which may not be invertible, and hence are not group elements in $GL(n,\mathbb{R})$. It is, therefore, worth to add the extra invertible constraint in RESCAL, which requests matrices constructed through group parametrization rather than assigned arbitrary matrix elements. A similar analysis holds for the affine group Aff$(\mathbb{R}^n)$.

\section{Group Embedding for Knowledge Graphs}\label{implementing}
In this section, we would firstly provide a general recipe for the group embedding implementation, and then provide two explicit examples of \textbf{NagE}, both of which apply a continuous non-Abelian group that has not been studied in any precedent works before.

\subsection{A general group embedding recipe}\label{recipe}
We summarize the group embedding procedure as following:
\begin{enumerate}
    \item Given a graph, choose a proper group $\mathcal{G}$ for embedding. The choice may concern property of the task, such as commutativity and so on. And as stated above, in most general cases, a non-Abelian continuous group should be proper.
    \item Choose a rep-space for the entity embedding. For simplicity, one could use multiple ($n$) copies of the same rep $\rho$, which is the case of most existing works. Suppose $\rho$ is a $p$-dim rep, then the total dimension of entity embedding would be $pn$, which is written as a vector $\Vec{v}_{\mathbf{e}}$. Roughly speaking, $k$ captures the relational structure and $n$ encodes other feature.
    \item Choose a proper parametrization of $\mathcal{G}$, that is, choose a set of parameters indexing all group elements in $\mathcal{G}$. Suppose the number of parameters required to specify a group element is $q$, then the total dimension of relation embedding $\Vec{v}_{\mathbf{r}}$ would be $qn$. A group element can now be expressed as a block-diagonal matrix $\mathbf{R}_{\mathbf{r}}$, with each block $\mathbf{M}_i$ being a $p\times p$ matrix whose entries are determined by the vector $\Vec{v}_{\mathbf{r}}$.
    \item Choose a similarity measure $d(\cdot)$, the score value $s_{\mathbf{r}}(\mathbf{e}_1,\mathbf{e}_2)$ of a triplet $(\mathbf{e}_1,\mathbf{r},\mathbf{e}_2)$ is then:
    \begin{align}
        s_{\mathbf{r}}(\mathbf{e}_1,\mathbf{e}_2) \equiv d\big( \mathbf{R}_{\mathbf{r}}\cdot\Vec{v}_{\mathbf{e}_1}, \;\; \Vec{v}_{\mathbf{e}_2} \big)
    \end{align}
\end{enumerate}
Below we demonstrate the group embedding approach by implementing it with exampling continuous non-Abelian groups. As shown in table~\ref{tab:groups}, two simple continuous non-Abelian groups that have not been studied are $SO(3)$ and $SU(2)$, we will implement them as relation embedding manifolds, which, as a result, produce two \textbf{NagE models}.

\subsection{\textbf{SO3E}: NagE with group \texorpdfstring{$SO(3)$}{Lg}}
The 3D special orthogonal group $SO(3)$ is one of the simplest continuous non-Abelian group. 
As an illustrative demonstration, we construct an embedding model with $SO(3)$ structure and implement it in real experiments. Following the general recipe above, after determining the group $\mathcal{G}=SO(3)$, we choose a proper rep-space for entity embedding: $[\mathbb{R}^3]^{\otimes n}$, which consists $n$-copies of $\mathbb{R}^3$. Each $\mathbb{R}^3$ subspace transforms as the standard rep-space of $SO(3)$. All relations thus act as $3n\times 3n$ block diagonal matrix, with each block being a $3\times 3$ complex matrix carrying the standard representation of $SO(3)$.

Next, we choose a proper parametrization of $SO(3)$. Instead of the more general angular momentum parametrization, due to our choice of using the standard representation, we could parameterize the $SO(3)$ elements using Euler angles $(\phi, \theta, \psi)$, which is easier for implementation.

Put all together, our group embedding is then fixed as:
\begin{align*}
    \mathbf{e} \; \Longrightarrow \; 
    \Vec{v}_{\mathbf{e}}&= \big(x_1, y_1, z_1, \;\;\cdots, \;\;x_n, y_n, z_n\big), \quad \forall \mathbf{e}\in \mathcal{E}; \nonumber \\
    \mathbf{r} \; \Longrightarrow \; 
    \Vec{v}_{\mathbf{r}} &= \big(\phi_1, \theta_1, \psi_1, \;\; \cdots, \;\phi_n, \theta_n, \psi_n\big), \quad \forall \mathbf{r}\in \mathcal{R};
\end{align*}
both of which are $3n$-dim. In a triplet, the relation vector $\Vec{v}_{\mathbf{r}}$ acts as a block diagonal matrix $\mathbf{R}_{\mathbf{r}}$, with each block matrix $M_i$ acting in the subspace of $(x_i, y_i, z_i)$:

\begin{align}
    \setlength{\tabcolsep}{10pt}
    \centering
    \left[
    \begin{tabular}{ccc|ccc|ccc|ccc}
     {} & {} & {} & {} & {} & {} & {} & {} & {} & {} & {} & {} \\
     {} & $\mathbf{M}_1$ & {} & {} & $0$ & {} & {} & $\ldots$ & {} & {} & $0$ & {} \\
     {} & {} & {} & {} & {} & {} & {} & {} & {} & {} & {} & {} \\
     \hline
     {} & {} & {} & {} & {} & {} & {} & {} & {} & {} & {} & {} \\
     {} & $0$ & {} & {} & $\mathbf{M}_2$ & {} & {} & $\ldots$ & {} & {} & $0$ & {} \\
     {} & {} & {} & {} & {} & {} & {} & {} & {} & {} & {} & {} \\
     \hline
     {} & {} & {} & {} & {} & {} & {} & {} & {} & {} & {} & {} \\
     {} & $\vdots$ & {} & {} & $\vdots$ & {} & {} & $\ddots$ & {} & {} & $\vdots$ & {} \\
     {} & {} & {} & {} & {} & {} & {} & {} & {} & {} & {} & {} \\
     \hline
     {} & {} & {} & {} & {} & {} & {} & {} & {} & {} & {} & {} \\
     {} & $0$ & {} & {} & $0$ & {} & {} & $\ldots$ & {} & {} & $\mathbf{M}_n$ & {} \\
     {} & {} & {} & {} & {} & {} & {} & {} & {} & {} & {} & {} 
    \end{tabular}
    \right]
    \left[
    \begin{tabular}{c}
         $x_1$ \\
         $y_1$ \\
         $z_1$ \\
         \hline
         $x_2$ \\
         $y_2$ \\
         $z_2$ \\
         \hline
         \\
         \vdots \\
         \\
         \hline
         $x_n$ \\
         $y_n$ \\
         $z_n$
    \end{tabular}
    \right].
\end{align}
And each $3\times 3$ block $\mathbf{M}_i$ is parametrized as following:
\begin{align}
    \mathbf{M}_i^{11} &= \quad\cos{\psi_i}\cos{\phi_i} - \cos{\theta_i}\sin{\psi_i}\sin{\phi_i},\nonumber\\ 
    \mathbf{M}_i^{12} &= \quad\cos{\psi_i}\sin{\phi_i} + \cos{\theta_i}\cos{\psi_i}\cos{\phi_i},\nonumber\\ 
    \mathbf{M}_i^{13} &= \quad\sin{\psi_i}\sin{\theta_i},\nonumber\\ 
    \mathbf{M}_i^{21} &=-\sin{\psi_i}\cos{\phi_i} - \cos{\theta_i}\sin{\psi_i}\cos{\phi_i},\nonumber\\ 
    \mathbf{M}_i^{22} &=-\sin{\psi_i}\sin{\phi_i} + \cos{\theta_i}\cos{\psi_i}\cos{\phi_i},\nonumber\\ 
    \mathbf{M}_i^{23} &= \quad\cos{\psi_i}\sin{\theta_i},\nonumber\\
    \mathbf{M}_i^{31} &= \quad\cos{\psi_i}\sin{\theta_i}, \nonumber\\ 
    \mathbf{M}_i^{32} &=-\cos{\psi_i}\cos{\theta_i}, \qquad
    \mathbf{M}_i^{33} = \cos{\theta_i}.
\end{align}

\subsection{\textbf{SU2E}: NagE with group \texorpdfstring{$SU(2)$}{Lg}}
The 2D special unitary group $SU(2)$ is another simple continuous non-Abelian group. 
We choose the rep-space for entity embedding as: $[\mathbb{C}^2]^{\otimes n}$, which consists $n$-copies of $\mathbb{C}^2$. Each $\mathbb{C}^2$ subspace transforms as the standard rep-space of $SU(2)$. All relations thus act as $2n\times 2n$ block diagonal matrix, with each block being a $2\times 2$ complex matrix carrying the standard representation of $SU(2)$.

Next, we choose a proper parametrization of $SU(2)$. An analysis with the corresponding Lie algebra $\mathfrak{su}(2)$ shows that any group element could be written as~\cite{lie}:
\begin{align}
    e^{i\alpha [\hat{\mathbf{n}}\cdot\Vec{\mathbf{J}}]} = 
    \cos{\alpha}\hat{1} + i\sin{\alpha}\hat{\mathbf{n}}\cdot\Vec{\mathbf{J}},
\end{align}
where $\alpha$ is a rotation angle taken from $[0,2\pi]$, and $\hat{\mathbf{n}}$ is a unit vector on $S^2$, represented by two other angles $(\theta, \phi)$; moreover, the symbol $\hat{1}$ means an identity matrix, and $\Vec{\mathbf{J}}$ are three generators of the group: $(\mathbf{J}_x, \mathbf{J}_y, \mathbf{J}_z)$, which, in the standard rep have the following form:
\begin{align*}
    \centering
    J_x = 
    \left[
    \begin{tabular}{c c}
    0 & 1 \\
    1 & 0
    \end{tabular}
    \right],\; 
    J_y = 
    \left[
    \begin{tabular}{c c}
    0 & -i \\
    i & 0
    \end{tabular}
    \right],\; 
    J_z = 
    \left[
    \begin{tabular}{c c}
    1 & 0 \\
    0 & -1
    \end{tabular}
    \right].
\end{align*}
Put all together, our group embedding is then fixed as:
\begin{align*}
    \mathbf{e} \; \Longrightarrow \; 
    \Vec{v}_{\mathbf{e}}&= \big(x_1, y_1, \; x_2, y_2, \;\cdots, \;x_n, y_n\big), \quad \forall \mathbf{e}\in \mathcal{E}; \nonumber \\
    \mathbf{r} \; \Longrightarrow \; 
    \Vec{v}_{\mathbf{r}} &= \big(\alpha_1, \hat{\mathbf{n}}_1, \alpha_2, \hat{\mathbf{n}}_2,\; \cdots, \alpha_n, \hat{\mathbf{n}}_n\big), \quad \forall \mathbf{r}\in \mathcal{R};
\end{align*}
where $x_i$ and $y_i$ are complex numbers, and $\alpha_i$ and $\hat{\mathbf{n}}_i=(\theta_i, \phi_i)$ represent angles. 
In a triplet, the relation $\Vec{v}_{\mathbf{r}}$ acts as a block diagonal matrix $\mathbf{R}_{\mathbf{r}}$, with each block matrix $M_i$ acting in the subspace of $(x_i, y_i)$. And each $2\times 2$ block $\mathbf{M}_i$ is parametrized as~\cite{lie}:
\begin{align*}
    \centering
    \left[
    \begin{tabular}{c c}
         $\cos{\alpha_i} + i\sin{\alpha_i}\sin{\theta_i}$ & 
         $ie^{-i\phi_i}\cdot\sin{\alpha_i}\cos{\theta_i}$
         \\
         $ie^{-i\phi_i}\cdot\sin{\alpha_i}\cos{\theta_i}$ & 
         $\cos{\alpha_i} - i\sin{\alpha_i}\sin{\theta_i}$
    \end{tabular}
    \right]. 
\end{align*}

\subsection{Similarity measure and loss function}
We choose $L_2$-norm as the similarity measure $d(\cdot)$ to compute the score value:
\begin{align}
     s_{\mathbf{r}}(\mathbf{e}_1,\mathbf{e}_2) = \|\mathbf{R}_{\mathbf{r}}\cdot\Vec{v}_{\mathbf{e}_1} - \Vec{v}_{\mathbf{e}_2}\|_2
\end{align}
We design the model loss function for a triple $(\mathbf{e_1}, \mathbf{r}, \mathbf{e_2})$ as follows:
\begin{align*}
    &L =-\log{\sigma\left[\gamma-s_{\mathbf{r}}(\mathbf{e_1}, \mathbf{e_2})\right]}\nonumber\\
    &\qquad -\sum_{i=1}^{n} p\left(\mathbf{e_{1i}^{\prime}}, \mathbf{r}, \mathbf{e_{2i}^{\prime}}\right) \log{\sigma\left[s_r\left(\mathbf{e_{1i}}^{\prime}, \mathbf{e_{2i}}^{\prime}\right)-\gamma\right]} \\
    &p\left(\mathbf{e_{1j}^{\prime}}, \mathbf{r}, \mathbf{e_{2j}^{\prime}} |\left\{\left(\mathbf{e_{1i}}, \mathbf{r}, \mathbf{e_{2i}}\right)\right\}\right) =\frac{e^{\alpha[\gamma-s_{\mathbf{r}}(\mathbf{e'_{1j}}, \mathbf{e'_{2j}})]}}{\sum_{i} e^{\alpha [\gamma-s_{\mathbf{r}}(\mathbf{e'_{1i}}, \mathbf{e'_{2i}})]}}
\end{align*}
where $\sigma$ is the Sigmoid function, $\gamma$ is the margin used to prevent over-fitting. $\mathbf{e_{1i}^{\prime}}$ and $\mathbf{e_{2i}^{\prime}}$ are negative samples while $p\left(\mathbf{e_{1i}^{\prime}}, r, \mathbf{e_{2i}^{\prime}}\right)$ is the adversarial sampling mechanism with temperature $\alpha$ we adopt self-adversarial negative sampling setting from ~\cite{rotate}. We term the resulting model as \textbf{SO3E} and \textbf{SU2E} respectively for the above two groups.
We mention other implementation details in the next section.

\section{Experiments}\label{experiments}

\subsection{Experimental Setup}

\paragraph{Datasets: }

\begin{table*}[!ht]
\setlength{\tabcolsep}{2pt}
\begin{center}
\begin{tabular}{|c|c|cccc|cccc|c|}
\hline
\multirow{2}*{Group} & \multirow{2}*{Commutativity} & \multicolumn{4}{|c|}{WN18}  & \multicolumn{4}{|c|}{FB15k} & \multirow{2}*{Example} \\ \cline{3-10}

& & MRR & H@1 & H@3 & H@10 & MRR & H@1 & H@3 & H@10 & \\ \hline

T & Abelian & 0.495 & 0.113 & 0.888 & 0.943 & 0.463 & 0.297 & 0.578 & 0.749 & TransE \\
U$(1)$ & Abelian & \underline{0.949} & \textbf{0.944} & 0.952 & \underline{0.959} & \textbf{0.797} & \textbf{0.746} & \underline{0.830} & \underline{0.884} & RotatE \\ 
U$(1)$ & Abelian & 0.947 & \underline{0.943} & 0.950 & 0.954 & 0.733 & 0.674 & 0.771 & 0.832 & TorusE \\  
GL$(1,\mathbb{R})$ & Abelian & 0.822 & 0.728 & 0.914 & 0.936 & 0.654 & 0.546 & 0.733 & 0.824 & DistMult \\ 
GL$(1,\mathbb{C})$ & Abelian & 0.946 & 0.942 & 0.949 & 0.954 & 0.692 & 0.599 & 0.759 & 0.840 & ComplEx \\ \hline
SO$(3)$ & non-Abelian & \textbf{0.950} & \textbf{0.944} & \underline{0.953} & \textbf{0.960} & \underline{0.794} & \underline{0.740} & \textbf{0.831} & \textbf{0.886} & SO3E \\ 
SU$(2)$ & non-Abelian & \textbf{0.950} & \textbf{0.944} & \textbf{0.954} & \textbf{0.960} & 0.791 & 0.734 & \textbf{0.831} & \textbf{0.886} & SU2E \\ \hline
\end{tabular}
\caption{\label{tab:1815}Link prediction on WN18 and FB15k (\textbf{bold} represent the best scores, \underline{underlined} represent the second best).}
\end{center}
\end{table*}

\begin{table*}[!ht]
\setlength{\tabcolsep}{2pt}
\begin{center}
\begin{tabular}{|c|c|cccc|cccc|c|}
\hline
\multirow{2}*{Group} & \multirow{2}*{Commutativity} & \multicolumn{4}{|c|}{WN18RR}  & \multicolumn{4}{|c|}{FB15k-237} & \multirow{2}*{Example} \\ \cline{3-10}

& & MRR & H@1 & H@3 & H@10 & MRR & H@1 & H@3 & H@10 & \\ \hline

T & Abelian & 0.226 & - & - & 0.501 & 0.294 & - & - & 0.465 & TransE \\
U$(1)$ & Abelian & \underline{0.476} & 0.428 & \underline{0.492} & 0.571 & \underline{0.338} & 0.241 & 0.375 & \textbf{0.533} & RotatE \\ 
GL$(1,\mathbb{R})$ & Abelian & 0.430 & 0.390 & 0.440 & 0.490 & 0.241 & 0.155 & 0.263 & 0.419 & DistMult \\ 
GL$(1,\mathbb{C})$ & Abelian & 0.440 & 0.410 & 0.460 & 0.510 & 0.247 & 0.158 & 0.275 & 0.428 & ComplEx \\ \hline
SO$(3)$ & non-Abelian & \textbf{0.477} & \textbf{0.432} & \textbf{0.493} & \underline{0.574} & \textbf{0.340} & \textbf{0.244} & \textbf{0.378} & 0.530 & SO3E \\ 
SU$(2)$ & non-Abelian & \underline{0.476} & \underline{0.429} & \textbf{0.493} & \textbf{0.575} & \textbf{0.340} & \underline{0.243} & \underline{0.376} & \underline{0.532} & SU2E \\ \hline
\end{tabular}
\caption{\label{tab:237rr}Link prediction on WN18RR and FB15k-237 (\textbf{bold} represent the best scores, \underline{underlined} represent the second best).}
\end{center}
\end{table*}

The most popular public knowledge graph datasets include FB15K~\cite{freebase} and  WN18~\cite{wordnet}. FB15K-237~\cite{fb237} and WN18RR~\cite{wnrr} datasets were derived from these two, in which the inverse relations were removed. FB15K dataset is a huge knowledge base with general facts containing 1.2 billion instances of more than 80 million entities. For benchmarking, usually, a frequency filter was applied to obtain occurrence larger than 100 resulting in 592,213 instances with 14,951 entities and 1,345 relation types. WN18 was extracted from WordNet~\cite{wordnet} dictionary and thesaurus, the entities are word senses and the relations are lexical relations between them. It has 151,442 instances with 40,943 entities and 18 relation types.
\paragraph{Evaluation Protocols: }
We use three categories of protocols for evaluations, namely, cut-off Hit ratio (H@N), Mean Rank(MR) and Mean Reciprocal Rank (MRR). H@N measures the ratio of correct entities predictions at a top $n$ prediction result cut-off. Following the baselines used in recent literature, we chose $n=1,3,10$. MR evaluates the average rank among all the correct entities. MRR is the average rank inverse rank of the correct entities.

\paragraph{Implementation Details: }
We implemented our models using pytorch\footnote{https://www.pytorch.org} framework and experimented on a server with an Nvidia Titan-1080 GPU. The Adam~\cite{Adam} optimizer was used with the default $\beta_{1}$ and $\beta_{2}$ settings. A learning rate scheduler observing validation loss decrease was used to reduce learning rate by half after patience of 3000. Batch-size was set at 1024.  We did a grid search on the following hyper-parameters: embedding dimension $d\in\{100, 250, 400, 500\}$; learning rate $\eta\in\{3e-4, 1e-4, 3e-5, 1e-5, 3e-6\}$; number of negative samples during training $n_{neg}\in\{128, 256, 512\}$; adversarial negative sampling temperature $\alpha \in\{0.5, 0.75, 1.0, 1.25\}$; loss function margin $\gamma \in \{6, 9, 12, 20, 22, 24, 26\}$.

\subsection{Results and Model analysis}

Empirical results on FB15k and WN18 are reported in Table \ref{tab:1815}. We compared the embedding results of different groups, including $T$, $U(1)$, $GL(1, \mathbb{R})$, $GL(1,\mathbb{C})$, $SO(3)$ and $SU(2)$, which are mainly categorized by the commutativity. As discussed in Sec.~\ref{3.3}, the former four groups have been implicitly applied in existing models. 
For $SO(3)$ and $SU(2)$, we report the result of our own experiments. Results of the other models are taken from their original literature: TransE using group $T$ was proposed in \cite{transe}; RotatE using group $U(1)$ was proposed in \cite{rotate} while TorusE with the same group was proposed in \cite{toruse}; group $GL(1,\mathbb{V})$ was implemented in DisMult~\cite{dismult} with $\mathbb{V}=\mathbb{R}$ and in ComplEx~\cite{complex} with $\mathbb{V}=\mathbb{C}$.

Results on datasets FB15K-237 and WN18RR are demonstrated in Table \ref{tab:237rr} respectively. We remove TorusE from the tables due to the absence of results in the original work, and refer to \cite{nguyen2017novel} for TransE.

In the FB15k dataset, the main hyper-relation is anti-/symmetry and inversion. The dataset has a vast amount of unique entities. Shown in Table \ref{tab:1815}, the RotatE model achieved good performance in this dataset. 
SO3E and SU2E achieved comparable result across the metrics. On the other hand, since inversion relations are removed in FB15k-237, the dominant portion of hyper-relations becomes the composition. We can see RotatE fail on this task due to non-Abelian hyper-relations. Shown in Table \ref{tab:237rr}, the continuous non-Abelian group method SO3E and SU2E outperformed most of the metrics.

In the WN18 dataset, SO3E and SU2E outperformed all the baselines on all metrics shown in Table \ref{tab:1815}. The WN18RR dataset removes the inversion relations from WN18, left only 11 relations and most of them are symmetry patterns. We can see from Table \ref{tab:237rr}, SO3E and SU2E model performed well due to their non-Abelian nature.

Drawn from the experiments, two factors significantly impact the embedding model performance: the embedding dimension, and group attributes (including commutativity and continuity). As theoretically analyzed in Section \ref{theory}, and empirically shown above, continuous non-Abelian groups are more reasonable choices for general tasks. It is important to note that SO3E and SU2E proposed above are exampling models for our group embedding framework, and they use the simplest continuous non-Abelian groups. Much more efforts could be devoted in this direction in the future.

\section{Conclusions}\label{conclusion}

We proved for the first time the emergence of a group definition in the KG representation learning. This proof suggests that relational embeddings should respect the group structure. 
A novel theoretic framework based on group theory was therefore proposed, termed as the \emph{group embedding} of relational KGs. Embedding models designed based on our proposed framework would automatically accommodate all possible hyper-relations, which are building-blocks of the link prediction task. 

From the group-theoretic perspective, we categorize different embedding groups regarding commutativity and the continuity and empirically compared their performance. We also realize that many recent models correspond to embeddings using different groups. Generally speaking, a continuous non-Abelian group embedding should be powerful for a generic KG completion task. We demonstrate this idea by examining two simple exampling models: \textbf{SO3E} and \textbf{SU2E}. With $SO(3)$ and $SU(2)$ as embedding groups, our models showed promising performance in challenging tasks where hyper-relations become crucial.

\section{Discussions}
We close our first exhibition of guided model design by mentioning several potential directions which deserve further discussions.

Restricted to the introduced group theoretic framework, there are several aspects under-explored.
Beside embedding relations as group elements, entity embeddings live in different representation space of the corresponding group. 
And therefore an investigation of group representation theory in entity embedding is highly demanded. We leave this in future works.
On the other hand, although empirical evaluations focus on linear models, it is important to note that the proof of the group structure only relies on the KG task itself. This means our conclusion also works for more general models, including neural-network-based ones. 
Even beyond KG embeddings, the same analysis could be applied to other representation learning where intrinsic relational structures are prominent. 
An implementation of group structures in more general cases would be very interesting.

At the same time, there are in fact some drawbacks of the proposed framework.
The most significant one is that the whole study has been restricted to relations of $1-1$ mappings.
However, $N-1$ relations are very common in real life datasets.
For example, \textbf{isParentOf} is a typical $N-1$ relation where $N=2$.
The required existence of inverse elements prohibits a proper embedding of this type of relations.
To capture this more general scenario, therefore, it is inevitable to discuss algebraic structure where non-invertible elements could be accommodated.
In fact, semi-group structure provides a promising ground, which differs from a group exactly on the existence of inverse elements.
Semi-group based relational embedding therefore deserves a deep exploration.

% \newpage

\setcounter{equation}{0}
\chapter{Spectral Induced Equivariant Network}
% \epigraph{ The best of artists hath no thought to show	\\	
%       which the rough stone in its superfluous shell	\\
%     doth not include; to break the marble spell	\\
%       is all the hand that serves the brain can do.		
% 	}{  {\it-Michelangelo}}

In this chapter, we would provide another instantiation of guided model design.
This time, a more classical area would be discussed for the demonstration: image classification.
From a high level abstraction with respect to guided model design, we would be interested in investigating spatial orders in the signal of images.
More specifically, the information represented by an image would transform in a principled way if the underlying image experiences certain geometric transformation.
This implies a global spatial correlation among pixels at different location.
To explore this correlation and the resulting global constraints on model design, we exploit the power of group representation theory in capturing symmetry characteristics of the signal.
Eventually, a novel deep neural network architecture is proposed based on irreducible representation of space groups.

\section{Background}

To fully specify all features of an image taken in real life, a large number of variables is required: position, pose, shape, color, and so on. Besides, for a given image, there could be a series of potentially meaningful transformations. These transformations bring in extra difficulty for machine learning models to generalize. 
More importantly, even if the same kind of transformation can mean differently on different tasks. For example, most identification tasks are insensitive to simple geometric transformations like translations and rotations, and therefore we say this task holds certain \emph{invariance}~\cite{6522407}; on the other hand, however, when a translated/rotated input mapping to an output with different category, the task is then sensitive to those transformations, and there is no invariance.

Invariance is a special case. 
In general, one would request a network structure capable of capturing the difference brought by transformations, instead of pre-assuming impacts brought by transformations (e.g. invariance). 
More importantly, the design of a general network should be able to capture the change of information~\cite{Lowe03distinctiveimage, NIPS2015_33ceb07b} within all the intermediate representations of hidden layers, up to the output one, at which impacts brought by transformations could be defined by the task itself.

Among all possible transformations, the most intuitive and important ones are geometric transformations, including translations, rotations, inversions, and general affine transformations. 
All possible geometric transformations could form a closed algebraic structure, called \emph{group}. 
And in this case, the above requirement on the network is called \emph{equivariance} in group theory \cite{10.1007/978-3-642-21735-7_6}.

There have been many significant works~\cite{TransformingA, TransformationEB, sohn2012learning} tackling the equivariance property in recent years. 
A series of works done by Cohen et al.\cite{DBLP:conf/icml/CohenW16, DBLP:journals/corr/CohenW16a} already proposed various generalized CNNs by incorporating different symmetry transformations. 
The recent seminal work \cite{DBLP:journals/corr/abs-1811-02017} further provided a generic recipe for constructing real-space equivariant models. 
Besides, continuous-rotational invariant tasks have also been successfully tackled by \cite{DBLP:journals/corr/WorrallGTB16} and \cite{DBLP:journals/corr/GonzalezVKT16} in practice.

Nearly all prior works~\cite{DBLP:journals/corr/abs-1811-02017, DBLP:conf/icml/CohenW16, DBLP:journals/corr/CohenW16a, OyallonM14, 6619007} start from conventional CNN structure~\cite{8308186}, and propose different forms of kernels. Conventional CNN has already kept translation equivariance, therefore, as long as one stays in the parameter space with the same setup, this equivariance property is attained. 
Several works~\cite{DBLP:journals/corr/WorrallGTB16} focus on the rotation and inversion equivariance especially for 2-dimension images. 
And in fact, without translations, the concerned group becomes a point-group, which is much easier to be dealt with.

Our present work treats the problem from a completely different perspective. 
Rather than starting from the conventional CNN structure and then adding more constraints for additional symmetry transformations, we initiate our work from a pure mathematical concern: achieve the equivariance of the full symmetry group $G$ in one-shot, which leads to the analysis using group representation theory~\cite{ping2002group}. 
A natural representation of signals turns out to be hosted by the spectral domain (i.e. frequency space) rather than the real space  (pixel space). 
In spectral domain, we derived general rules for equivariance, a simple implementation of which leads to the spectral induced equivariant network (SieNet).

\section{Equivariance of Symmetry Transformations}

It is well-known that the conventional CNN is equivariant w.r.t the translation~\cite{DBLP:conf/icml/CohenW16}, that is: if the input image is translated by a certain amount (in the unit of pixels), the feature maps created by a convolution layer would also be translated by the same amount, which suggests the information change from the translations has been captured. 
This example is intuitive and easy to be demonstrated. However, to rigorously study equivariance and, at the same time, without being tautology, one needs to firstly sharply define the equivariance in a practical image processing task.

In this section, we would firstly provide a proper definition of equivariance by considering a signal in the real (pixel) space. 
The real space signals are more intuitive and straightforward to be comprehend, and we would use real space signals to argue that only the trivial representation of the quotient group $G/T$ is necessary for a pure geometric concern. 
Based on these analysis, we then use a simple example with only translation group to introduce group representation theory for full $G$-equivariance construction, which naturally leads to a study in the spectral (frequency) space.

\subsection{Definition of Equivariance}
We firstly provide a formal definition of equivariance. Considering the following two feature maps produced by a CNN model~\cite{8308186}:
\begin{equation}
    f(\vec{r})\in\mathcal{F}, \qquad\quad \hat{f}(\vec{r})\in\hat{\mathcal{F}},
\end{equation}
in two adjacent layers, $l$ and $l+1$, that are connected by certain operation $F(\cdot)$, which in general can be a convolution, a nonlinearity, or a pooling without sub-sampling:
\begin{equation}
F[f](\vec{r}) =: \hat{f}(\vec{r}),
\end{equation}
where, for simplicity, we have assumed the size of feature maps does not change through the operation $F$. If we specify \emph{in advance} the action of certain 2-dimensional Wallpaper group $G$ on both the function space $\mathcal{F}$ and $\hat{\mathcal{F}}$ as the following automorphism:
\begin{align}
\mathcal{G}&: \mathcal{F}\mapsto\mathcal{F} \nonumber \\
\hat{\mathcal{G}}&: \hat{\mathcal{F}}\mapsto\hat{\mathcal{F}},
\end{align}
or, more specifically, we could denote as:
\begin{align}
g&: f \mapsto g\circ f \qquad \forall g\in G, f\in \mathcal{F} \nonumber \\
g&: \hat{f} \mapsto g\circ \hat{f} \qquad \forall g\in G, \hat{f}\in \hat{\mathcal{F}},
\end{align}
where $g\circ f$ (or $\hat{f}$) are also elements of $\mathcal{F}$ (or $\hat{\mathcal{F}}$).
\begin{Def}
The \emph{\textbf{$G$-equivariance condition} of $F$} is:
\begin{equation}\label{eq_def}
g\circ F[f](\vec{r}) = F[g\circ f](\vec{r}), \qquad \forall g\in G, f\in \mathcal{F}.
\end{equation}
\end{Def}

We would like to emphasize that: equivariance is a condition for the operation $F(\cdot)$ required by the predefined group actions $\mathcal{G}$ and $\hat{\mathcal{G}}$. That is to say: only when both $\mathcal{G}$ and $\hat{\mathcal{G}}$ are specified in advance, one could derive a meaningful equivariance condition. 
%In most cases $\mathcal{G}$ is defined properly by the problem itself (\egthe geometric transformation of an image). On the other hand, if one firstly writes down a form for $F(\cdot)$ and then tries to prove a so-called equivariance, it would be, actually, as demonstrated below, a tautology which is constantly true by design.

\subsection{Trivial Equivariance versus Correct Formulation}

\subsubsection{Trivial equivariance}
In this section, we would firstly demonstrate an "equivariance" achieved trivially by fixing the operation $F(\cdot)$ without a definition of $\hat{\mathcal{G}}$ in advance, and then would provide the proper way to study equivariance.

Now without any concern about equivariance, given the form of operation $F(\cdot)$, for any feature map $f\in\mathcal{F}$ in the previous layer, we \emph{\textbf{define}} the group action $\hat{\mathcal{G}}$ on the operated feature map $\hat{f}:=F[f]$ in the next layer as:
\begin{align}
g\circ \hat{f} := F[g\circ f].
\end{align}
This provides a definition of $\hat{\mathcal{G}}$ for any $\hat{f}$ obtained from $F(\cdot)$. The above definition is equivalent to:
\begin{align}\label{trivial_eq}
g\circ F[f](\vec{r}) := F[g\circ f](\vec{r}),
\end{align}
which looks nearly the same as the condition stated in \eqref{eq_def}, however it is a defining statement rather than an independent condition as before. In other words, given any form of $F(\cdot)$, as long as we intentionally define that the group action in the next layer is generated from the one in the previous layer, then equivariance is achieved trivially by definition.

This demonstration is not completely rigorous: in the case where $F(\cdot)$ is a many-to-one map, \eqref{trivial_eq} might not be a legitimate definition of group action. However, the above analysis indeed describes the cases of some existing works. For example, in the case of Harmonic Networks, the harmonic function used to construct a convolution kernel is chosen before analyzing the group action in the next layer; therefore, although mathematically the claimed equivariance seems to be correct, it is a result of definition derived from the designed kernel form.

\subsubsection{Correct formulation of equivariance}
As stated before, the proper way to study equivariance is to firstly identify the two group actions $\mathcal{G}$ and $\hat{\mathcal{G}}$. In conventional CNNs, the action of translations, before and after a convolution, is always defined as translations of pixels, which is intuitive and makes feature maps in higher layers more interpretable: pixels still correspond to spatial locations.
Usually, $\mathcal{G}$ on the input images (e.g. \textit{RGB} features) is well-defined, and we would like to find an appropriate $\hat{\mathcal{G}}$ for higher layers. 
%In the case of steerable CNNs, many possible $\hat{\mathcal{G}}$'s are proposed, while an assessment of different $\hat{\mathcal{G}}$'s is absent.

A general signal of 2D image problems consists of multiple feature maps and therefore has (2+1)-dimensions, and we would call them: 2 \textit{spatial dimensions} and 1 \textit{on-site dimension}. In most studies, geometric transformations on 2D images can be described by permutations of pixels (becoming more complicated for a generic affine transformation). Therefore, to keep the geometric meaning, we would consider that \textbf{pixels in higher layers still represent real space locations/coordinates}. A symmetry transformation would transform higher layer features in a way as it acting on the real images, i.e. by permuting pixels:
\begin{align}\label{pix_permut}
\hat{f}(\mathbf{r}) \quad\longrightarrow\quad \hat{f}(g^{-1}\mathbf{r}),
\end{align}
where $g^{-1}\mathbf{r}$ represents another coordinate (pixel) in the 2D space.

Besides pixel permutations, previous works also considered certain global on-site action by incorporating nontrivial irreducible representations of the quotient group $G/T$, where $T$ is the translation group. With these two types of transformation, one could define a general group action $\mathcal{G}$($\hat{\mathcal{G}}$) on the function space $\mathcal{F}$($\hat{\mathcal{F}}$) as:
\begin{align}\label{general_action}
g: f_i(\mathbf{r}) &\mapsto [g\circ f]_i(\mathbf{r}) := \sum_{i'} W_{ii'}^{g} f_{i'}(g^{-1}\mathbf{r}) \nonumber \\
g: \hat{f}_i(\hat{\mathbf{r}}) &\mapsto [g\circ \hat{f}]_i(\hat{\mathbf{r}}) := \sum_{i'} \hat{W}_{ii'}^{g} \hat{f}_{i'}(g^{-1}\hat{\mathbf{r}}),
\end{align}
where $\mathbf{W}$ ($\hat{\mathbf{W}}$) are globally defined matrices acting on the on-site dimension, which form a representation of the quotient group $G/T$. With this predefined group action for each layer, we obtain the following general equivariance condition for a linear operation $\mathbf{\kappa}$:
\begin{The}\label{theo_1}
\textbf{Real Space Theorem:} Consider two adjacent layers connected by the following linear operation:
\begin{align}\label{kappa}
\hat{f}_{i}(\hat{\mathbf{r}})  = \sum_{\mathbf{r},j}\kappa_{ij}(\hat{\mathbf{r}}, \mathbf{r})f_j(\mathbf{r}),
\end{align}
and the group actions of certain symmetry group $G$ on these two layers are defined as in \eqref{general_action}. Then the operation $\mathbf{\kappa}$ would be $G$-equivariant if it is \textit{symmetric} in the following sense:
\begin{equation}\label{sym_kern0}
\kappa_{ij}(\hat{\mathbf{r}}, \mathbf{r}) = \sum_{i'j'}\hat{W}^{g}_{ii'}\kappa_{i'j'}(g^{-1}\hat{\mathbf{r}}, g^{-1}\mathbf{r})[W^{g}]^{-1}_{j'j},
\end{equation}
which should hold for an arbitrary element $g\in G$.
\end{The}
\bproof
Signals represented in the real 2D space transform under a geometric transformation as in Eq.\ref{general_action}, where $g$ is any group element in $G$. Equivariance requires the following equation holds:
\begin{align}
[g\circ \hat{f}]_i(\hat{\mathbf{r}})=\sum_{\mathbf{r},j}\kappa_{ij}(\hat{\mathbf{r}}, \mathbf{r})[g\circ f]_j(\mathbf{r}),
\end{align}
which is equivalent to:
\begin{align}\label{form1}
\sum_{i'} \hat{W}_{ii'}^{g} \hat{f}_{i'}(g^{-1}\hat{\mathbf{r}})
=\sum_{\mathbf{r},j,j'}\kappa_{ij}(\hat{\mathbf{r}}, \mathbf{r})W_{jj'}^{g} f_{j'}(g^{-1}\mathbf{r}).
\end{align}
Since $g^{-1}\mathbf{r}$ (or $g^{-1}\mathbf{\hat{r}}$) is still a coordinate in the real 2D space, we could obtain from \eqref{kappa} that:
\begin{align}
\hat{f}_{i}(g^{-1}\hat{\mathbf{r}})  = \sum_{g^{-1}\mathbf{r},j}\kappa_{ij}(g^{-1}\hat{\mathbf{r}}, g^{-1}\mathbf{r})f_j(g^{-1}\mathbf{r}).
\end{align}
Multiply both side by $\hat{W}^{g}$:
\begin{align}
\sum_{i'}\hat{W}_{ii'}^{g}\hat{f}_{i}(g^{-1}\hat{\mathbf{r}})  = \sum_{g^{-1}\mathbf{r}}\sum_{i',j}\hat{W}_{ii'}^{g}\kappa_{i'j}(g^{-1}\hat{\mathbf{r}}, g^{-1}\mathbf{r})f_j(g^{-1}\mathbf{r}),
\end{align}
and then insert the identity matrix: $[W^{g}]^{-1}W^{g}=\mathbb{I}$, we obtain:
\begin{align}\label{form2}
\sum_{i'}\hat{W}_{ii'}^{g}\hat{f}_{i}(g^{-1}\hat{\mathbf{r}})  = \sum_{\mathbf{r},j,j'}\bigg[\sum_{i',j"}\hat{W}_{ii'}^{g}\kappa_{i'j"}(g^{-1}\hat{\mathbf{r}}, g^{-1}\mathbf{r})
[W^{g}]_{j"j}^{-1}\bigg]W^{g}_{jj'}
f_{j'}(g^{-1}\mathbf{r}),
\end{align}
where we have replaced $\sum_{\mathbf{r}}$ with $\sum_{g^{-1}\mathbf{r}}$, since both summations go over all coordinates in the 2D space (pixel-grid). Compare \eqref{form1} with \eqref{form2}, we have:
\begin{align}
\sum_{\mathbf{r},j,j'}\kappa_{ij}(\hat{\mathbf{r}}, \mathbf{r})W_{jj'}^{g} f_{j'}(g^{-1}\mathbf{r}) = 
\sum_{\mathbf{r},j,j'}\bigg[\sum_{i',j"}\hat{W}_{ii'}^{g}\kappa_{i'j"}(g^{-1}\hat{\mathbf{r}}, g^{-1}\mathbf{r})
[W^{g}]_{j"j}^{-1}\bigg]W^{g}_{jj'}f_{j'}(g^{-1}\mathbf{r}),
\end{align}
which should hold for arbitrary function $f\in\mathcal{F}$ and element $g\in G$. Therefore we obtain a simple general solution:
\begin{align}\label{k_form1}
\kappa_{ij}(\hat{\mathbf{r}}, \mathbf{r})  = 
\sum_{i',j'}\hat{W}_{ii'}^{g}\kappa_{i'j'}(g^{-1}\hat{\mathbf{r}}, g^{-1}\mathbf{r})
[W^{g}]_{j'j}^{-1} \qquad \forall g, \mathbf{r}, \hat{\mathbf{r}},
\end{align}
which is precisely the equation in \textbf{Theorem \ref{theo_1}. Q.E.D.}
\eproof

We already mentioned that both $W^g$ and $\hat{W}^g$ form reps of $G$, one can then show that the following matrices also form a rep of $G$:
\begin{align}
\tilde{W}^g = [W^{(g^{-1})}]^T,
\end{align}
where $T$ means the transpose, by noticing:
\begin{align}
\tilde{W}^{g_1}\cdot\tilde{W}^{g_2} &= [W^{(g_1^{-1})}]^T\cdot[W^{(g_2^{-1})}]^T \nonumber \\
&= [W^{(g_2^{-1})}\cdot W^{(g_1^{-1})}]^T \nonumber \\
&= [W^{(g_2^{-1}g_1^{-1})}]^T \nonumber \\
&= [W^{(g_1g_2)^{-1}}]^T \nonumber \\
&= \tilde{W}^{g_1g_2}.
\end{align}
And therefore we could further rewrite \eqref{k_form1} as:
\begin{align}\label{k_form2}
\kappa_{ij}(\hat{\mathbf{r}}, \mathbf{r}) 
&= \sum_{i',j'}\hat{W}_{ii'}^{g}\tilde{W}^{g}_{jj'}\cdot\kappa_{i'j'}(g^{-1}\hat{\mathbf{r}}, g^{-1}\mathbf{r}) \nonumber \\
&= \sum_{i',j'}\big[\hat{W}^{g}\otimes\tilde{W}^{g}\big]^{i'j'}_{ij}\cdot\kappa_{i'j'}(g^{-1}\hat{\mathbf{r}}, g^{-1}\mathbf{r}).
\end{align}
One immediately realizes that $\hat{W}^{g}\otimes\tilde{W}^{g}$ forms a tensor representation of group $G$.
Now if we regard $\kappa$ as rank-2 tensor functions on $(\mathbf{\hat{r}}, \mathbf{r})\in \mathbb{R}^4$, which form the basis of the tensor representation, the above equation then rigorously has the following meaning: $\kappa$ lives in the symmetric subspace of the tensor-function space.

\subsubsection{Role of on-site representations}

We now further discuss the role of on-site rep, i.e. $\mathbf{W}^g$ (or $\mathbf{\hat{W}}^g$). 
In general, a globally defined on-site matrix would mix feature maps with a set of linear superpositions. Therefore, in the case with a nontrivial on-site rep, the group action of $g$ on a function would:
\begin{enumerate}
\item
firstly transfer the function value from another coordinate at $g^{-1}\mathbf{r}$;
\item
then, with the non-identity matrix $W^g$, would mix feature maps along the on-site dimension.
\end{enumerate}
Apparently, the first step corresponds to a geometric transformation in the real space, e.g. reshaping or moving the image, which has an intuitive meaning. While it is not clear what role the second step plays, at least in the geometric context. 

The above discussion is related in one's explanation of pixel value. In our current set-up, as mentioned in the main part, we regard pixels as real 2D space coordinates, which is the most intuitive way and makes possible to visualize and interpret higher layer feature maps. This is not the only choice: in the work of Group Equivariant CNN~\cite{DBLP:conf/icml/CohenW16}, authors regard pixels as group elements in the group space, in which case mathematical derivations become simpler at the expense of losing an intuitive interpretation of signals.

On the other hand, however, if one takes a consideration more general than the geometric context, the feature map mixing might be capable of describing other interesting information. For example, in the case of 3D pose/angle-change on a projected 2D image, as one rotate the 2D projection, there might be an associating "shadow" change, which might be achieved by a feature maps mixing.

Therefore, to study only 2D geometric transformations, we believe a nontrivial on-site transformation $\mathbf{W}$ ($\hat{\mathbf{W}}$) is unnecessary, since \emph{a 2D geometric symmetry transformations only acting on spatial dimensions, and if we deem pixels in higher layers still represent real space coordinates, then a geometric symmetry transformation should only act spatially, not on-site}. 

Due to the above reason, we would use only trivial on-site representation (all $\mathbf{W}$'s are identity matrices) in our work, and the equivariance condition for a linear operation $\kappa$ then becomes:
\begin{align}\label{sym_kern}
\kappa_{ij}(\hat{\mathbf{r}}, \mathbf{r}) = \kappa_{ij}(g\hat{\mathbf{r}}, g\mathbf{r}) \qquad \quad \forall g\in G,
\end{align}
which implies a kernel symmetric in the real space. Special solutions of the above equation yield to conventional CNNs, Harmonic Networks, and Steerable CNNs, as introduced below.

\subsection{Special solution and related work}
The above equation \eqref{k_form1} constraining the form of $\kappa$ to be symmetric as a tensor-function. Now we would talk about some special solution forms satisfying the general constraint.

To compare with related works, we would like to simplify the discussion on $\mathbf{W}^g$ and $\hat{\mathbf{W}}^g$, by requiring a trivial on-site rep for the translation group $T$:
\begin{align}\label{trans_trivial}
\mathbf{W}^g = \hat{\mathbf{W}}^g = \mathbf{1}, \qquad \forall g \in T.
\end{align}
Although the most general consideration indeed allows a nontrivial on-site $T$-rep, on the one hand, it has never been shown in previous works and hence is irrelevant for our comparison; and, on the other hand, as we discussed in the main text and above, a non-trivial on-site rep is unnecessary for a pure 2D geometric concern.

\begin{Exp}
\textbf{Conventional CNN with translation group}

As a starting point, we consider all translation transformations: $g\in T$. As required in \eqref{k_form1} and the above:
\begin{align}
\kappa(\hat{\mathbf{r}}, \mathbf{r})  =  \kappa(\hat{\mathbf{r}}-\hat{\tau}, \mathbf{r}-\hat{\tau}), \qquad \forall\tau,
\end{align}
which means $\kappa$ is invariant if both coordinates are shifted by the same amount. The only solution would be:
\begin{align}
\kappa(\hat{\mathbf{r}}, \mathbf{r}) \equiv \kappa(\hat{\mathbf{r}}-\mathbf{r}),
\end{align}
i.e. $\kappa$ only depends on the difference of two coordinates in the adjacent layers. This is precisely the conventional CNN kernel, if one further requires that the function value vanishes outside a finite region, corresponding to the kernel size in literature.
\end{Exp}

\begin{Exp}\textbf{Harmonic Networks with $SE(2)$ group}

Now we consider the case of $G=SE(2)$, where, in addition to the translation sub-group $T$, we further consider a Lie group $SO(2)$ describing continuous rotations in 2D real space. From the discussion above, we start with the operation $\kappa(\hat{\mathbf{r}}-\mathbf{r})$, which is already $T$-equivariant. Now to be rotational equivariant, we further have the following condition:
\begin{align}
\kappa_{ij}(\hat{\mathbf{r}}-\mathbf{r})  
&= \sum_{i',j'}\hat{W}_{ii'}^{h}\cdot\kappa_{i'j'}(h^{-1}\hat{\mathbf{r}}-h^{-1}\mathbf{r})
\cdot[W^{h}]_{j'j}^{-1} \nonumber \\
&= \sum_{i',j'}\hat{W}_{ii'}^{h}\cdot\kappa_{i'j'}\big[h^{-1}(\hat{\mathbf{r}}-\mathbf{r})\big]
\cdot[W^{h}]_{j'j}^{-1} \qquad \forall h\in SO(2).
\end{align}
Now the matrices $\mathbf{W}^h$ (or $\mathbf{\hat{W}}^h$) form an on-site rep of $SO(2)$. From representation theory, we could always decompose it into irreps of $SO(2)$, which are all 1-dimensional, usually labeled by an integer $m$. The matrix (actually a scalar) of a rotation by an angel $\theta$, i.e. $h_{\theta}\in SO(2)$, in the irrep-$m$ has the form:
\begin{align}
W^{h_{\theta}}_m = e^{im\theta}.
\end{align}
Since all irreps are 1-dim, one could always transform all $\mathbf{W}^h$ (or $\mathbf{\hat{W}}^h$) into diagonal matrices, with each diagonal element written in the above form, and the condition becomes:
\begin{align}
\kappa_{ij}(\hat{\mathbf{r}}-\mathbf{r})  
&= \kappa_{ij}\big[h_{\theta}^{-1}(\hat{\mathbf{r}}-\mathbf{r})\big] \cdot e^{i(\hat{m}_i-m_j)\theta}, \qquad \theta\in [0, 2\pi],
\end{align}
where $\hat{m}_i$ (or $m_j$) labels the irrep assigned to the feature map $i$ (or $j$). Therefore, if we label: $m_{ij} = \hat{m}_i-m_j$, the above form becomes precisely the Harmonic Networks. 

As stated in the main article, we emphasize the difference here: the original Harmonic Network in chooses $m$ value directly as a hyperparameter, without specifying $\hat{m}_i$ and $m_j$ first; while our discussion above do it inversely: firstly find proper $\hat{m}_i$ and $m_j$, and then determine $m$ as their difference. Although it seems the two above logic lead to the same result, they are conceptually different in the discussion of equivariance: since to assign directly an irrep for $\kappa$ without knowing the proper irrep of the feature maps does not really produce an equivariance condition but is more or less a tautology, and mathematically the meaning of irrep on $\kappa$ is hard to be comprehended. 

Actually, again as discussed already, with a pure geometric concern, non-trivial on-site irreps (hence non-identity $\mathbf{W}$) may not be necessary at all; while they may play a role in more complicated tasks.
\end{Exp}

\begin{Exp}\textbf{Steerable CNN with general Wallpaper group}

Now we discuss a general 2D Wallpaper group $G$, with a 2D point group $H$ as a subgroup.
With a constraint in \eqref{trans_trivial}, we have the following general condition:
\begin{align}
\kappa_{ij}(\hat{\mathbf{r}}-\mathbf{r})  
&= \sum_{i',j'}\hat{W}_{ii'}^{h}\cdot\kappa_{i'j'}\big[h^{-1}(\hat{\mathbf{r}}-\mathbf{r})\big]
\cdot[W^{h}]_{j'j}^{-1} \qquad \forall h\in H.
\end{align}
$\mathbf{W}^h$ (and $\mathbf{\hat{W}}^h$) is a globally defined on-site rep for $H$, which could always be decomposed into irreps of $H$ and hence written in a block-diagonal form. One could choose in advance irreps contained in both $\mathbf{W}$ and $\mathbf{\hat{W}}$, and then the above condition corresponds to the Steerable CNN structure.

Another special case is when one chooses explicitly both $\mathbf{W}$ and $\mathbf{\hat{W}}$ as the regular representation of $H$, then the result would corresponds to Group Equivariant CNNs.
\end{Exp}

\subsection{Equivariant Structures via Group Representation Theory}
%The above discussion focuses on real 2D-space, which results into a relatively complicated constraint on the kernel form as in \eqref{sym_kern} due to a nontrivial group action in the function space $\mathcal{F}$.
%In fact, a much simpler form could be obtained by decomposing the whole function space $\mathcal{F}$ into subspaces corresponding to irreducible representations (irreps) of the concerned group $G$. The form of $\kappa$ becomes simple because: 
%\begin{enumerate}
%\item
%different subspaces do not get mixed under group transformations;
%\item
%all irreps of a given group $G$ can be explicitly fixed beforehand.
%\end{enumerate}

The discussion above has specified that a 2D geometric transformations transform a real space feature map in $\mathcal{F}$ by a pixel permutations (with necessary interpolations). From now on, we would only use this type of transformations and the associated function space $\mathcal{F}$, and would transfer from real (pixel) space to frequency space.

As mentioned above, most proceeding works use real space signals and discuss irreps of the quotient point group $G/T$, which raises two problems:
\begin{enumerate}
\item 
Translations and point-group transformations (rotation, mirrors, inversions, and so on) are discussed separately, bringing in the extra difficulty in solving and comprehending the interplay between these two types of transformations;
\item 
As we argued above, for a pure geometric consideration, only the trivial irrep of $G/T$ with identity matrices $\mathbf{W}$ is explainable, and other irreps usually lack interpretations.
\end{enumerate}
It is actually more convenient to study equivariance through irreps of the whole group $G$ itself rather than the subgroup $G/T$. On the one hand, the two above problems get solved directly, and, on the other hand, a discussion of $G$-irrep decomposition usually provides more general forms for equivariant operations. The idea of directly considering the whole group $G$ has actually been implied in the case of 3D CNNs and Clebsch-Gordan Nets, where $G=SO(3)$ whose irreps are well-known as spherical harmonics. However, to the best of what we know, in the case where translation exists, there has not been a discussion on implementing irreps of the whole group $G$. 

In the following discussion, we use the conventional CNN as an example with $G=T$, to show that a $G$-irrep analysis would provide more general forms of equivariant operations, which are also easier to implement.

\begin{Exp}\textbf{Pure translation group}

In this case with $G=T$, irreps of $G$ actually correspond to different spectral components $d(\mathbf{k})$ obtained from the Fourier transformation:
\begin{align}
d(\mathbf{k}) &= \frac{1}{N}\sum_{x,y=1}^{N}e^{-i[k_xx + k_yy]}\cdot f(\mathbf{r}), \nonumber \\
\text{with }\:\: k_{\alpha}&:=\frac{2\pi n_{\alpha}}{N},\;\: n_{\alpha}\in[-N/2,\; N/2],
\end{align}
which are all 1-dimensional: the action of any group elements, $T_x^{\tau_x}T_y^{\tau_y}\in T$, is simply:
\begin{equation}
d(\mathbf{k}) \;\longrightarrow\; e^{-i[k_x\tau_x + k_y\tau_y]}\cdot d(\mathbf{k}),
\end{equation}
i.e. multiplying a $\mathbf{k}$-dependent phase factor. Therefore, \emph{a $T$-equivariant operation should not depend on the phase of $d(\mathbf{k})$}. The \textbf{simplest solution} could be a point-wise multiplication, which do not mix different irreps at all:
\begin{equation}\label{point_prod}
d(\mathbf{k}) \;\longrightarrow\; \lambda(\mathbf{k})\cdot d(\mathbf{k}).
\end{equation}
And this actually contains conventional CNNs as a subset: any convolution (with finite-size kernels) in the real space corresponds to a point-wise multiplication in the frequency domain. 

More importantly, however, the above analysis actually implies other more complicated $T$-equivariant candidates: \textbf{\emph{any operations that do not rely on the phase of $d(\mathbf{k})$}}. With this in mind, one could also try to compose equivariant nonlinear operations and "pooling" operations, such that the whole network structure would become rigorously equivariant.
\end{Exp}

The above analysis starts with a completely mathematical concern: to achieve full $T$-equivariance, and then provides general equivariant solutions which contain conventional CNNs as a subset. The spectral method therefore enriches the equivariance-allowed solution space. 

In practice, $G=T$ is the simplest case. In the next section, we would discuss the general theory of $G$-irrep decomposition and introduce the proposed model.

%The above analysis and example demonstrate the advantage of the $G$-irrep decomposition. Now, instead of only considering the pure translation group $T$, we would like to decompose $\mathcal{F}$ into irreps of a general 2D Wallpaper group $G$. It turns out that frequency domain is again important in the study. In the next section, we discuss the general theory of $G$-irrep decomposition, and derive more complicated equivariant operations.

\section{Frequency Space for 2D Equivariance}
In this section, we firstly introduce the general result of $G$-irrep decomposition for 2D Wallpaper groups. 
By studying the whole group directly, all possible symmetry transformations are considered systematically within one shot. 

% After the general discussion, we construct part by part a full-symmetry equivariant model. 

\subsection{Group action in frequency space}
As we would use spectral representation later (i.e. signals in the frequency space), it is necessary to clarify the group action on different frequency branches $d(\mathbf{k})$. Recall the definition of $d(\mathbf{k})$:
\begin{align}
d(\mathbf{k}) = \frac{1}{\sqrt{N^2}}\sum_{x,y=1}^{N} e^{-i[k_xx + k_yy]}\cdot f(x,y).
\end{align}
The group action on $d(\mathbf{k})$ is then determined by the group action defined on $f(\mathbf{r})$:
\begin{align}
g: f(\mathbf{r}) \mapsto f(g^{-1}\mathbf{r}), 
\end{align}
which, combined with the above definition of $d(\mathbf{k})$, implies:
\begin{align}
g: d(\mathbf{k}) \mapsto [g\circ d](\mathbf{k}) = &\frac{1}{N}\sum_{\mathbf{r}} e^{-i\mathbf{k}\cdot\mathbf{r}}\cdot f(g^{-1}\mathbf{r}) \nonumber \\
= &\frac{1}{N}\sum_{\mathbf{r}} e^{-i\mathbf{k}\cdot [g\mathbf{r}]}\cdot f(\mathbf{r}).
\end{align}
In the case of translations, i.e. $g= (\tau_x, \tau_y)\in T$:
\begin{align}
[g\circ d](\mathbf{k}) &= \frac{1}{N}\sum_{\mathbf{r}} e^{-i\mathbf{k}\cdot \mathbf{r}} \cdot e^{-i[k_x\tau_x + k_y\tau_y]}\cdot f(\mathbf{r}) \nonumber \\
&= e^{-i[k_x\tau_x + k_y\tau_y]} \cdot d(\mathbf{k}),
\end{align}
which is exactly the transformation rule of Fourier components: point-wise multiplication with a $\mathbf{k}$-dependent phase factor, corresponding the the 1-dimensional nature of $T$-irreps.

In the case of other point-group (or continuous Lie group) elements including rotation, mirror, and inversion, i.e. $g\in H$, since they all preserve the origin $\mathbf{0}$ and are linear unitary operations, the following equation holds:
\begin{align}
\mathbf{k}\cdot [g\mathbf{r}] = [g^{-1}\mathbf{k}]\cdot \mathbf{r}.
\end{align}
One could check each case and verify the above relation for all discrete point groups or continuous orthogonal Lie group $O(2)$. (General affine transformation with non-unitary scaling does not hold, in which case, actually, $\mathbf{k}$ may not be well-defined.)

\subsection{Induced Irreducible Representations of Wallpaper Groups}
The correct way to construct all irreps for a Wallpaper group is the so-called induced representation method. More specifically, irreps of a larger group could be induced from irreps of its subgroups. A general Wallpaper group includes at least two nontrivial subgroups: a point group $H$, and a translation group $T$ which is a normal subgroup. In general we could write as: $G = T\rtimes H$. In this work, we construct irreps of G via an induction from irreps of $T$, which, as stated above, are simply Fourier components in the spectral domain, with each vector $\mathbf{k}=(k_x, k_y)$ labeling a one-dimensional irrep of $T$.

There are three advantage to start from $T$ rather than $H$: (i) $T$ is a normal subgroup, which simplifies the interplay with other elements in $H$; (ii) all irreps of $T$ is one-dimensional, resulting into a simpler representation matrix form; (iii) the discussion can be easily generalized to continuous symmetry groups, e.g. $E(2)$ and $SE(2)$, which also contain $T$ as a normal subgroup. Now we state the general recipe to induce irreps of $G$ from the spectral domain:
\begin{Def} \textbf{Spectral Induction for $G$-irreps}

Take spectral domain (irreps of $T$) as a group action space of $H$, by considering each $\mathbf{q}$ transform as a 2-dim vector: $\mathbf{q}\rightarrow h\mathbf{q}, \; \forall h\in H$. Now, consider the group action of $H$ on any $\mathbf{q}$, find the stablizer $H_0^{\mathbf{q}}$, and the orbit $O_{H}^{\mathbf{q}}$. Then we could construct basis functions of a $G$-irrep as: $d_{\mathbf{k}, i}$, which are in general labeled by two indexes: $\mathbf{k}$ labels a frequency vector in the orbit $O_{H}^{\mathbf{q}}$, and $i$ labels the basis functions of certain $H_0^{\mathbf{q}}$-irrep. 
\end{Def}

Following the above definition, one could construct basis functions for all $G$-irreps. Once basis functions are known, representation matrices of $G$ are also determined. In addition, as discussed in the last section, since a trivial representation of $H$ (hence $H_0^{\mathbf{k}}$) is adequate for a geometric concern, we could omit the index $i$, and only use $d_{\mathbf{k}}\equiv d(\mathbf{k})$ to label basis functions, which goes back to the usual spectral representation.

In practice, we mainly focus on transformation rules of different $G$-irreps, which is important for deriving the equivariance constraint. A general element $g$ in $G$ can be written as $g \equiv \big\{h_g|T_x^{\tau_x}T_y^{\tau_y}\big\}$, where $h_g$ is an element in point group $H$.
\begin{align}\label{act_spectr}
\big\{h|T_x^{\tau_x}T_y^{\tau_y}\big\}:\, d(\mathbf{k}) \;\longrightarrow\; e^{-i[k'_x\tau_x + k'_y\tau_y]}\cdot d(\mathbf{k}'),
\end{align}
where $\mathbf{k}'=h_g^{-1}\mathbf{k}$. Hence a general group action on the spectral signal would transfer the frequency vector into another one, and then multiply a $\mathbf{k}$-dependent phase factor. With this in mind, we know a general equivariant operation $F(\cdot)$ should respect the following two constraints:
\begin{enumerate}
\item The form of $F(\cdot)$ should not depend on the phase of signals;
\item If $F(\cdot)$ mix $d(\mathbf{k})$ with other Fourier components at frequency $\mathbf{q}[\mathbf{k}]$, which in general should depend on $\mathbf{k}$, then the following condition should hold: 
\begin{equation}\label{mixed_freq}
h\circ \mathbf{q}[\mathbf{k}] = \mathbf{q}[h\circ \mathbf{k}], \qquad \forall h\in H.
\end{equation}
\end{enumerate}
Given any group $G$, the above results are rigorous. For continuous groups, however, such as $SE(2)$ whose subgroup is $H=SO(2)$, it is impossible to construct a rigorously equivariant network structure. The reason is actually due to the pixel-grid nature of signals, where a continuous rotation cannot be rigorously defined. A trade-off between the mathematical rigorousness and the concerned symmetry has to be made.

\subsection{A heuristic explanation of induced representation method}
The induced representation method, used in the main article to construct irreps and then leads to irrep decomposition of the whole group $G$, is the core of our theoretical analysis. Some abstract mathematical concepts are introduced to make the method rigorous. Here, we use some simple examples to help readers understand this approach, by plotting the group action in spectral domain, i.e. the frequency space (or $k$-space).

\subsubsection{Examples of group actions in spectral domain}
We would first use examples to clarify the concepts of group orbits and stablizer generated from group actions. Since the group action of $T$ only modifies the phase of $d(\mathbf{k})$, we would focus on the action of the subgroup $H$. With these helpful concepts, we would then explain how to achieve equivariance using induced representation, which are actually just $H$-orbits in the spectral domain.

\begin{Exp}
\textbf{Group action and orbits with $G= P_4$}

In this case, the point group is $H=C_4$. The $C_4$ rotation (here we abuse the name for the group generator the same as group's name), which is the only generator of the group, would transform a general $\mathbf{k}$-vector in the following way as shown in Figure~\ref{c4rot}, like an ordinary 2D real space vector.

\begin{figure}[h!]
\centering
\begin{subfigure}{.5\textwidth}
	\centering
	\includegraphics[width = 5cm]{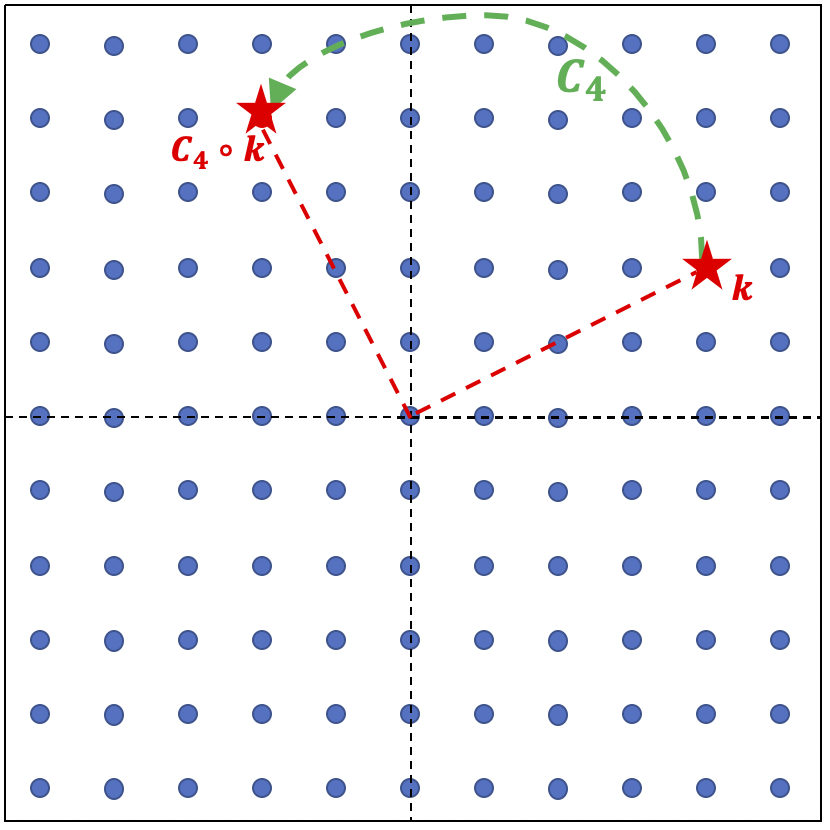}
	\caption{$C_4$ rotation in $k$-space.}
	\label{c4rot}
\end{subfigure}%
\begin{subfigure}{.5\textwidth}
	\centering
	\includegraphics[width = 5cm]{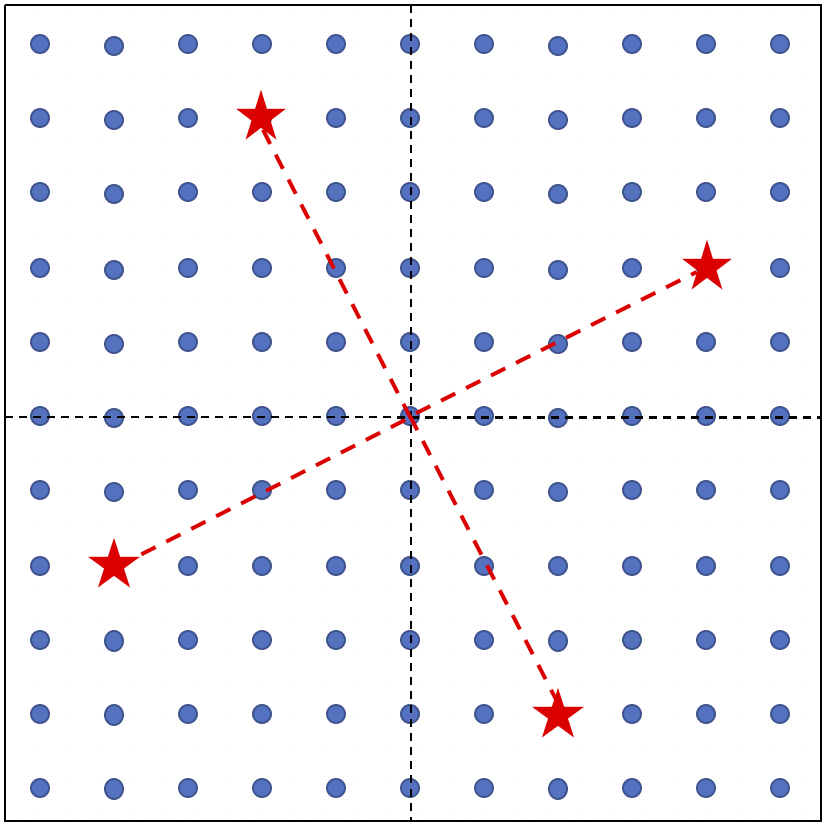}
	\caption{A typical $C_4$ orbit with 4 $\mathbf{k}$'s.}
	\label{c4orb}
\end{subfigure}
\caption{$C_4$ group demonstration}
\end{figure}

There are in total 4 elements in the group-$C_4$: $\{E, C_4, C_4^2, C_4^3\}$. Acting all elements on any vector $\mathbf{k}$ would produce a $H$-orbit: $O_{H}^{\mathbf{k}}$, which usually contains 4 different frequencies, except for special cases (such as the origin). This is shown in Figure~\ref{c4orb}, where the 4 red stars are frequencies contained in the orbit.
\end{Exp}

\begin{Exp}
\textbf{Orbits and stablizers with $G= P_{4m}$}

In this case, the point group is $D_4$. There is one more generator than the $C_4$ case: a mirror operation $\sigma$ along certain axis (for example we could choose $\sigma_x$, a flipping around $x$-axis as the generator), and hence in total 8 group elements. Most orbits of $D_4$ therefore includes 8 frequencies as shown in Figure~\ref{d4orb1}.
However, there are special cases, where fewer frequencies contained in the same orbit, as shown in Figure~\ref{d4orb2}: red (or yellow) stars represents a special orbit consisting of frequencies with high-symmetry. For these high-symmetry frequencies, there is a nontrivial stablizer $H_{0}^{\mathbf{k}}$ associated with each $\mathbf{k}$, which in this example is isomorphic to group $Z_2 = \{E, \sigma_{\mathbf{k}}\}$, where $\sigma_{\mathbf{k}}$ is the axis containing $\mathbf{k}$ itself.

\begin{figure}[h!]
\centering
\begin{subfigure}{.5\textwidth}
	\centering
	\includegraphics[width = 5cm]{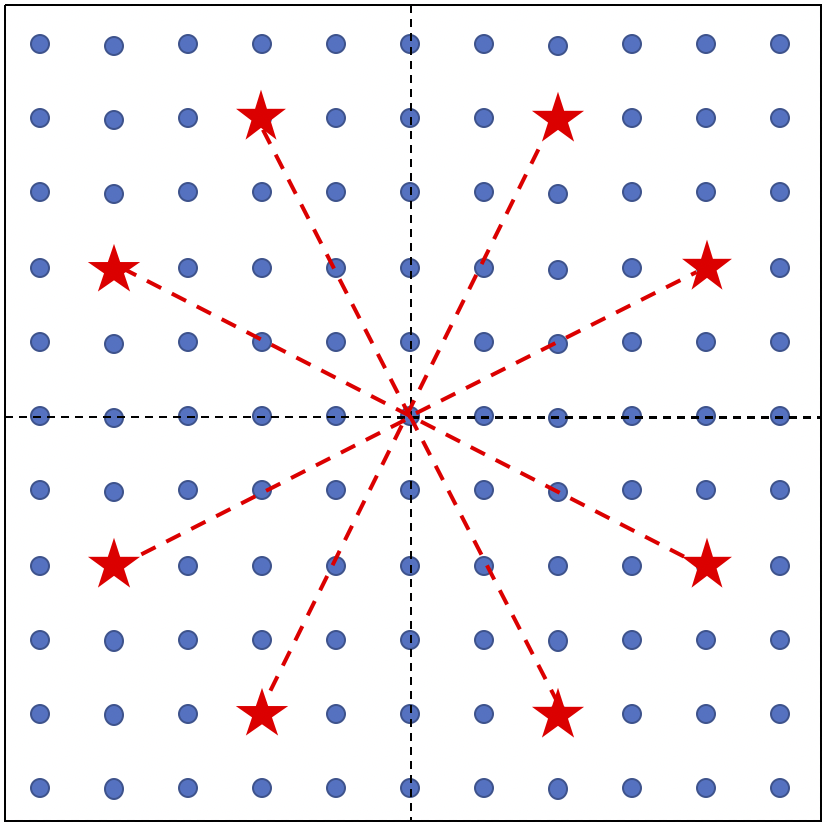}
	\caption{A typical $D_4$ orbit with 8 $\mathbf{k}$'s.}
	\label{d4orb1}
\end{subfigure}%
\begin{subfigure}{.5\textwidth}
	\centering
	\includegraphics[width = 5cm]{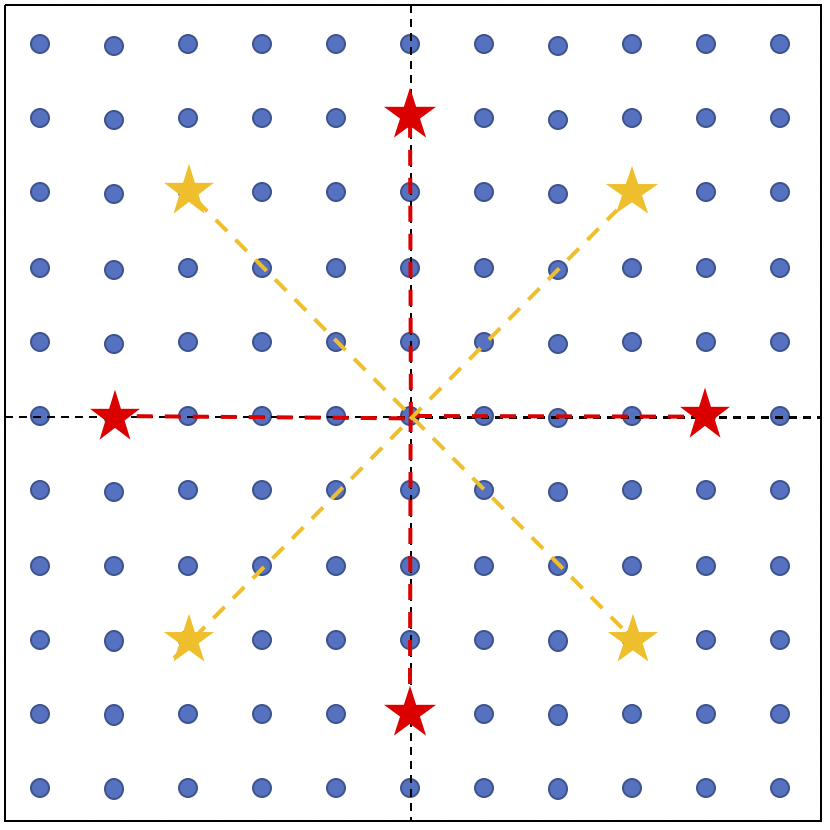}
	\caption{Two special $D_4$ orbits with only 4 $\mathbf{k}$'s.}
	\label{d4orb2}
\end{subfigure}
\caption{$D_4$ group demonstration}
\end{figure}
\end{Exp}

\begin{Exp}
\textbf{Group orbits with $G=SE(2)$ and approximations}

As the last example, we examine the case of $G=SE(2)$ with $H=SO(2)$. This is a continuous Lie group, and the frequency space, rigorously, should be continuous and infinite. All group orbits would be rings in frequency space, with each radius value $\rho$ represents one orbit, as shown in Figure~\ref{so_orb1}.

\begin{figure}[h!]
\centering
\begin{subfigure}{.5\textwidth}
	\centering
	\includegraphics[width = 5cm]{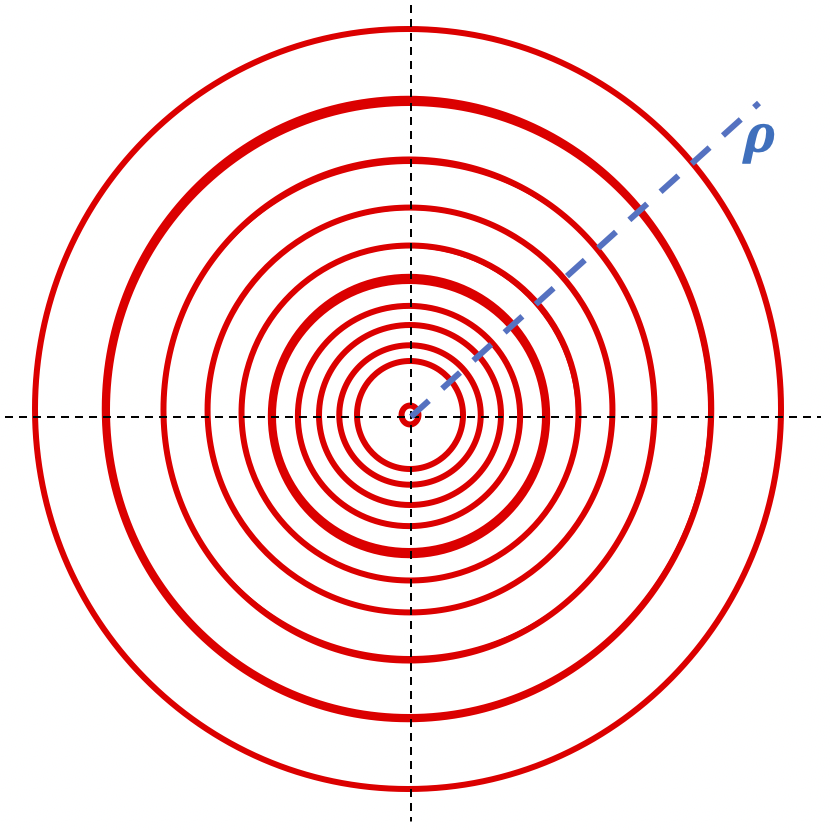}
	\caption{Typical $SO(2)$ orbits in continuous $k$-space.}
	\label{so_orb1}
\end{subfigure}%
\begin{subfigure}{.5\textwidth}
	\centering
	\includegraphics[width = 5cm]{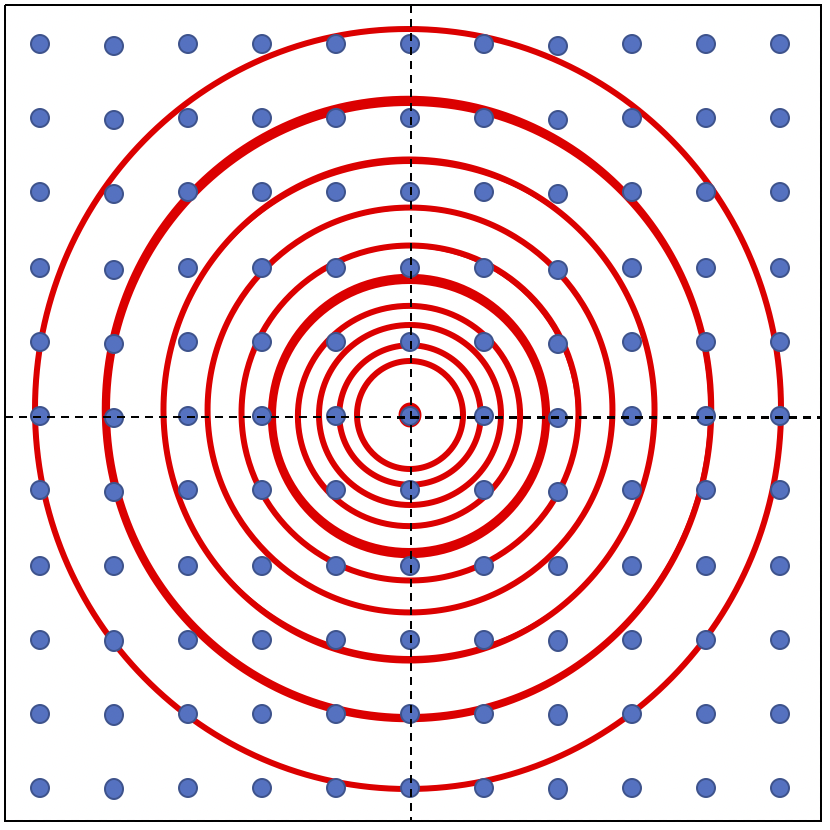}
	\caption{Approximate $SO(2)$ orbits in discrete $k$-space.}
	\label{so_orb2}
\end{subfigure}
\caption{$D_4$ group demonstration}
\end{figure}

Continuous frequency space cannot be used directly. In practice, we always have a discrete frequency space, and, therefore, approximations are necessary. For example, for square images as in examples above, the frequency is also square shaped. and we can use the approximating scheme shown in Figure~\ref{so_orb2}: choose rings with discrete (instead of continuous) radius values $\rho$ to approximate the continuous orbits.
\end{Exp}

\subsubsection{Equivariance from induced representation}
The three above examples has visualized group orbits in the spectral domain. The importance of them is due to the fact that all the frequency branches (components from Fourier transformation) within a $H$-orbit, $\{d_i(\mathbf{k})\}_{\mathbf{k}\in O_H}$, where $i$ labels the on-site dimension (feature maps' index), actually form a basis of $G$-reps; and this $G$-rep is irreducible iff, at any $\mathbf{k}$ in the orbit $O_H$, all $d_i(\mathbf{k})$'s form an irrep of the stablizer $H_0^{\mathbf{k}}$. A $G$-irrep decomposition of signals in the spectral domain could then simply achieved by decomposing the whole spectral domain into different $H$-orbits as shown in the above examples: 
\begin{itemize}
\item
$G=P_4$: $G$-irreps correspond to "4-stars" in the spectral domain; 
\item
$G=P_{4m}$: $G$-irreps correspond to "8-stars" (mostly) in the spectral domain; 
\item
$G=SE(2)$: $G$-irreps correspond to rings in the spectral domain.
\end{itemize}
The above procedure construct $G$-irreps (corresponding to orbits) induced from $T$-irreps (corresponding to single frequency points/branches), and is therefore called \emph{induced representation method}. 

\begin{figure}[ht]
\centering
\begin{subfigure}{.3\textwidth}
	\centering
	\includegraphics[width = 4cm]{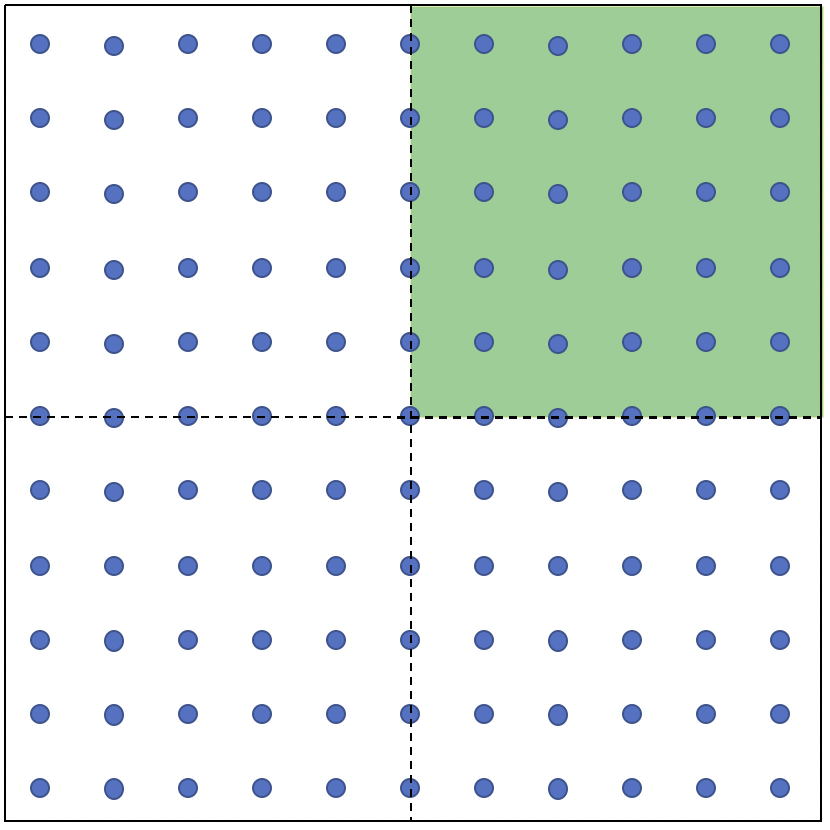}
	\caption{$G=P_4$.}
	\label{c4w}
\end{subfigure}%
\begin{subfigure}{.3\textwidth}
	\centering
	\includegraphics[width = 4cm]{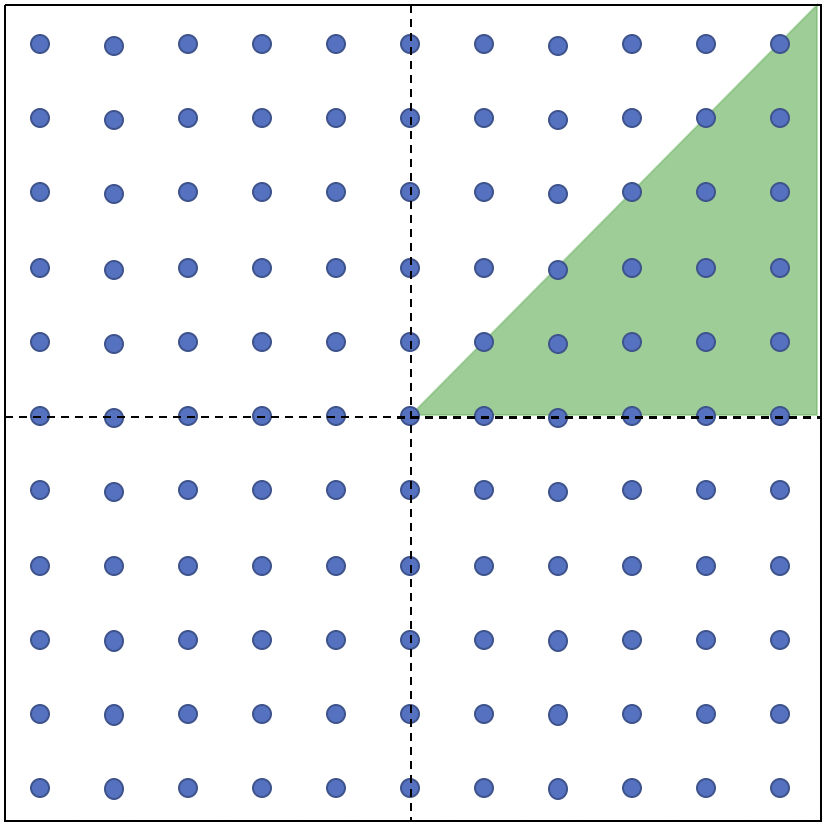}
	\caption{$G=P_{4m}$.}
	\label{d4w}
\end{subfigure}%
\begin{subfigure}{.3\textwidth}
	\centering
	\includegraphics[width = 4cm]{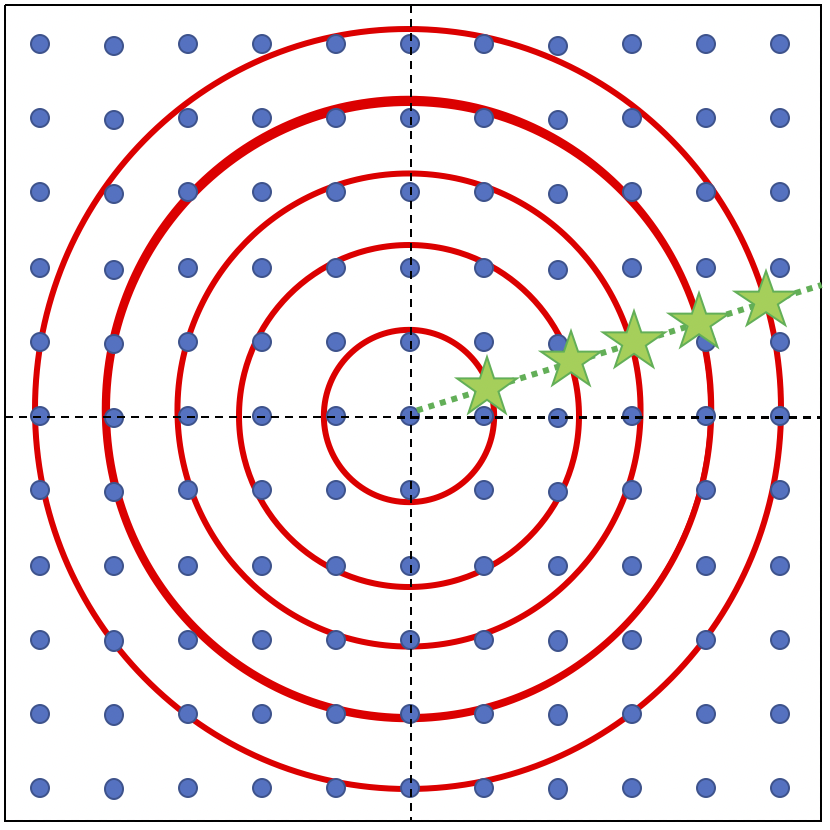}
	\caption{$G=SE(2)$.}
	\label{sow}
\end{subfigure}
\caption{Independent weights: green colored regions are independent parameters during learning.}
\end{figure}

The reason one wants to decompose signals into $G$-irreps, as demonstrated in previous works, is due to the fact that group transformations do not mix different irreps, and, within each irrep, always have a simple transformation rule. 
In the case where all stablizers are trivial ($G=P_4$ and $G=SE(2)$) or only trivial on-site $H_0^{\mathbf{k}}$-irrep is taken into account ($G=P_{4m}$), a single orbit in each feature map forms an $G$-irrep, which is exactly the case in our study, and, as stated in the main article, to achieve equivariance for a convolution operation, one could simply assign the same weights to all coordinates $\mathbf{k}$ in the same orbit:
\begin{itemize}
\item
$G=P_4$: Only $1/4$ parameters are independent, e.g. the green square in Figure~\ref{c4w}; 
\item
$G=P_{4m}$: Only $1/8$ parameters are independent, e.g. the green triangle in Figure~\ref{d4w}; 
\item
$G=SE(2)$: Only a vector of radius values are independent, i.e. the green stars in Figure~\ref{sow}.
\end{itemize}

We would like to mention some important advantage of this approach.
\begin{enumerate}
\item 
With the $G$-irrep decomposition, group actions do not mix different orbits, therefore the equivaraince constraint on learnable parameters could be easily implemented, which has a much simpler form than in the real space signals.
\item
Convolution in spectral domain is a point-wise multiplication, which, in general, could accommodate more learnable parameters. As a comparison, in real space, convolution kernels are usually small compared with the whole. One could use larger kernels with padding, which, however, is redundant as padding does not contain additional information. (For example, in the case of Harmonic Networks, although a large number of radius values could be applied, the number of independent parameters is still determined purely by kernel size, which then results into an over-parametrization.)

It is important to allow more parameters: equivariance of symmetry operations puts constraint on parameters, and therefore reduce the number of independent learnable ones. This, however, is from a purely geometric consideration which is not sufficient in practice. Therefore, a method, which could respect all geometric (equivariant) constraints and, at the same time, accommodate as much learnable parameters as required by other aspects of learning, should be highly appreciated.

\item 
From the illustration above, we notice there is a nice bridge from discrete point group $H$ and continuous Lie group $H$: as $\mathbf{k}$-space become denser and more rotations are considered, group orbits simply change from discrete points to continuous rings, as shown below in Figure~\ref{orbs}:

\begin{figure}[h!]
\centering
\includegraphics[width = 10cm]{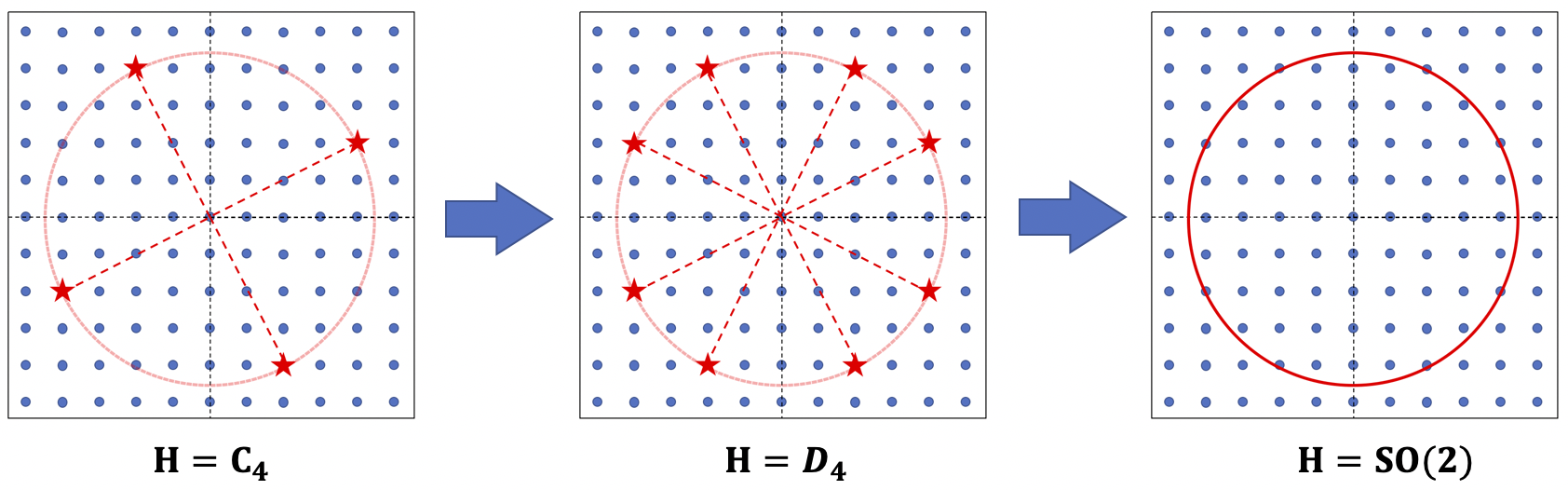}
\caption{Group orbits change from discrete to continuous.}
\label{orbs}
\end{figure}

Actually, $C_4$ is a subgroup of $SO(2)$, and the orbit change can therefore be easily comprehend in the induced representation method. Ignoring the change of $\mathbf{k}$-space shape, the above transition provides a clear interpretation of the discrete-orbit approximation we made for continuous Lie groups.
\end{enumerate}

\section{SieNet: Spectral Induced Equivariant Network}
Now we construct operations respecting equivariance of the full symmetry group $G$ using the frequency space, i.e. a spectral representation, which we term as \textbf{s}pectral-\textbf{i}nduced \textbf{e}quivariant \textbf{net}work (SieNet). More specifically, we would construct $G$-equivariant convolutions, pooling, and nonlinearity. We have rigorous results for discrete symmetry groups, e.g. $P_4$ and $P_{4m}$, but make necessary trade-offs for continuous groups, e.g. Lie groups $E(2)$ and $SE(2)$.

\subsection{G-equivariant Multiplication (convolution)}
We firstly consider the simple operation: a point-wise multiplication. This is a close analogy to conventional convolutions in the real space, and can be viewed as a "convolution" in the spectral domain.

According to \eqref{act_spectr}, a general element in $G$ would firstly permute the frequency coordinate, from $\mathbf{k}$ to $\mathbf{k}'$ in the same orbit $O_{H}^{\mathbf{k}}$, and then shift the phase of the component $d(\mathbf{k}')$, by an amount $-(\mathbf{k}'\cdot\hat{\tau}_g)$ where $\hat{\tau}_g = (\tau_x, \tau_y)$ is the vector representing the translation. The simplest solution would be a point-wise multiplication:
\begin{align}
d(\mathbf{k}) \;\longrightarrow\; \lambda[O_{H}^{\mathbf{k}}]\cdot d(\mathbf{k}).
\end{align}
Compared with \eqref{point_prod}, the only difference is that the learnable parameter $\lambda$ is now a function of orbits $O_H$ rather than coordinate $\mathbf{k}$. As a result, all coordinates $\mathbf{k}$ in the same orbit would share the same $\lambda$.

In the case of $G=P_4$, nearly (except the origin) each orbit contains 4 points. Therefore, the number of free parameters actually get reduced by $3/4$. In the case of $G=SE(2)$ with, each orbit is a ring in the spectral domain, and therefore $\lambda$ should only depends on the radius of the ring, which is achieved with approximations (Gaussian interpolations) in practice. 

\subsection{G-equivariant Orbit Pooling}

Pooling is an operation that reduces the number of variables, and compresses information. Since we are already using the spectral representations, a natural choice for pooling would be \emph{spectral pooling}. We would further impose the equivariance condition. 

Since spectral pooling is actually a (linear) point-wise mask, we could then apply the lesson learned above in the case of multiplication: treat all $\mathbf{k}$ coordinates in the same orbit equally. Therefore, we propose an "orbit pooling": each pooling contains complete orbits. And in general we "peel" orbits on the boundary of the spectral domain, which are high frequency components. For square image cases, this simply corresponds to cut the two spatial (frequential) dimensions by the same size.

Reminded that the conventional max-pooling in real space is not equivariant even for translation symmetry alone, the current orbit-pooling method provides a simple yet rigorous access to equivariance. This further justifies the advantage of spectral representation.

\subsection{G-equivariant Nonlinearity}
The above multiplication and pooling are linear operations. As required by general function approximator theory, there must be a proper nonlinear operation. Besides, since both the multiplication and pooling are point-wise operations while pooling throws high frequency information, the nonlinearity must include an interplay between different frequency components, to transfer the information into lower frequency branches before the higher frequency ones being abandoned.

In practice, it's not easy to design a nonlinear function directly in the spectral domain that could satisfy the above requirements. Therefore, instead of being restricted in the spectral domain, we use the traditional $ReLU(\cdot)$ function, which is a simple point-wise function in the real space but a function with sophisticated interplay between branches in the spectral domain. As this operation in real space is point-wise (or global), it leaves the geometric relation unchanged, and hence is equivariant w.r.t arbitrary symmetry groups.

\setcounter{equation}{0}
\chapter{Recurrent Neural Network as a Dynamical System}\label{ch:rnn_dyn}

In this chapter, we would deliver a demonstration of secondary measurement ($\partial$M) methodology, through an indirect $\partial$M.
As introduced in the prologue, a $\partial$M treats a trained NN model as an efficient approximating representation of the original underlying system.
With this representation, an indirect $\partial$M measures quantities of interests about the original system.
In this chapter, the underlying systems we would investigate are nonlinear dynamical systems, where chaotic regimes exist that are extremely challenging for forecasting.
A recurrent neural network model is trained as a representation of the original dynamical system, and we study the dynamics it exhibits by exploiting characteristics and techniques developed in dynamical system theories, e.g. Lyapunov exponents and dimensions.
On the one hand, these measurements produce a description about the generating dynamics hidden in a finite sample of collected data, and hence provide further insights in theoretical modeling as the next stage of research;
on the other hand, the study of dynamical characteristics enables us to evaluate the asymptotic behavior and error accumulation of the trained machine, which offers a way to estimate its reliability in forecasting tasks.

\section{Background}\label{sec:intro}
% Modeling temporal data, i.e. time series, has been an important topic in various fields. 
% Depending on the complexity of the generating dynamics, different approaches have been implemented. 
% For example, 
% in econometrics, Auto-Regressive (AR) model, Moving-Average (MA) model and their variants have been thoroughly explored;
% and in control system theory, state-space model is proposed to explicitly capture the status of the internal dynamics.
% Among different treatments, there are two closely related methods: the Recurrent Neural Network (RNN) model used in Deep Learning (DL), and the Reservior Computer (RC), which share exactly the same model structure. 
% Different from models restricted to linear dynamics or dynamics with a fully parametrized structure, a RNN (or RC) with large enough capacity has the potential to model arbitrarily complicated nonlinear dynamics, and has drawn attention of researchers from various areas.

Machine learning in data science is presented as a black-box approach 
where an inferential training process applied to a data set
results in the machine's ability to reproduce and
potentially classify, predict or synthesize data.
These methods typically provide limited deductive insights into the 
relationships in training data.

One interesting class of training data are time series, and pioneering work 
\cite{maass2002real,jaeger2004harnessing} 
on Echo State Machines introduced the idea of 
training on a discretely sampled trajectory of a 1-parameter non-linear (delay) model with
chaotic dynamics.  
A natural motivation is that the exponential sensitivity to initial conditions, due to a positive
Lyapunov exponent, make such trajectories difficult to predict.  
Subsequently, the dynamical systems community employed the Echo State architecture, which is an
example of a Recurrent Neural Network, to study Reservoir Computing (RC) networks 
\cite{pathak2017using,pathak2018model} trained on chaotic time series sampling chaotic 
trajectories of the Lorenz system, a 3-dimensional system of non-linear ODE's 
(as well as a compliceted 2D PDE).
In \cite{pathak2017using}, Pathak and coworkers showed that an RC network
trained on a time series of length $T$ 
can simulate a continuation of this time series after the training period, that has the same generic 
long term behavior of the generating chaotic system.

In this paper we build on this approach.
The underlying philosophy is to use a training time series sampling a class of dynamical 
behavior in some ``low-dimensional data space," while the trained machine effectively attempts 
to duplicate this dynamics in a ``high-dimensional machine space."  While stability generally 
suggests some kind of attractor structure, using data sampled from a chaotic ODE allows for a 
more rigorous characterization of this concept.  The beauty of (continuous time) systems of 
chaotic ODE's is that a single, sufficiently long trajectory samples the attractor ergodically.  
Consequently, an analysis of a single trajectory can reveal the fractal dimension of the attractor, 
which can be associated with entropic and information measures.
In \cite{pathak2017using}, Pathak and coworkers explicitly 
calculate an approximation of the Lyapunov exponents of the machine dynamics
using the continuation of the (discrete) training trajectory from which they 
can estimate the machine attractor dimension through the Kaplan-Yorke conjecture.
The degree to which such an analysis works in general, depends to both the training method and the complexity of the underlying attractor.
Of course the machine cannot infer more information than that in the training data.

We extend this approach by training and analyzing the attractor structure more broadly,
and by considering a data space with a more complex attractor structure.
RC training uses an inversion process to determine a linear projection from the
machine space to the data space, and is trained on a single (relatively long) 
training time series. 
We develop a Deep Learning (DL) training algorithm using standard tools from the DL 
community based on auto-regressive prediction, that is trained on multiple shorter training time series.
Additionally, instead of focusing on a single example of a well-performing trajectory
on a well-performing machine,
we analyze multiple machine trajectories and multiple trained machines.
We find that the DL training method is noticeably better at reproducing the attractor dimension
for training data from the Lorenz system, and that this improvement becomes very striking
when the complexity of the attractor structure increases, using hyper-chaotic trajectories of the 
4-dimensional R\"ossler system, i.e. when there are two positive Lyapunov exponents.

The dynamical systems community may find the DL training method beneficial for sampling the data space 
much more broadly. 
The DL community by find value in the explicit analysis of the autonomous trained RNN machine using tools 
from non-linear dynamics – the (scale-free) attractor dimension provides insight into machine stability 
and the inferred dynamical structure of the process generating training data; which may complement 
linear statistical approaches.

\section{RNN model for Sequential Learning}

Given a time series $\{\mathbf{u}_t\}_{t=1}^{T}$, an RNN model implements the following dynamics:
\begin{align}\label{eq:rnn}
    \mathbf{h}_{t+1} &= \tanh{\big[\mathbf{W}\cdot\mathbf{h}_t + \mathbf{W}_{in}\cdot\mathbf{u}_t + \mathbf{b}\big]}, \nonumber \\
    \mathbf{u}'_{t+1} &= \mathbf{W}_{out}\cdot\mathbf{h}_{t+1}.
\end{align}
In general cases, $\mathbf{u}'_t\in\mathbb{R}^{k'}$ and $\mathbf{u}_t\in\mathbb{R}^k$ are \textit{output} and \textit{input signals}, respectively. 
And $\mathbf{h}_t\in\mathbb{R}^d$ is called \textit{hidden variables} in DL, or referred as the \textit{state vector} of RC. 
In prediction tasks, one sets $k=k'$, as $\mathbf{u}_t$ and $\mathbf{u}'_t$ aim to capturing the same signal, where $\mathbf{u}_t$ is the true data, while $\mathbf{u}'_t$ represents the predicted value in Eq.\ref{eq:rnn}.

The $d\times d$ matrix $\mathbf{W}$ captures the information of recurrent connections, while the matrix $\mathbf{W}_{in}$ ($d\times k$) projects information from the signal space to the hidden space and $\mathbf{W}_{out}$ ($k\times d$) does it inversely.
The vector term $\mathbf{b}$, termed as bias in DL research, provides a constant shift to improve the modeling capacity.
Formally, an RNN model applies the following projections:
\begin{equation}
    \centering
    \begin{tabular}{ccc}
        $\mathbf{u}_t$ &  & $\mathbf{u}'_{t+1}$\\
        $\big\downarrow$ & & $\big\uparrow$ \\
        $\mathbf{h}_t$ & $\xrightarrow{\text{RNN}}$ & $\mathbf{h}_{t+1}$
    \end{tabular}.
\end{equation}
The difference between DL and RC treatments lies in the training mechanism.
In DL, all three matrices, i.e. the matrix tuple $\mathcal{W} = (\mathbf{W}_{in}, \mathbf{W}, \mathbf{W}_{out})$, are optimized simultaneously through back-propagation of error:
\begin{align}\label{eq:dl_train}
    \mathcal{W}_0 = \operatorname*{argmin}_{\mathcal{W}}F_{\mathcal{W}}\big[\mathbf{u}_{1:T}, \mathbf{u}'_{1:T}\big],
\end{align}
where $\mathbf{u}_{1:T} = \{\mathbf{u}_1, \mathbf{u}_2, \cdots, \mathbf{u}_T\}$ is the ground-truth sequence of the variable $\mathbf{u}_{t}$ from $t=1$ to $t=T$, and $\mathbf{u}'_{1:T}$ being the corresponding one output by the model. And $F_{\mathcal{W}}[\cdot]$ is a objective \textit{loss function} to be minimized, which defines a measure of distance between two arbitrary time sequences.
The training in RC, however, optimizes the readout mapping only, i.e. $\mathbf{W}_{out}$, in the following way:
given a RNN with hidden dimension $d$, a data sequence of length $T$ could be fitted as below:
\begin{align}\label{eq:rc_fit}
    \mathbf{W}_{out} &= \mathbf{U}\mathbf{H}^{T}(\mathbf{H}\mathbf{H}^T)^{-1},
\end{align}
where $\mathbf{u}$ is a $T\times 3$ matrix with each row representing a 3-dimensional vector $[x_t, y_t, z_t]$, and $\mathbf{H}$ is a $T\times d$ matrix with each row representing a $d$-dimensional hidden state $\mathbf{h}_t$.
This is motivated from the following deduced normal form:
\begin{align}\label{eq:rc_normal_form}
    \mathbf{W}_{out}\mathbf{H} = \mathbf{U} \quad\Rightarrow\quad 
    \mathbf{W}_{out}\mathbf{H}\mathbf{H}^T = \mathbf{U}\mathbf{H}^T
\end{align}

Intuitively, the recurrent connection $\mathbf{W}$ among hidden variables (i.e. the state vector) iteratively integrates the impact of the past into the next-step evolution, hence makes RNN a natural candidate for learning dynamics.
To make accurate prediction from the readout mapping $\mathbf{W}_{out}$, the internal dynamics of the reservior should not lose information about the evolution.
In practice, especially in Reservior Computing, $\mathbf{h} \in \mathbb{R}^d$ usually lives in a very high dimensional space, to provide sufficient capacity to store information from the past evolution.
There are two different scenarios of prediction:
\begin{enumerate}
    \item \textbf{One-step-ahead prediction}: at each step of the prediction, there is a ground truth value $\mathbf{u}_t$ to forecast only the very next step $\mathbf{u}'_{t+1}$;
    \item \textbf{Auto-regressive prediction}: the prediction process uses a predicted value $\mathbf{u}'_t$ to derive the next step $\mathbf{u}'_{t+1}$.
\end{enumerate}
The difference between the two above scenarios results into different internal dynamics of the reservior:
\begin{enumerate}
    \item in one-step-ahead prediction, there is constantly an input signal fed into the reservior, resulting into a \emph{driven dynamics} as depicted in Fig.~\ref{fig:driven_rnn}:
    \begin{figure}[!ht]
        \centerline{
        \includegraphics[height=1.0in]{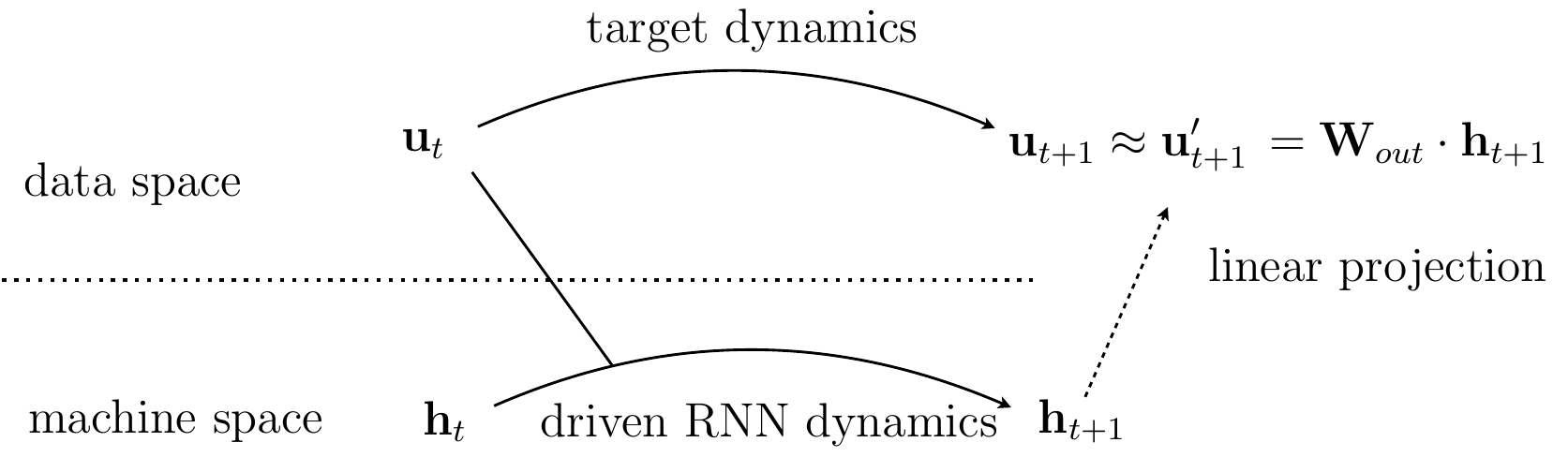}
        }
        \caption{data-driven RNN dynamics and projection;}
        \label{fig:driven_rnn}
    \end{figure}
    \item in auto-regressive prediction, as one takes the predicted value $\mathbf{u}'_t$ as the input signal, Eq.\ref{eq:rnn} can be rewritten as:
    \begin{align}\label{eq:auto}
        \mathbf{h}_{t+1} &= \tanh{\big[(\mathbf{W} + \mathbf{W}_{in}\mathbf{W}_{out})\cdot\mathbf{h}_t + \mathbf{b}\big]}, \nonumber \\
        \mathbf{u}'_{t+1} &= \mathbf{W}_{out}\cdot\mathbf{h}_{t+1},
    \end{align}
    which is an \emph{autonomous dynamics} without external driving signal as depicted in Fig.~\ref{fig:auto_rnn}:
    \begin{figure}[!ht]
        \centerline{
        \includegraphics[height=1.0in]{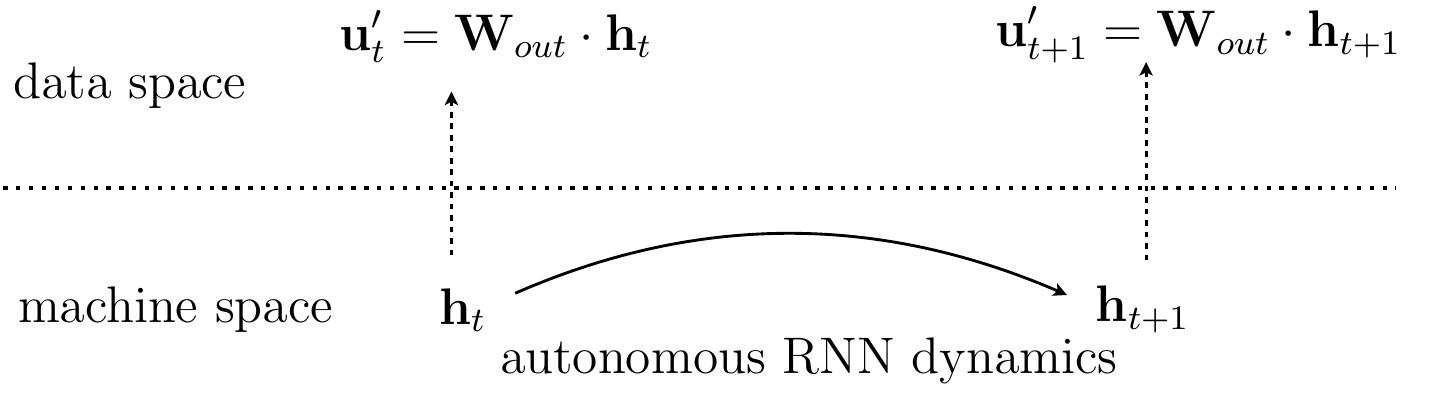}
        }
        \caption{Autonomous phase: RNN dynamics simulates data;}
        \label{fig:auto_rnn}
    \end{figure}
\end{enumerate}
In real-life applications, auto-regressive prediction, i.e. forecasting multiple steps ahead, is more useful while challenging due to the error accumulation.
To achieve better prediction, a deeper understanding about the internal dynamics is required.

In fact, for time series prediction, it is debatable that if error calculated on finite sequences is an appropriate measure of the model performance.
All techniques for temporal data modeling, e.g. ARIMA model in econometrics and state-space model in control systems, are aiming at modeling the \textit{generating process} of the data, which is the \textit{underlying dynamics}, rather than any \textit{finite length sequence}.
However, in practice, the model training (or, learning or fitting) procedure could merely make advantage of a finite sample of data, which is the only available observation, and a direct comparison with the true dynamics is not applicable.
The autonomous nature of long-term prediction tasks further complicates the problem, as error calculated on a set of finite sequences does not guarantee a statistically stable error accumulation behavior.
Therefore, one could not derive an error bound estimation as achieved in modeling with feedforward neural networks, where one could draw conclusions about the statistical relation between input and output signals.

Beyond prediction accuracy, it is also of researchers' interest to understand the underlying dynamics itself.
While some glassbox modeling approaches, e.g. ARIMA models, directly parameterize the data generation process, RNN models, after being trained on a sample of sequence, does not provide an explicit description about the dynamics.
Therefore, a post-training investigation is desired to deliver further insights about the underlying dynamics.

The above question is closely related to the representation learning in machine learning area as well.
A representation of a raw signal is an efficient alternative which can be used in various downstream tasks.
It is crucial that the representation contains all valuable information of the original signal.
A typical example is the representation of a sentence or paragraph used in general natural language processing (NLP), where, taken a textual sequence as input, a RNN model is used to produce either an output sequence or a single token as the representation of the original data.
While in general NLP, all sequences are of finite length which supports a finite sequence representation naturally, for time series data, any finite sequence does not guarantee a proper representation.
A representation of a sentence is designed to contain all semantic meaning, while a representation of a time series should be designed to contain the underlying dynamics of the data generating process.
In this scenario, instead of using any sequence or token generated by a RNN model, the model itself should be regarded as a "representation" of the time series (or, more precisely, of the underlying dynamics), which brings an additional reason to explore the dynamics rather than focusing on any finite sequence.

Motivated by the above two questions, we provide a dynamical system perspective to analyze a general RNN model trained to learn time series dynamics.
In previous works, RNN has been deemed either as a dynamical system where the internal connection $\mathbf{W}$ is generated by a pre-defined random distribution, or purely as a black-box model to learn time series data.
We combine these two perspectives and study a trained RNN, rather than a randomly generated one, as a dynamical system.
To obtain a model with good performance in time series prediction tasks, we train a RNN by applying state-of-the-art techniques used popularly within the DL community.
And we also explain each technique used in practice from a dynamical system perspective, which further justifies the theoretical motivation.
After the training, many numeric indices, including Lyapunov exponents and attractor dimensions, can be applied as a measure on the trained RNN model to quantitatively describe the internal machine dynamics and provide insights about both the data-generation process and an error bound estimation.
By taking advantage of this combination of methods and concepts from two fields, we aim at bridging the DL and RC communities, which, to our knowledge, is the first attempt of a bi-directional application.

\section{Instantiating Dynamical Systems}

In this work, we would use a instantiating dynamical system, Lorentz System, as the target dynamics to learn, to exploit the investigation on DL based approaches.
The Lorentz system has continuous 3 dimensional dynamics governed by the differential equations:
\begin{align}\label{eq:lorentz}
    \Dot{x} &= \sigma(y-x), \nonumber \\
    \Dot{y} &= x(\rho-z) - y, \nonumber \\
    \Dot{z} &= xy - \beta z
\end{align}
with parameters set as $\sigma =10$, $\rho=28$, and $\beta=8/3$,
for which this system is known to have chaotic dynamics.  So the differences between trajectories grow exponentially on the Lyapunov time scale which is order unity for these parameters.
To generate a reasonably accurate dataset, we generate discrete time series using a 4th order Runge-Kutta simulation with 
time step $\delta t=\text{1E-3}$.
However, for training purposes, we sampling this dataset using a significantly larger time step.  Here is a sample trajectory $(x, y, z)$ rescaled to the cube $(-1, 1)^3$.
\begin{figure}[!ht]
    \centering
    \includegraphics[height=1.85in]{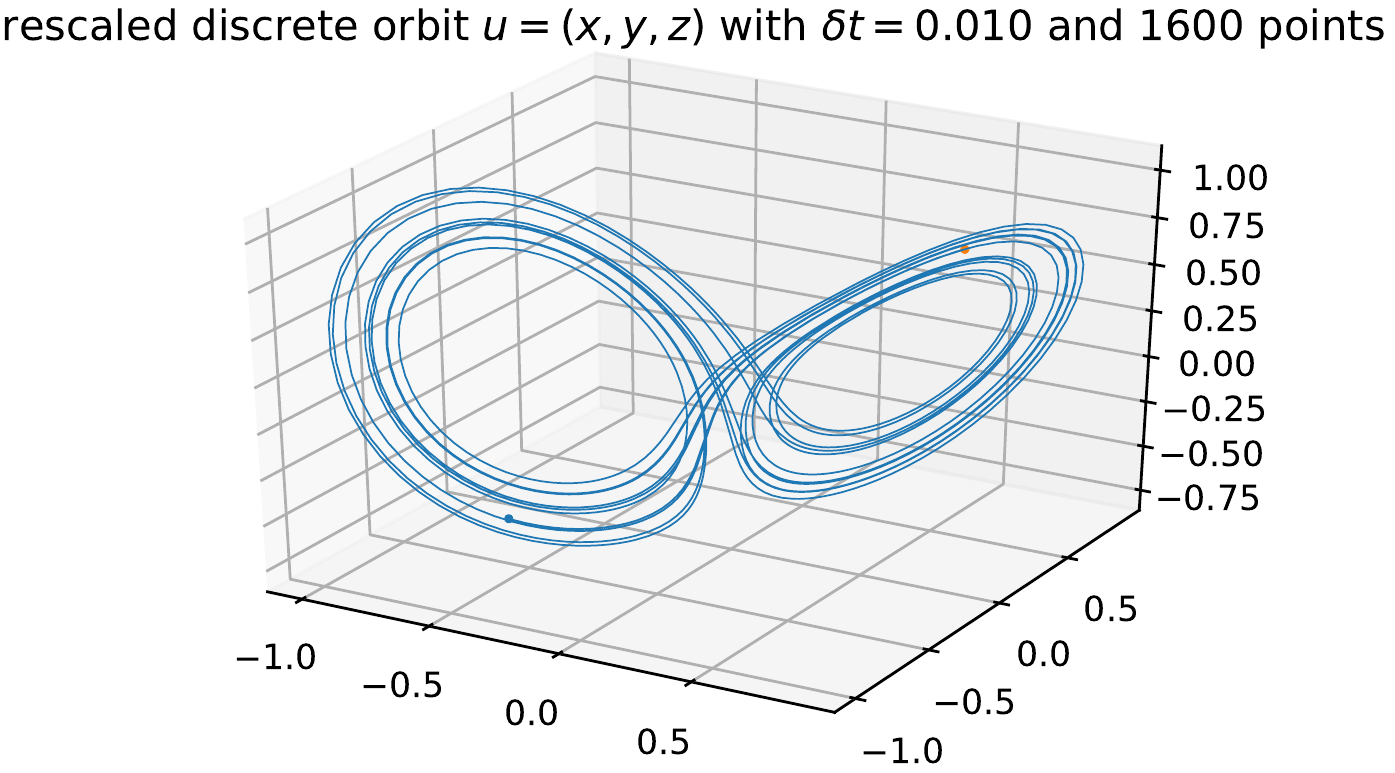}
    \caption{Sample trajectory $(x, y, z)$ over 16 time units, rescaled to the cube $(-1, 1)^3$.}
    \label{fig:LorentzSample}
\end{figure}

\section{Learning Dynamics: a deep learning approach for time series}\label{sec:dl}
In this section, we introduce a deep learning setup for time series data.
Specifically, we are more interested in time series generated by nonlinear dynamics that is technically more challenging and related to real life scenarios, among which, chaotic behavior is the most challenging case to learn and would be chosen as a demonstration in our present work.

Deep learning is a delicate combination of several key components: loss functions, dataset collections, training scheme, parameter initialization, and, the model structure.
Most cross-disciplinary research applying deep learning methods usually focus on directly applying a neural network model.
However, the model structure alone could not produce a good performance.
We now investigate each component in details, and attempt to relate them with topics more familiar to experts in dynamical systems and applied mathematics.

\subsection{Ergodicity and Batch Learning}\label{sec:dl:batch}

As introduced earlier, the "learning" mechanism in traditional RC implementations is achieved through a fitting procedure: one keeps both $\mathbf{W}_{in}$ and $\mathbf{W}$ matrices fixed from a random initialization, and fits a sequence of data points to obtain an optimal $\mathbf{W}_{out}$ matrix by minimizing the difference between the true sequence and the sequence generated from a driven dynamics as in Eq.\ref{eq:rnn}.
This approach, although widely applied, is limited to a single sequence of data during training (i.e fitting), and becomes problematic in learning chaotic dynamics.

The problem of fitting method can be revealed by analyzing details of numeric computation in Eq.\ref{eq:rc_fit}.
As the hidden dimension $d$ has been fixed while the sequence length $T$ being varying (usually larger than $d$), a Ridge regression is usually applied in practice to improve the stability of computation:
\begin{align}\label{eq:fitting}
    \mathbf{W}_{out} &= \mathbf{U}\mathbf{H}^{T}(\mathbf{H}\mathbf{H}^T + \epsilon\mathbb{I})^{-1},
\end{align}
where $\epsilon$ is a small valued constant parameter controlling the regularization strength, and $\mathbb{I}$ is a $d\times d$ identity matrix.
The accuracy of the inverse calculation would significantly drop if $T>>d$.
However, a long sequence with a large value of $T$ is desired to provide a complete description of chaotic dynamics, due to the asymptotic nature of its behavior.
Therefore, in practice, a trade-off between the amount of data and computation accuracy has to be made.

It turns out this trade-off could be mitigated by applying a DL training procedure.
Specifically, one could implement the so-called \textit{Batch Learning} approach, which, instead of using a single sequence, trains a model by simultaneously investigating a batch of $N$ sequences.
With a large batch value $N$, a model can be trained to minimize deviations across a much larger phase space while keeping the length of any single sequence to be small.
It is worth to mention that, although the above error minimization is implemented as a statistical averaging, it is very different from directly obtaining a matrix by averaging across several separately fitted $\mathbf{W}_{out}$'s.
The key reason that Batch Learning DL works well is rooted in the optimization algorithm: stochastic gradient descent (SGD).
SGD optimizes all matrix parameters through multiple steps with a small learning rate, while the traditional fitting method obtains a solution in one shot.
The latter one would easily exaggerates error accumulations in sequence data.

\subsection{Residual Connection for Continuous Dynamics}\label{sec:dl:residual}
We have introduced the standard RNN model structure which provides a promising candidate to learn dynamics. The vanilla version of RNN in either Eq.\ref{eq:rnn} or Eq.\ref{eq:auto}, however, is designed to describe a discrete dynamics, and may experience extra difficulty when learning continuous dynamics where the incremental change in a single step is usually very small.
To mitigate this difficulty, we apply a modified version of hidden variable (state variable) update as following:
\begin{align}\label{eq:res_rnn}
    \mathbf{h}_{t+1} &= (1-\alpha)\mathbf{h}_{t} + \alpha\tanh{\big[\mathbf{W}\cdot\mathbf{h}_t + \mathbf{W}_{in}\cdot\mathbf{u}_t + \mathbf{b}\big]}.
\end{align}
This modification has been widely applied in the RC community~\cite{pathak2017using}, usually termed as a type of Echo State Machine.
With a proper value of $\alpha$, the above form could improve the stability of state space dynamics.

From a Deep Learning perspective, this could in fact be viewed as a \textit{residual connection}, as introduced in the well-known work of He~\cite{he2015deep}.
Similar to the idea of ResNet, by explicitly modeling an identity map in the first part, this modification enforces a continuously evolving trajectory.

\subsection{From 1-Step-Ahead to Seq2Seq}\label{sec:dl:seq2seq}
The training setup is of great importance when applying deep learning method.
Specifically, the performance of the model highly depends on the design of the loss function, which is the objective of the optimization.
In most conventional methods, the time series forecasting focuses on predicting the state one step ahead in the future, termed as \textit{1-step-ahead} learning, where, at each iteration step in the time dimension, the optimization minimizes the following learning objective:
\begin{align}\label{eq:loss_1step}
    F_{\mathcal{W}}[\mathbf{U}_{t}, \mathbf{U}'_{t}] = \frac{1}{k}\sum_{i=1}^{k}|u_{t, i}- u'_{t, i}|^2, \quad \forall t \in [1, T],
\end{align}
where at each step $\mathbf{u}'_{t+1}$ is obtained via Eq.\ref{eq:rnn} by substituting a true value $\mathbf{u}_{t}$. 
While usually showing nice precision in 1-step-ahead prediction, this setup would fail rapidly in the case of long term forecasting with error accumulation.
This is particularly problematic when the time series is generated by nonlinear dynamics, which is the most common scenario in real life applications. 
To boost the model performance in long term forecasting tasks, we instead apply the following objective function:
\begin{align}\label{eq:loss_auto}
    F_{\mathcal{W}}\big[\mathbf{U}_{1:T}, \mathbf{U}'_{1:T}\big] = \frac{1}{kT}\sum_{i=1, t=1}^{k, T}|u_{t, i}- u'_{t, i}|^2,
\end{align}
where each value of $\mathbf{U}'_{t+1}$ in the sequence $\mathbf{U}'_{1:T}$ is generated autoregressively by substituting a predicted value $\mathbf{u}'_{t}$ into the model.

For the deep learning community, this actually can be viewed as a Seq2Seq modeling, with the same weights assigned to both the encoder and the decoder.
Seq2Seq setup has been widely applied in natural language processing tasks.
It is also well known among the DL community, that the tied weights could accelerate the convergence of training. 

From the perspective of a dynamical system, the two objectives actually target at two distinct dynamics.
The 1-step-ahead loss function Eq.\ref{eq:loss_1step} models a system which ground true information are fed into continuously at each step. 
Therefore, the model is instructed to learn a driven dynamics.
As a consequence, the dependence of external input would be amplified through larger weights in $\mathbf{W}_{in}$
As a contrast, the Seq2Seq loss function Eq.\ref{eq:loss_auto} minimizes the error in an autoregressive process, which directly captures an autonomous dynamics.

\subsection{Weight Initialization in Spectral Domain}\label{sec:dl:init}
There is a well-known challenge in DL with sequential model: the \textit{gradient vanishing} problem.
It is rooted in the current framework of DL, which is based on the \textit{backpropagation} (BP) of error terms.
For general RNN models, during the training with a long sequence, the gradient easily gets either amplified (exploding) or diminished (vanishing) when propagating backward in time, resulting into an information loss on earlier steps of the sequence.
However, as mentioned above, a training setup targeting at a multi-step prediction is desired to learn the correct dynamics, therefore preventing gradient vanishing or exploding in practice becomes necessary.

While, during the training, all matrix weights keep being updated such that a direct control of dynamics becomes infeasible, it is helpful to discuss the question of weight initialization.
In DL, initializing weights into appropriate regions could significantly accelerate the learning process.
Current initialization methods, including \textit{He initialization} and \textit{Xaiver initialization}, target at BP in deep networks based on the idea of normalization.
To the best of what we know, there is no initialization approach working for general recurrent structures.
We hence propose a new method to initialize the recurrent connection, $\mathbf{W}$, to prevent unwanted machine behaviors. 
The method normalizes all weights in the recurrent connection $\mathbf{W}$ by scaling them through the following formula:
\begin{align}\label{eq:spectral_norm}
    W_{ij} = \frac{\rho_0}{\rho_{\text{max}}}\hat{W}_{ij},
\end{align}
where $\hat{W}_{ij}$ are obtained by sampling from a uniform distribution $\mathbf{U}[0,1]$, and $\rho_{\text{max}}$ is the largest eigenvalue of the original sampled matrix $\mathbf{\hat{W}}$, which is also called the \textit{spectral radius}.
The scaling factor $\rho_0$ is a hyperparameter empirically determined by researchers, usually taking values slightly above $1$, e.g. within $[1.1, 1.4]$.
We term this new method as \textit{spectral initialization}.

Although theoretically not a fully founded approach, the above proposed method in fact has a quite simple intuition.
It has been proven in \cite{pmlr-v9-glorot10a} that, for any matrix $\mathbf{W}$, the spectral radius $\rho_{\text{max}} < 1$ would be a sufficient condition for gradient vanishing, which should be prevented.
On the other hand, an extremely large value of $\rho_{\text{max}}$ would result into more dimensions in the phase space expanding and therefore more likely into a gradient exploding scenario.
Indeed the argument above is not mathematically rigorous, however, empirically performs well.
It is also interesting to notice that, the gradient problem in the DL community already take advantage of a dynamical system perspective to obtain intuitions~\cite{pmlr-v9-glorot10a}.

\subsection{Experimental Setup}\label{sec:dl:experiment}

We sample the original data simulated from Lorentz dynamics with a frequency $1/20$, resulting in a time interval $\text{2E-2}$ in practice.
The first $1000$ steps in the simulation are skipped before collecting the training data, to focus more on the attractor space.
During training, a Seq2Seq method is implemented, with the initial $100$ steps fed as a warm-up driving without error calculation.
To apply batch training, we generate $N=8000$ sequences of length $T=120$, and use mini-batch size $n=200$ to implement a stochastic training to prevent being trapped in local minimum.
The sequence lengths used in validation and testing are fixed as $T_v = 400$ and $T_t = 150$ respectively.

For simplicity, we build a single layer RNN model with hidden dimensions $\{600, 900, 1200\}$.
We implement the standard DL framework \textit{PyTorch} and train neural network models using \textit{RMSprop}~\cite{HintonSlides}, a variant of gradient based optimization algorithms that is popular among sequential tasks, with learning rate fixed as $\textit{LR}=\text{1E-3}$. 
Training terminates either after $100$ epochs of batch iteration or when an early-stopping signal is detected\footnote{This can be seen when the validation error starts increasing due to an overfitting}.

\section{What Have We Learned: Hidden Space Dynamics and Attractor Dimension}\label{sec:result}

DL based methods are often debated due to their black-box nature.
Interpretibility, however, is an essential problem in time series modeling, where, besides prediction, a comprehension about the underlying dynamics is desired.
To achieve this, we view a trained RNN model itself as a dynamical system, and provide a description for the learnt dynamics by utilizing concepts and attributes studied in the dynamical system theory.

\subsection{A Geometric Perspective: Attractors in Hidden Space}\label{sec:result:attractor}

In the theory of dynamical system, the asymptotic behavior is the most important feature of the dynamics, which produces different types of attractors that describe the system status when time approaches infinity.
For instance, a fixed-point represents a constant state in the phase space, while a limit cycle describes a periodic asymptotic behavior.
Strange attractors, on the other hand, is a special class of attractors which features chaotic dynamics.

When modeling time series data obtained from real life, researchers usually start with some basic intuitions. 
For example, the market value of a financial product should not stay as a constant in a long term forecasting, while a periodic behavior could approximate certain business cycles.
These intuitions are closely related to asymptotic behavior of the underlying dynamics, and provide a strategy to examine the plausibility of the trained model.
Corresponding to the two examples above, a trained model with a fixed point attractor could not provide a reasonable description of a market value time series, while a model with a limit cycle attractor could indeed capture a time series of certain business cycle.
In our current work, we aim at learning chaotic dynamics, whose asymptotic behavior exhibits strange attractors in the phase space.

Paradoxically, despite the inherent unpredictability of chaotic trajectories over sufficiently long times,
distinct chaotic Lorenz trajectories are essential equivalent when viewed over long times.
They sample the strange attractor which has a fractal dimension, ergodically.
The widely 
used Kaplan-Yorke hypothesis provides a procedure to determine the
Lyapunov dimension $D_L$ from the Lyapunov exponents.
The Lyapunov dimension is generally believed to be a good measure of the
attractor dimension, and typically is very close to the Renyi entropy 
or dimension $S_1$.
The Lyapunov exponents, $\lambda_i$ are formally the principal values of
the time-$t$ flow-map in the large $t$ limit.
For an ODE-based dynamical process,
an infinitesimal time advance is a neutral perturbation 
and one Lyapunov exponent vanishes identically.

For chaotic parameters, ergodicity implies 
generic trajectories approach other trajectories arbitrarily closely. 
Consequently, the Lyapunov exponents can then be estimated from 
a single trajectory, provided the sampling time-step is sufficiently small
and the trajectory sufficiently long.
Given a discretely sampled trajectory of $T$ points 
i.e. $\mathbf{u}(j\delta t)$, with $j=1,\ldots,T$
\begin{equation}\label{lexp}
\lambda _i\approx 
\frac1{T\delta t}\sum_{j=1}^T\log(p_i^{\scriptscriptstyle j})
\quad\Leftrightarrow\quad
e^{\lambda_i(T\delta t)}\approx\prod_{j=1}^Tp_i^{\scriptscriptstyle j}
\end{equation}
where $p_i^{\scriptscriptstyle j}$ is the $i$-th principle value of
a local deformation of the time-$t$ flow at $\mathbf{u}(j\delta t)$, 
lineraly estimated using the ODE's Jacobian evaluated at that point.
This procedure
"naturally" weights contributions more heavily when the dynamics is slow.
The rate of convergence of (\ref{lexp}) depends quite sensitively on
the time-step $\delta t$, the trajectory length $T$ and the overall 
flow pattern on the attractor.

Even for the well-studied Lorenz system, convergence is quite slow.
As an example, Table~1 presents numerical calculations of the Lorenz system exponents
over a fixed time interval $T\delta t=100$ (about 100 Lyapunov times).
We use the popular parameters $\rho=28$, $\sigma=10$, $\beta=8/3$ 
and vary $\delta t$. 
We present averages of the three exponents over 64 trajectories (all essentially on the attractor) in Tab.\ref{tab:lyp_deltaT}.
\begin{table}[!ht]
\centering
\begin{tabular}{|l|c|c|c|}
\hline
$\delta$  &  $\lambda_1$  &  $\lambda_2$ & $\lambda_3$  
\\ \hline
0.05      & 3.03$\pm$0.03 &  1.102 $\pm$ 0.04 & -33.84 $\pm$ 0.24
\\ \hline
0.01      & 1.18$\pm$0.02 &  0.512 $\pm$ 0.01 & -15.86 $\pm$ 0.03
\\ \hline
$10^{-3}$ & 0.94$\pm$0.02 &  0.027 $\pm$ 0.02 & -14.67 $\pm$  0.02
\\ \hline
$10^{-4}$ & 0.90$\pm$0.02 & -0.006 $\pm$ 0.02 & -14.57 $\pm$  0.02
\\ \hline
\end{tabular}
\caption{Lyapunov exponents of Lorentz system with varying sampling step sizes.}
\label{tab:lyp_deltaT}
\end{table}

It can be seen that, even when averaging (\ref{lexp}) over 100 Lyapunov times one needs a small value of $\delta t$ to retrieve the
vanishing exponent for continuous time dynamics as well as the conservation of $\sum\lambda_i$
for the Lorenz system since the diagonal of the Lorenz Jacobian is constant.
Moreover, the deviation for the known exponents is systematically larger than the statistical
variation.

As a comparison, the Kaplan-York hypothesis for the Lyapunov dimension is
\begin{align}\label{eq:dim}
    D_L = k + \frac{\sum_{i=1}^{k}\lambda_i}{|\lambda_{k+1}|},
\end{align}
where $\{\lambda_i\}_{i=1}^d$ are the Lyapunov spectrum, and $k$ is the maximum value such that $\sum_{i=1}^{k}\lambda_i > 0$ with sorted Lyapunov exponents: $\lambda_i\ge\lambda_{i+1}$.
Tab.\ref{tab:dim_deltaT} shows the average of $D_L$ calculated this way for each of the 64 trajectories used in
Tab.\ref{tab:lyp_deltaT} for each $\delta t$ value.
Notice that the error in the dimension is considerably smaller and hence providing a stable measure.
\begin{table}[!ht]
\centering
\begin{tabular}{|l|c|}
\hline
$\delta$  & $D_L$                
\\ \hline
0.05      & 2.122$\pm$0.0012  
\\ \hline
0.01      & 2.107$\pm$0.0009  
\\ \hline
$10^{-3}$ & 2.066$\pm$0.0012 
\\ \hline
$10^{-4}$ & 2.061$\pm$0.0012 
\\ \hline
\end{tabular}
\caption{Lyapunov dimensions of Lorentz system with varying sampling step sizes.}
\label{tab:dim_deltaT}
\end{table}

To explore more dynamics features, in the present work we therefore use the Lyapunov dimension to characterize attractors of trained machine.
Though defined completely based on dynamic attributes (i.e. Lyapunov spectrum), $D_L$ also suggests an upper bound for the information dimension of the system, and hence provides a clear description of hidden dynamics.

The above definition in Eq.\ref{eq:dim} requires the complete Lyapunov spectrum, which we implement a numeric approach~\cite{gramshimidt} to obtain.
The calculation applies a Gram-Schimidt orthonormalization on Lyapunov vectors at each step of iterations, to avoid a misalignment of all vectors along the direction of maximal expansion.
If the dynamics is smooth between two arbitrary time steps, the linearization of the iteration map could describe the change of an infinitesimal perturbation through the Jacobian matrix $\mathbf{J}_t$, where entries are defined as:
\begin{align}\label{eq:jacob}
    \big[J_t\big]_{ij} = \frac{\partial u_{t+1,i}}{\partial u_{t,j}},
\end{align}
with $\mathbf{u}_t$ labeling the system state at time $t$.
Suppose $\{\hat{\mathbf{e}}^i_t\}_{i=1}^d$ is a set of base vectors in the state space of hidden variables $\mathbf{h}_t$, then one could approximate the Lyapunov spectrum $\{\lambda_i\}_{i=1}^d$ from the following formula:
\begin{align}\label{eq:exponent}
    \lambda_i = \lim_{T\rightarrow\infty}\lim_{\delta t\rightarrow 0} \frac{1}{T\;\delta t}\sum_{t=0}^T\frac{|\hat{\mathbf{q}}^i_{t+1}|}{|\hat{\mathbf{e}}^i_t|},
\end{align}
where $\hat{\mathbf{q}}^i_{t+1}$ is the Gram-Schimidt orthogonalized vector obtained by rotating the original vector $\hat{\mathbf{e}}^i_{t}$ using the Jacobian $\mathbf{J}_t$:
\begin{align}\label{eq:gs_orth}
    \hat{\mathbf{q}}^1_{t+1} &= \mathbf{J}_t\cdot\hat{\mathbf{e}}^1_{t}, \nonumber \\
    \hat{\mathbf{q}}^2_{t+1} &= \mathbf{J}_t\cdot\hat{\mathbf{e}}^2_{t} - \frac{[\mathbf{J}_t\cdot\hat{\mathbf{e}}^2_{t}]^{T}\cdot\hat{\mathbf{q}}^1_{t+1}}{\hat{\mathbf{q}}^1_{t+1}{}^{T}\cdot\hat{\mathbf{q}}^1_{t+1}}\hat{\mathbf{q}}^1_{t+1}, \nonumber \\
    &\vdots \nonumber\\
    \hat{\mathbf{q}}^i_{t+1} &= \mathbf{J}_t\cdot\hat{\mathbf{e}}^i_{t} - \sum_{j=1}^{i-1} \frac{[\mathbf{J}_t\cdot\hat{\mathbf{e}}^i_{t}]^{T}\cdot\hat{\mathbf{q}}^{j}_{t+1}}{\hat{\mathbf{q}}^{j}_{t+1}{}^{T}\cdot\hat{\mathbf{q}}^{j}_{t+1}}\hat{\mathbf{q}}^{j}_{t+1}.
\end{align}
And new base vectors are obtained after normalization:
\begin{align}\label{eq:norm}
    \hat{\mathbf{e}}^i_{t+1} = \frac{\hat{\mathbf{q}}^i_{t+1}}{|\hat{\mathbf{q}}^i_{t+1}|}.
\end{align}

In real life application scenarios, it is likely that the time interval $\delta t$ of the raw signal between two moments $t$ and $t+1$ is unknown.
In this case, the Lyapunov spectrum calculated in Eq.\ref{eq:exponent} is not available.
However, it is interesting to note that the Lyapunov dimension of an attractor defined in Eq.\ref{eq:dim} does not rely on a specific value of $\delta t$ as long as all time steps are equally spaced.
This suggests the stability of Lyapunov dimension in more general cases, and is one of the most important reason that we choose the dimension rather than exponents or spectrum as the evaluation criteria.

\subsection{RNN Model Trained via Deep Learning Approaches}\label{sec:result:dl_rnn}

Now we evaluate RNN models trained using DL approach introduced in the last section.

\begin{figure}[!ht]
    \centering
    \includegraphics[height=2.8in]{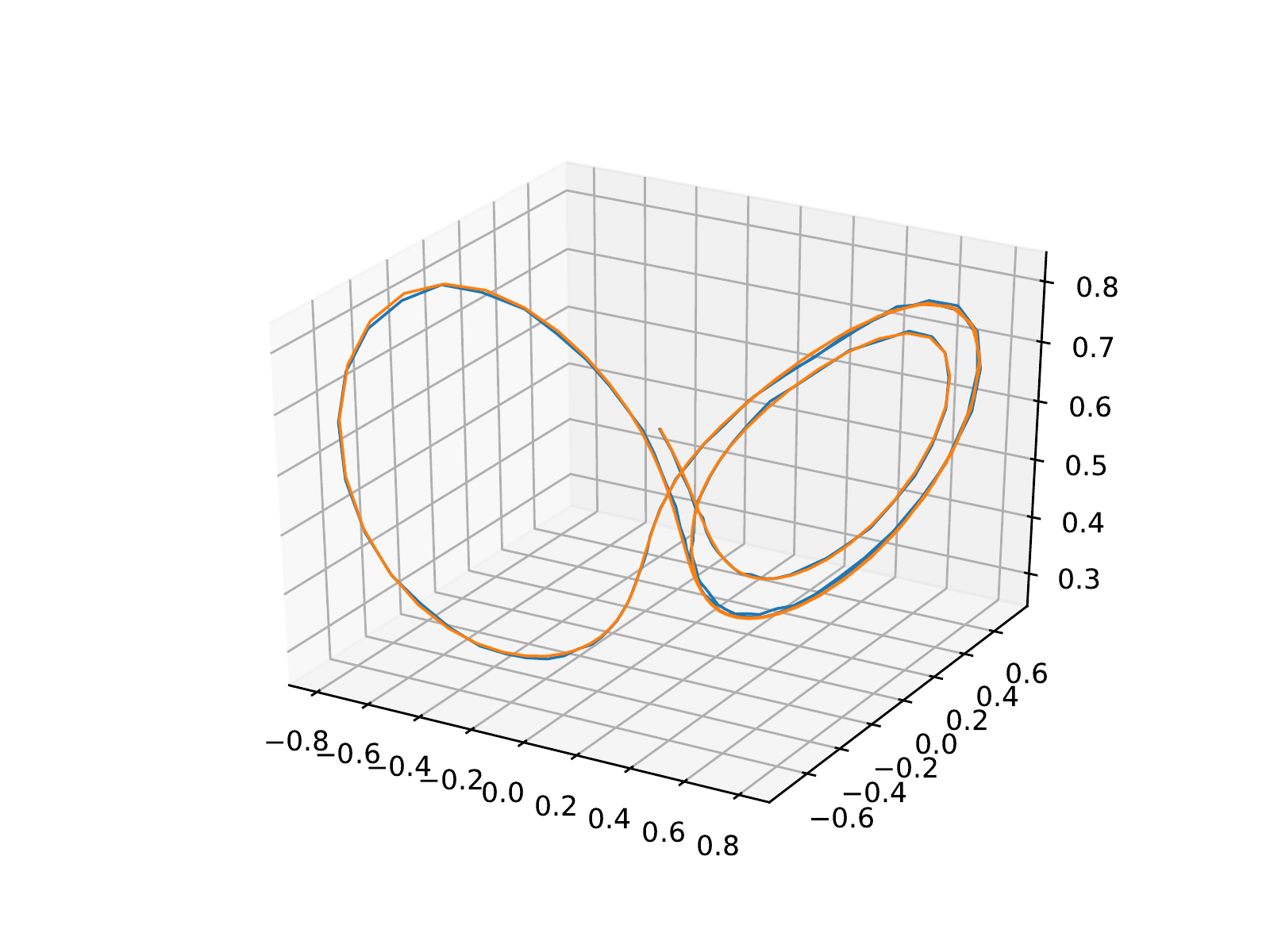}
    \caption{Testing results of a trained machine: comparing the ground truth values and model-generated values on a sequence of length $T_t=150$ steps, with a time interval $\delta t = 2E-2$.}
    \label{fig:TestingCurves}
\end{figure}

We trained RNN models learning Lorentz dynamics using the setup introduced in Sec.\ref{sec:dl:experiment}.
To test the model performance, we examine the autonomous dynamics of trained machines by substituting $\mathbf{h}_{T}$ (i.e. the hidden state obtained by driving the machine with the true data sequence $\{\mathbf{u}_{t}\}_{t=1}^T$) as the initial hidden state.
We then generate a 3-dimensional sequence through the following autonomous dynamics:
\begin{align}
    \mathbf{h}_{t+1} &= (1-\alpha)\mathbf{h}_t + \alpha\tanh{\big[(\mathbf{W} + \mathbf{W}_{in}\mathbf{W}_{out})\cdot\mathbf{h}_t + \mathbf{b}\big]}, \nonumber \\
    \mathbf{u}'_{t+1} &= \mathbf{W}_{out}\cdot\mathbf{h}_{t+1}, \qquad\qquad \forall t\in[T, T+T_t].
\end{align}
The autonomously generated sequence $\mathbf{U}'_{T:T+T_t}$ is then compared with the ground truth sequence $\mathbf{U}_{T:T+T_t}$ generated by Lorentz dynamics. 
Fig.\ref{fig:TestingCurves} shows an instance of model testing with length $T_t=150$.
The negligible deviation suggests that the instance model has indeed learnt a dynamics similar to Lorentz system.
It is worth to mention that the testing setup above implements an autonomous dynamics, or equivalently, targets at a multi-step prediction (i.e. $150$ steps ahead in our setup).
This task is much harder than learning a driven dynamics which in fact only predicts one step ahead.

As elaborated in previous sections, error analysis on a finite length sequence prediction could not serve as a comprehensive model assessment.
In addition to quantifying prediction error, we investigate the attractor behavior of a trained machine to provide more information about the dynamics.
Specifically, we calculate the Lyapunov Dimension of machine attractors as introduced in Sec.\ref{sec:result:attractor}, and compare with the dimension of the original data generating process, i.e. Lorentz system.
The text book value of the strange attractor dimension of Lorentz system is $2.06$, which is based on theoretical derivation of the continuous dynamics.
Since in our study, the training data is obtained from discrete numeric simulation, it is more reasonable to compare the trained machine with the simulation process.
With a time step $\text{5E-2}$, we obtain the approximated attractor dimension of the simulation process as $d\simeq 2.32$.

The Jacobian of a single layer RNN can be derived from Eq.\ref{eq:jacob} and Eq.\ref{eq:auto}:
\begin{align}
    \mathbf{J}_t = (1-\alpha)\mathbb{I} + \alpha \mathbf{D}_{t+1}[\mathbf{W} + \mathbf{W}_{in}\mathbf{W}_{out}],
\end{align}
where we have defined:
\begin{align}
    \big[D_t\big]_{ij} = \big[1-h_{t, i}^2\big] \cdot \delta_{ij}.
\end{align}
The above calculation can be easily generalized to multi-layer RNN models and their variants, e.g. \textit{Long-Short-Term-Machine} (LSTM) and \textit{Gated Recurrent Units} (GRU).
For instance, we could show the calculation of multi-layer RNNs with an arbitrary activation function $Q(\cdot)$. 
For simplicity we assume that different layers share the same width (i.e. same numbers of neurons), a general $L$-layer RNN model then can be written as:
\begin{align}\label{multilayer_rnn}
    h^{(1)}_{t+1} =& Q\big[W^{(1)}h^{(1)}_{t} + W^{(1)}_{in}W_{out}\cdot h^{(L)}_{t} + B^{(1)}\big], \nonumber \\
    h^{(2)}_{t+1} =& Q\big[W^{(2)}h^{(2)}_{t} + W^{(2)}_{in}\cdot h^{(1)}_{t} + B^{(2)}\big], \nonumber \\
    &\vdots \nonumber \\
    h^{(l)}_{t+1} =& Q\big[W^{(l)}h^{(l)}_{t} + W^{(l)}_{in}\cdot h^{(l-1)}_{t} + B^{(l)}\big], \nonumber \\
    &\vdots \nonumber \\
    h^{(L)}_{t+1} =& Q\big[W^{(L)}h^{(L)}_{t} + W^{(L)}_{in}\cdot h^{(L-1)}_{t} + B^{(L)}\big].
\end{align}
Note that the shape of $W_{in}^{l}$ matrix in layer-1 differs from that in other layers: while $W_{in}^{1}$ is $[N\times d]$, where $d$ is the output dimension, other matrices are of shape $[N\times N]$ in the equal-width setup. According to the above expression, we are aware of the following dependence when calculating the matrix $R$:
\begin{equation}
R = 
    \setlength{\tabcolsep}{4pt}
    \centering
    \left[
    \begin{tabular}{cccccccc}
      $\frac{\partial h_{t+1}^{(1)}}{\partial h_{t}^{(1)}}$ & 0 & 0 & $\ldots$ & 0 & 0 & $\ldots$ & $\frac{\partial h_{t+1}^{(1)}}{\partial h_{t}^{(L)}}$ \\
      $\frac{\partial h_{t+1}^{(2)}}{\partial h_{t}^{(1)}}$ & $\frac{\partial h_{t+1}^{(2)}}{\partial h_{t}^{(2)}}$ & 0 & $\ldots$ & 0 & 0 & $\ldots$ & $\frac{\partial h_{t+1}^{(2)}}{\partial h_{t}^{(L)}}$ \\
      $\vdots$ & $\vdots$ & $\ddots$ & {} & $\vdots$ & $\vdots$ & {} & $\vdots$ \\
      $\vdots$ & $\vdots$ & {} & $\ddots$ & $\vdots$ & $\vdots$ & {} & $\vdots$ \\
      $\frac{\partial h_{t+1}^{(l)}}{\partial h_{t}^{(1)}}$ & $\frac{\partial h_{t+1}^{(l)}}{\partial h_{t}^{(2)}}$ & $\ldots$ & $\ldots$ & $\frac{\partial h_{t+1}^{(l)}}{\partial h_{t}^{(l)}}$ & 0 & $\ldots$ & $\frac{\partial h_{t+1}^{(l)}}{\partial h_{t}^{(L)}}$ \\
      $\vdots$ & $\vdots$ & {} & {} & $\vdots$ & $\ddots$ & {} & $\vdots$ \\
      $\vdots$ & $\vdots$ & {} & {} & $\vdots$ & {} & $\ddots$ & $\vdots$ \\
      $\frac{\partial h_{t+1}^{(L)}}{\partial h_{t}^{(1)}}$ & $\frac{\partial h_{t+1}^{(L)}}{\partial h_{t}^{(2)}}$ & $\ldots$ & $\ldots$ & $\frac{\partial h_{t+1}^{(L)}}{\partial h_{t}^{(l)}}$ & $\frac{\partial h_{t+1}^{(L)}}{\partial h_{t}^{(l+1)}}$ & $\ldots$ & $\frac{\partial h_{t+1}^{(l)}}{\partial h_{t}^{(L)}}$
    \end{tabular}
    \right] \nonumber
\end{equation}
which then leave us the task for calculating the following four different types of terms:
\begin{align}
    \text{diagonal terms: }&\quad\frac{\partial h_{t+1}^{(l)}}{\partial h_{t}^{(l)}}; \\
    \text{last-column terms: }&\quad\frac{\partial h_{t+1}^{(l)}}{\partial h_{t}^{(L)}}; \\
    \text{last term: }&\quad\frac{\partial h_{t+1}^{(L)}}{\partial h_{t}^{(L)}}; \\
    \text{ordinary terms: }&\quad\frac{\partial h_{t+1}^{(l)}}{\partial h_{t}^{(m)}}, \qquad \forall l > m+1.
\end{align}
Let us derive the formula for each case. We define the diagonal matrices:
\begin{align}
    \big[H_{t}^{(l)}\big]_{ij} = 
    \left\{
    \begin{tabular}{ll}
        $\big[1- \big(h^{(l)}_{t, i}\big)^2\big]\cdot \delta_{ij}$, & if $Q(\cdot) = \tanh{(\cdot)}$; \\
        $h^{(l)}_{t, i}\cdot\big[1- h^{(l)}_{t, i}\big]\cdot \delta_{ij}$, & if $Q(\cdot) = \sigma(\cdot)$;
    \end{tabular}
    \right. \nonumber
\end{align}
along with the following matrices:
\begin{align}
    M_{t}^{(l,m)} = \prod_{n=l}^{m+1} H^{(n)}_{t}\cdot W_{in}^{(n)}\cdot, \qquad \forall l>m+1,
\end{align}
where the series product indexed by $n$ is in descending order from $l$ to $m+1$, and the dot $\cdot$ denoted at the end of the matrices pair indicating the new terms are always attached from the right hand side. With the help of the above defined terms, we can write all derivatives as following:
\begin{itemize}
    \item \textbf{diagonal terms}:
    \begin{align}
        \frac{\partial h_{t+1}^{(l)}}{\partial h_{t}^{(l)}} = H_{t+1}^{(l)}\cdot W^{(l)}, \qquad \forall l < L;
    \end{align}
    
    \item \textbf{ordinary terms}:
    \begin{align}
        \frac{\partial h_{t+1}^{(l)}}{\partial h_{t}^{(m)}} = M_{t+1}^{(l, m)}\cdot H_{t+1}^{(l)}\cdot W^{(l)}, \qquad \forall l > m+1;
    \end{align}

    \item \textbf{last-column terms}:
    \begin{align}
        \frac{\partial h_{t+1}^{(l)}}{\partial h_{t}^{(L)}} = M_{t+1}^{(l, 0)}\cdot W_{out}, \qquad \forall l < L;
    \end{align}
    
    \item \textbf{last term:}
    \begin{align}
        \frac{\partial h_{t+1}^{(L)}}{\partial h_{t}^{(L)}} = H_{t+1}^{(L)}\cdot W^{(L)} + M_{t+1}^{(L, 0)}\cdot W_{out}.
    \end{align}

\end{itemize}

\begin{figure}[!ht]
    \centering
    \includegraphics[height=2.3in]{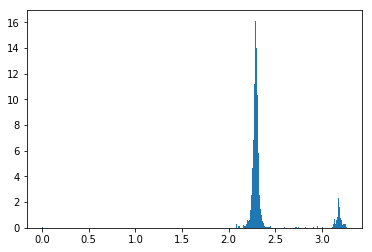}
    \caption{Distribution of attractor dimensions calculated on 1100 trajectories generated by 55 trained machines.}
    \label{fig:DimStats_train}
\end{figure}

As Lyapunov dimension of a specific trajectory from a single machine does not provide sufficient information about the learnt dynamics, we run experiments on multiple machines to apply a statistical investigation.
Based on Eq.\ref{eq:dim} and Eq.\ref{eq:exponent}, we calculate the attractor dimension from an average over $1500$ steps.
The eventual statistics is derived from $3000$ trajectories generated by $300$ independently trained machines.
The distribution of attractor dimensions is show in Fig.\ref{fig:DimStats_train}.
The majority of attractors' fractal dimensions is centered around $d\approx 2.3$, with a small portion centered at $d\approx 3.2$.
It can therefore be concluded that the deep learning based approach is indeed able to, with a large probability, obtain a model dynamics which produces attractors with a fractal dimension quite similar to the dynamics of the data generation process.
Compared with the above error analysis on a finite sequence, attractors analysis captures longer term and asymptotic behavior of the trained machine.
Besides, if the length of the time step was given, based on Eq.\ref{eq:exponent}, one could roughly infer an approximate rate of error accumulation or explosion.
This is quite useful in deriving an error bound for long term prediction tasks.

\subsection{Comparing with Fitted RC}

To understand the advantage of the proposed DL based approach for learning dynamics, we further compare it with the other alternatives.
Different from more conventional DL studies, we are not interested in comparing model structures but, rather, would investigate the impact of various training approaches given the same model structure, which, in our current work, is vanilla RNN.

\begin{figure}[!ht]
    \centering
    \includegraphics[height=2.3in]{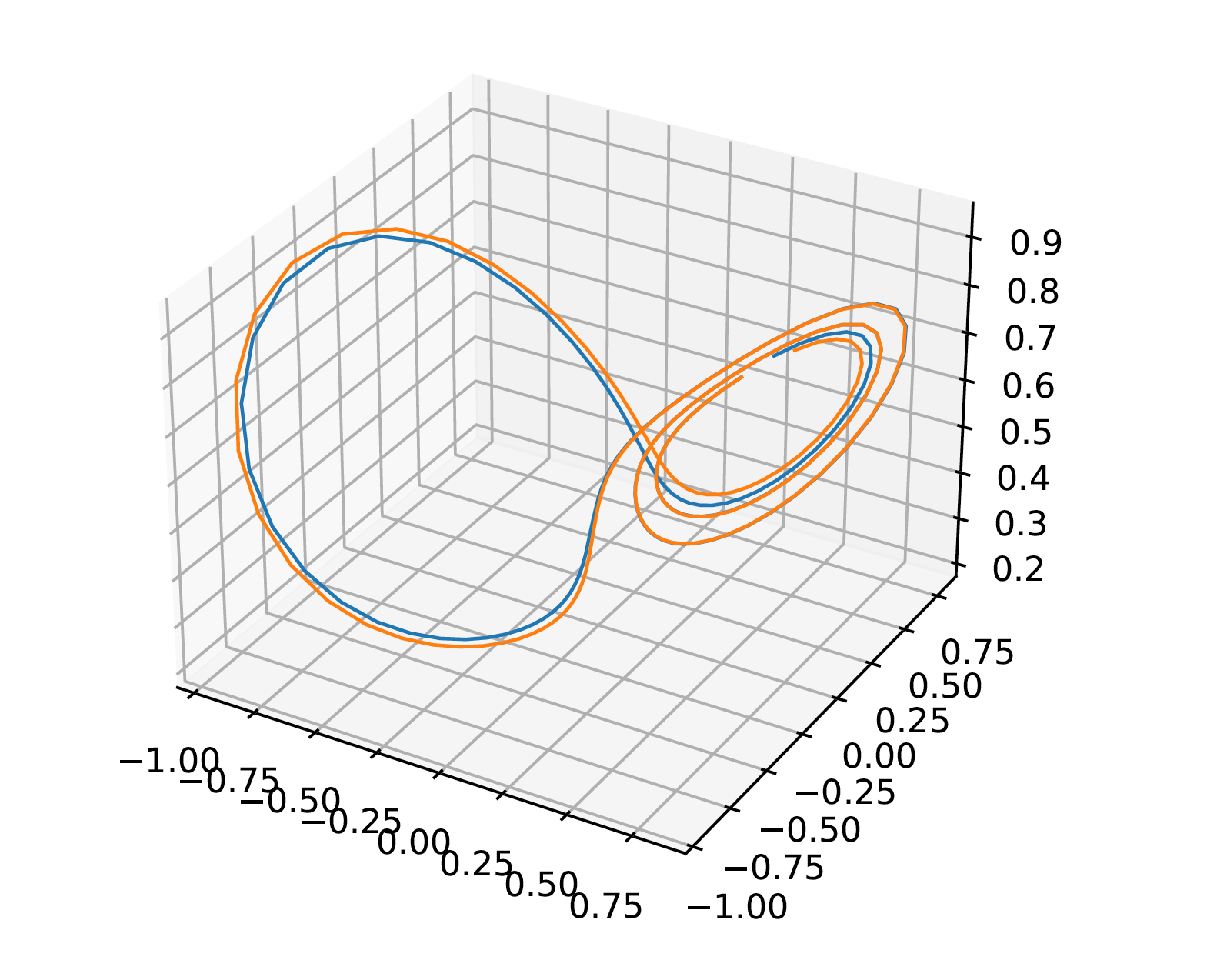}
    \caption{Distribution of attractor dimensions calculated on 1100 trajectories generated by 55 fitted machines.}
    \label{fig:testplot_fit}
\end{figure}

\begin{figure}[!ht]
    \centering
    \includegraphics[height=2.3in]{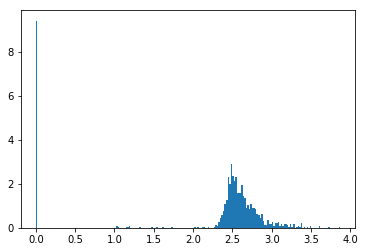}
    \caption{Distribution of attractor dimensions calculated on 1100 trajectories generated by 55 fitted machines.}
    \label{fig:DimStats_fit}
\end{figure}

It is interesting to compare with models obtained via a linear fitting procedure in Eq.\ref{eq:fitting}.
This procedure produces a $\mathbf{W}_{out}$ matrix alone.
Since the data matrix $\mathbf{u}$ is produced by a driven dynamics Eq.\ref{eq:rnn}, the procedure is actually minimizing errors in a series of 1-step-ahead forecasting tasks.
The procedure is widely applied in RC research and has been proven to be an efficient method for modeling dynamics.
The error in a validation phase (i.e. one-step prediction) is negligible, and could even remain to be small in a testing phase (i.e. multi-step prediction) through a few more steps further~\cite{mackyglass}.
A typical testing performance of a fitted machine is in Fig.\ref{fig:testplot_fit}. 
Similar to the trained machine, the deviation from the ground truth trajectory is also small, and therefore the fitted machine could still provide some insights for the prediction task.
To further investigate the true dynamics learnt by a fitted model and compare with the DL training method, we examine 1100 machines in total with different random matrices $\mathbf{W}$ by calculating attractor dimensions.
The statistics of Lyapunov dimensions is shown in Fig.\ref{fig:DimStats_fit}.
There is a portion of machines producing attractors with dimension values of $0$ and $1$, which are actually fixed points and limit cycles\footnote{For attractors with integer dimensions, the Eq.\ref{eq:dim} does not provide a correct answer. However, one could easily identify asymptotic behaviors as fixed points or limit cycles by running the autonomous dynamics longer in time.}.
This suggests that machines obtained via the above fitting procedure could not always capture the true hidden dynamics of the data generation process.
At the same time, fitted models also provide instantiation cases where the error analysis is not sufficient, or even misleading, for model assessment, while a dynamical system perspective instead could provide a better understanding about what has been learnt.
In practice, although the testing error (i.e. deviations in prediction tasks) could be further minimized by applying a larger hidden dimension $d$, the statistics of Lyapunov dimensions does not exhibit any significant improvement.

There are several limitations of the fitting method that could potentially affect the "learning capability" of the model.
Firstly, both the recurrent connection $\mathbf{W}$ and the input projection $\mathbf{W}_{in}$ are fixed from a random initialization, which restricts the overall model performance to be strongly dependent on the randomization.
The more complex an internal connection is, the easier the model could produce a non-trivial dynamics through a output projection $\mathbf{W}_{out}$.
This is confirmed by examining models with different hidden dimension values $d$: models with higher hidden dimensions exhibit both a better performance in testing phase and a larger portion of correct attractor dimensions (or, equivalently, a smaller portion of trivial attractors such as fixed points and limit cycles).
Secondly, the fitting method minimizes errors on a single trajectory with a finite length $T$.
As mentioned earlier in Seq.\ref{sec:dl:batch}, a trade-off between the amount of data (i.e. the length $T$ of the fitting sequence) and computation accuracy is inevitable, while complex dynamics usually could not be captured by only a small portion of an attractor.
Thirdly, as the objective of fitting consists of errors in 1-step-ahead predictions, the autonomous dynamics in Eq.\ref{eq:auto} is not directly addressed, which makes it harder to learn the true dynamics.

While the first two limitations above can be easily resolved via a DL implementation, it is interesting to analyze the impact of the third one within a DL setup.
More specifically, while still applying the \textit{PyTorch} DL framework, we train RNN models by minimizing the 1-step-ahead loss function in Eq.\ref{eq:loss_1step} instead of Eq.\ref{eq:loss_auto}.
When same hyperparameters are applied (hidden dimension $d$, residual connection $\alpha$, etc.), it turns out that the setup could hardly produce a model with good performance.
Compared with our proposed Seq2Seq setup, the only difference is the loss function.
Therefore this comparison clearly reveals the problem of 1-step-ahead setup in learning dynamics.

\section{Discussions and Future Work}

We now have delivered a hybrid approach combining the powerful modeling capability of deep learning and a comprehensive model evaluation analysis provided by dynamical system theory.
In the implementation of DL based training, we apply several crucial techniques, including residual connection, Seq2Seq training, spectral initialization, and batch learning, associated with a discussion of motivation from both a DL perspective and a dynamical system one.
By evaluating trained models based on both the standard multi-step test and attractors' asymptotic behavior, it can be concluded that the proposed DL method could not only accomplish the traditional time series prediction task, but also successfully capture the underlying data generating dynamics.
Compared with the linear fitting method widely used by RC community, DL based learning could make advantage of information collected from a larger portion of phase space. Besides, different from the fitting method where both $\mathbf{W}$ and $\mathbf{W}_{in}$ are fixed by a random initialization, all matrices are learnable in DL training framework and therefore are capable of capturing more complex dynamics while not limited by the odd of proper random seeds generated in the initialization.

Aside from the model performance, the asymptotic behavior analysis from the dynamical system theory perspective has benefited the learning in several aspects.
Models with seemingly good performance but improper internal dynamics can be ruled out by calculating their attractor dimensions.
At the same time, when time interval between steps are known, one could estimate the rate of error accumulation using approximate Lyapunov exponents.

The present study has already implied several intriguing questions.
On the one hand, asymptotic behavior of different random matrices in RNN has been studied in dynamical system theory.
It has been known that in large $d$ limit, both connectivity and generating distribution of the random matrix $(\mathbf{W}+\mathbf{W}_{in}\mathbf{W}_{out})$ would affect the resulting internal dynamics significantly.
These studies can benefit the initialization in DL method to accelerate the global convergence of training.
On the other hand, since, as discussed earlier, a trained RNN model can actually be deemed as a representation of the generating dynamics, it is hence valuable to infer information about the data generation process by measuring the trained model.
It is well-known that, as suggested by Kaplan-Yorke conjecture, the attractor dimension defined in Eq.\ref{eq:dim} is closely related to information dimension as well.
Therefore, it would be interesting to investigate the property of a data generation process with an effective dimension $D_L$ (i.e. the attractor dimension) which is either lower or higher than the original input dimension $k$.
In this way, with the assistance of dynamical system theory, one could investigate further into the data generation process rather than limited to a single DL model with seemingly good forecasting performance.

\setcounter{equation}{0}
\chapter{Algebraic Learning: a Novel Scheme for Statistical Learning}

In the three chapters above, we have demonstrated interpretable information modeling with a multi-scope investigation.
With regard to the type of signals\footnote{More precisely, the correlation in signals.}, Chapter 2, 3, and 4 explore problems with discrete correlation, spatial order, and temporal order, respectively.
From the perspective of interpretability methodologies, Chapter 2 and Chapter 3 exhibit the application of GuiMoD, while Chapter 4 implements an indirect $\partial$M.
While these three studies all exploit mathematical concepts, different areas of mathematics are covered: algebra emerges as the natural language for KGE problems in Chapter 2; geometry is of the central concern in image processing as discussed in Chapter 3; and techniques of analysis assist the model evaluation in Chapter 4.

With these instantiation studies of specific problems as concrete examples, we now zoom in the practice of GuiMoD, and reappraise the problem of information modeling from a more popular yet abstract perspective: artificial intelligence (AI).
Instead of debating the appropriate definition of intelligence, here our discussion would focus on a blueprint where machines attempt their best to imitate human behavior to understand the universe.
In real world scenarios, human draws conclusions based on two key modules~\cite{danks2009psychology}: \emph{perception}, and \emph{reasoning}.
Basically, the perception module provides atomic information pieces, i.e. entities, as the elementary input, and the reasoning module applies logic rules, by identify existing relations between observed entities, to derive a conclusion.
We would like to discuss the equivalent modules in AI.

\section{Perception Module in Intelligence}

A central question in perception module, is the granularity level of information piece.
Human does not perceive individual variables, e.g. single pixel in images or single letter in words, directly as meaningful signals.
And, as discussed earlier, high-dimensional signals in real world are usually highly organized with certain correlations/orders, e.g. spatial or temporal orders.
Indeed, more generally, organized low level variables could compose high level ones.
For instance, in the problem of face recognition, pixels could be organized to represent organs, e.g. eyes, ears, and noses, while organs could be further organized to compose a face.

To imitate the perception of human, \emph{machines are also expected to perceive signals of proper levels which contain meaningful information, rather than low level variables}.
To achieve this, on the one hand, tasks should be designed to offer explicit instructions to machines about level of signals to comprehend, which promotes the idea of supervised learning and labeled data;
on the other hand, \emph{the model structure should be explicitly designed to capture correlations and orders in high dimensional signals}, which often produces explicit constraints on the model structure and is exactly a deductive process of GuiMoD.

Now let us formalize the problem of a perception model design more explicitly.
Given a input signal as a set of variables in $\mathcal{F}$: $\mathbf{X}\equiv\{x_i\}_{i=1}^n$, where $\forall i, x_i\in\mathcal{F}$, a perception model aims at identifying a set of entities $\mathbf{E}_X=\{\mathbf{e}_{\alpha}\}_{\alpha=1}^K$ which are represented by $\mathbf{X}$ at certain granularity level.
Clearly, the set of variables $\mathbf{X}$ are not independent to each other.
The granularity level is also the level at which the desired "information" exhibits, and is determined by the scale $l$ of correlations between different $x_i$'s.
However, the \textit{correlations} discussed here is usually highly nonlinear and multi-variate, and is therefore difficult to be described by a traditional statistical approaches.

To formulate the discussion, we consider a transformation $g: \mathcal{F}^n\mapsto\mathcal{F}^n$:
\begin{align}
    g:\mathbf{X} \mapsto g\circ\mathbf{X} &\equiv \hat{\mathbf{X}}.
\end{align}
Clearly, an arbitrary transformation does not produce a new signal with meaningful information.
For example, a random permutation of letters in a sentence might result into nothing, while a permutation of words may indeed produce a new sentence.
The legitimacy of a transformation is determined by the task: for instance, permutations of sentences or paragraphs are semantically less volatile then permutations of words.
More explicitly, a transformation $g$ is considered as legitimate if the transformed signal $\hat{\mathbf{X}}$ still contains a set of meaningful entities $\hat{\mathbf{E}}_{\hat{X}}\equiv\{\mathbf{\hat{e}}_{\beta}\}_{\beta}^{K'}$.
The above discussion indicates that, given a perception task $\mathcal{T}$, we are in fact mostly interested in the set of legitimate transformations $\mathbf{G}_{\mathcal{T}}$, which captures correlations among variables in an implicit way.

Now we return to the discussion of GuiMoD for perception models.
A DL based perception model $\mathbf{M}$ takes a signal $\mathbf{X}$ as input, and outputs a set of labels representing identified entities $\mathbf{E}_M$:
\begin{align}
    \mathbf{M}: \mathbf{X} \mapsto \mathbf{E}_M(\mathbf{X})\equiv\{\mathbf{e}_{\alpha}\}_{\alpha=1}^m.
\end{align}
A model is perfect if at the data point $\mathbf{X}$:
\begin{align}
    \mathbf{E}_M(\mathbf{X}) = \mathbf{E}_X.
\end{align}
In the ideal case, one hopes the above relation holds for arbitrary $\mathbf{X}$.
While in practice, an arbitrary value assignment in $\{x_i\}_{i=1}^n$ does not produce a meaningful $\mathbf{X}$, and, one should only focus on the set of meaningful $\mathbf{X}$'s.
As discussed above, a legitimate transformation $g$ in the task specified set $\mathbf{G}_{\mathcal{T}}$ could transform a meaningful signal into another one.
Therefore, the model $\mathbf{M}$ is expected to perform ideally such that:
\begin{align}
    \mathbf{E}_M(g\circ\mathbf{X}) = \mathbf{E}_{g\circ X},\qquad \forall g\in\mathbf{G}_{\mathcal{T}}.
\end{align}
One should immediate notice that the above equation is actually a \emph{generalized equivariance condition} similar to our discussion in Chapter 3.
Instead of a group structure, in generic situations, the set $\mathbf{G}_{\mathcal{T}}$ could be any type of algebraic structures.
The reason that algebra emerges in the discussion here leads to a very noteworthy observation:
\emph{instead of studying changes due to a single transformation, one should actually focus on the relation between changes brought by different transformations}.

From a highly abstract level, the above discussion demonstrates the emergence of algebra in perception module and hence suggests the necessity of \emph{algebra-inspired model design}.

\section{Reasoning Module in Intelligence}

The reasoning module is based on perceptions, and explores \textit{relations} between observed entities, based on which the final conclusion could be drawn.
Although the word \textit{relation} appears again, we would like to further elaborate the meaning of it in the current context of reasoning.
Reasoning is an action that exploits logic to process information until conclusions being derived.
It outputs a value of true or false for a query $\mathbf{q}$, based on the logic process.
Therefore, to efficiently operate the reasoning process, a model should capture logic either explicitly or implicitly.

Logic is expressed by logic rules, which are composed of \emph{atoms} along with logic operators: conjunction $(\wedge)$, disjunction $(\vee)$, and negation $(\neg)$.
An \emph{atom}, $\mathbf{\alpha}$, consists of a predicate $P_{\alpha}$ associated with a set of terms $\mathbf{T}_{\alpha} = \{t_i\}_{i=1}^n$, where $P_{\alpha}$ maps $\mathbf{T}_{\alpha}$ to a logic variable $z$, which can be either binary $\{0,1\}$ in traditional logic or continuous $[0,1]$ in fuzzy logic.
Triplets in knowledge graphs can be viewed as a special type of atoms, whose predicates $P_{\alpha}$ is bi-variate, i.e. always taking two terms, the head entity $\mathbf{e}_h$ and the tail entity $\mathbf{e}_t$, as arguments.
And therefore the hyperrelation we proposed in Chapter 2 is a special type of logic rules, which is always chain-like and uses only conjunctions between atoms.
When all terms $\mathbf{T}_{\alpha}=\{t_i\}_{t=1}^n$ in an atom $\mathbf{\alpha}$ are replaced by constants, we call it a \emph{grounded atom}, and it becomes a \emph{fact} $f$ if its value holds true.
A general logic rule could be written in the following form:
\begin{align}
    \alpha_0 \dashv (\alpha_1, \alpha_2, \cdots, \alpha_m),
\end{align}
meaning the rule head $\alpha_0$ would be true iff all $\{\alpha_l\}_{l=1}^m$ in the rule body are true.

Now we discuss the problem of machine reasoning, i.e. to build a model which could operate the reasoning process.
As defined above, the targeted model $M_q$ should output a logic variable $z$ for a query $\mathbf{q}$.
Different from a proof construction in theorem provers, where an explicit proof path should be generated, machine reasoning may focus on the satisfaction score of the query alone, which is the conclusion.
To achieve the conclusion, during the model construction, $M_q$ should be fed with both facts and rules.
Only after both information being digested by $M_q$, it could become capable of reasoning for an input query $\mathbf{q}$.
While the feed of facts is always explicit, it is possible to feed $M_q$ with rules either explicitly or implicitly.
In traditional symbolic reasoning and programming, rules are explicitly fed into the module.
However, in practice, the number of rules could be extremely large, and a manual collection/construction would be intractable.
With the amount of available data increasing rapidly, a statistical learning based alternative for rules comprehension now has drawn significant attentions.
Basically, given a dataset of facts $\mathbf{F} = \{f_s\}_{s=1}^K$ which implicitly contains logic rules, the goal of a learning based approach is to train a model to comprehend rules hidden in $\mathbf{F}$ automatically.
After training, the trained model is expected to capture both facts and rules at the same time.

Note that the formulation of reasoning is very broad. For example, we could reformulate the supervised learning as a reasoning problem: a supervised learning model takes $\mathbf{x}_q$ as input and outputs $\mathbf{y}$, which could either be a classification or a regression.
In the context of reasoning, we could regard input features $\{x_i\}_{i=1}^n$ as a grounding of term variables $\mathbf{T}_{q}=\{t_i\}_{i=1}^n$ in the query atom $\mathbf{q}$, and the model $M_q$ implicitly captures the reasoning process which takes facts in and concludes if or not the query is true.

The reasoning of a query $\mathbf{q}$, in fact, is a process to check all possible proof paths, which forms a set $S_{q}$, termed as the \emph{complete proof set} of the query.
A proof of a query $\mathbf{q}$ could be constructed by proofs of its sub-queries $\{S_{q'}\}$, analogous to the relation between a theorem and lemmas.
In fact, given a task $\mathcal{T}$ defined by a set of facts and rules implicitly contained in a training dataset, it is proved that the collection $\mathbf{S}_{\mathcal{T}}=\{S_q\}_q$ of complete proof sets of all callable queries forms a \emph{semiring} structure\cite{huang2021scallop}, which is termed as the \textbf{proof semiring}, with each $S_q$ being an algebraic element while $\otimes$ and $\oplus$ resemble $\wedge$ and $\vee$ respectively.
For machine reasoning, the model $M_q$ trained for $\mathbf{q}$ plays the role of the complete proof set $S_q$
which is an element in $\mathbf{S}$, shown as the correspondence below:
\begin{align}
    M_q \quad \longleftrightarrow \quad S_q.
\end{align}
%That is to say, the learnt model correponds to an algebraic element on the proof semiring $\mathbf{S}$.
The neurons in the final layer of $M_q$ corresponds to the satisfaction of the query $\mathbf{q}$.
Since each $S_q$ could be further decomposed into sub-queries, it is natural to view sub-modules in $M_q$ as the corresponding equivalence of $\{S_{q'}\}$:
\begin{align}
    \{M_{q'}\}\quad \longleftrightarrow \quad \{S_{q'}\},
\end{align}
which may be directly related to neurons in hidden layers.
Taking all callable queries into consideration, one could obtain a set of models correspondingly, termed as $\mathbf{M}_{\mathcal{T}}$, which is isomorphic to $\mathbf{S}$ and therefore forms a semiring structure.

The above discussion, in fact, implies a new explanation of DL based models, from the perspective of reasoning.
Neurons, or sub-modules, are related to algebraic elements in a semiring.
The fact that both $\otimes$ and $\oplus$ are necessary operations to form a semiring, implies that corresponding arithmetic operations among neurons are favorable, and therefore provides a guidance for the model (operation) design.

\section{Algebraic Learning}

In Chapter 1, we introduced the idea of GuiMoD without an explicit recipe to implement in practice.
With the above discussion on two key modules of general intelligence, we now deliver a unified formalism for practicing GuiMoD.
Specifically, the observation of emergent abstract algebraic structures in both perception and reasoning modules of intelligence implies the importance of algebra in general modeling practice, motivated by which, we propose a novel modeling scheme for statistical learning: \textbf{\emph{Algebraic Learning} (AgLr)}.

As demonstrated respectively in the two modules, our formal analysis does not focus on any specific model or dataset.
Given a task $\mathcal{T}$, in perception module, we studied the set $\mathbf{G}_{\mathcal{T}}$ of transformations of data which produce a set of legitimate signals, while in reasoning module we investigated the set of models $\mathbf{M}_{\mathcal{T}}$ resulted from the set of all possible callable queries.
In fact, both cases are related to redundancy of certain space: the former is rooted in the redundancy of the data space, while the latter is related to the redundancy of the model space.
The huge amount of redundancy is either data or model spaces severely prohibits an efficient modeling practice.
It is therefore important to identify a target subset from the redundant space, e.g. $\mathbf{G}_{\mathcal{T}}$ with legitimate signals and $\mathbf{M}_{\mathcal{T}}$ of callable queries.
The fact that elements in the resulting set are not independent but can be related by certain operations leads to an emergence of certain type of abstract algebraic structure $\mathbb{A}$.
We can now define a general implementation of AgLr formalism as a practice of GuiMoD:
\begin{Def}
    Given a task $\mathcal{T}$, the idiosyncrasies of $\mathcal{T}$ would declare a set of legitimate signals or models which in general forms an abstract algebraic structure $\mathbb{A}_{\mathcal{T}}$.
    Based on a parametrization of $\mathbb{A}_{\mathcal{T}}$, one could derive explicit constraints on the model parameter space, which therefore provides a specific guidance for model design by identifying a legitimate model set $\mathcal{M}_{\mathcal{T}}$.
\end{Def}

The above formal discussion explains the origin of AgLr and its potential advantage to improve both efficiency and interpretability in generic information modeling.

\end{sloppypar}

\end{document}